\documentclass{article}

\usepackage{arxiv}

\usepackage[utf8]{inputenc} 
\usepackage[T1]{fontenc}    
\usepackage{hyperref}       
\usepackage{url}            
\usepackage{booktabs}       
\usepackage{amsfonts}       
\usepackage{nicefrac}       
\usepackage{microtype}      
\usepackage{lipsum}         
\usepackage{graphicx}
\usepackage{natbib}
\usepackage{doi}
\usepackage{subfigure}

\usepackage{amsmath}
\usepackage{cleveref}       
\usepackage{algorithm}
\usepackage{amsthm} 
\usepackage[shortlabels]{enumitem}

\newtheorem{theorem}{Theorem}
\newtheorem{lemma}[theorem]{Lemma}
\newtheorem{definition}[theorem]{Definition}
\newtheorem{remark}[theorem]{Remark}

\newcommand{\sectionref}[1]{Sect.\,\ref{#1}}
\newcommand{\figureref}[1]{Fig.\,\ref{#1}}
\newcommand{\equationref}[1]{Eq.\,\ref{#1}}
\newcommand{\algorithmref}[1]{Algorithm\,\ref{#1}}
\newcommand{\theoremref}[1]{Theorem\,\ref{#1}}
\newcommand{\lemmaref}[1]{Lemma\,\ref{#1}}
\newcommand{\tableref}[1]{Table\,\ref{#1}}
\newcommand{\abovestrut}[1]{\rule{0pt}{#1}}  
\newcommand{\acks}[1]{Acknowledgments: #1}

\graphicspath{{PICS/}}

\begin{document}

\title{Semiotics Networks Representing  Perceptual Inference}

\author{
\textnormal{David Kupeev}\thanks{Corresponding author}\\
Independent Researcher, Israel\\
kupeev@gmail.com \\ \\
\and 
Eyal Nitzany\\
Independent Researcher, Israel\\
eyalni@gmail.com \\ \\
}

\maketitle

\begin{abstract}

Every day, humans perceive objects and communicate these perceptions through various channels. 
In this paper, we present a computational model designed to track and simulate the perception of objects, 
as well as their representations as conveyed in communication.

We delineate two fundamental components of our internal representation, termed "observed" and "seen", 
which we correlate with established concepts in computer vision, namely encoding and decoding. 
These components are integrated into semiotic networks, which simulate perceptual inference of object perception and human communication.   

Our model of object perception by a
person allows us to define object perception by
{\em a network}. We demonstrate this
 with an example of an image baseline
  classifier by constructing a new network that includes the baseline
   classifier and an
additional layer. This layer produces
the images "perceived" by the entire network,
transforming it into a perceptualized image classifier.
This facilitates visualization of the acquired network.

Within  our network, the image representations become more efficient
 for classification tasks when they are assembled and randomized. In our
  experiments, the perceptualized network outperformed the baseline
   classifier  on a small training dataset. 

Our model is not limited to persons and
 can be applied
to any  system  featuring a loop involving the processing from
 "internal" to "external" representations.

\end{abstract}

\vspace{-0.2\baselineskip}
\hspace{0.9\baselineskip}
\begin{minipage}{0.9\linewidth}
  \textbf{Keywords:}
Network awareness, network interpretability, semiotic network, 
perceptualized classifier, limited training data
\\
dialog semiotics,
perceptualized classifier, limited training data
\end{minipage}

\section{Introduction}
\label{sec:intro}

Perception of objects by persons can be thought of as an internal representation of the outer world, which can be communicated via various modalities (for example, text, sound, vision, etc.). Furthermore, the same object can be described in different channels. For example, an image of a dog, or barking sound would set us to believe that a dog is around.

Perception possesses several properties,
which are in general agnostic to the modality
of the perceived input channel.
First, perception is mostly subjective.
This means that a specific object that
is perceived in on manner,
may be perceived differently by another person.
In other words, two persons may have different internal representation of the same object. For example, two persons that observe a dog might think that this is a nice dog (the first person) or a frighten dog (the second). Although, they both "observe" the same object (dog), they attend to different properties and thus may "see" other aspects of it. These "observe" and "seen" representations are the building block in our model and are used to mimic human perception. They enable one to observe an object and transform ("see") it.

Furthermore, this process can be applied to model 
human visual perception when only a single person 
is involved. We refer to this as an "internal cycle."
During this process,
an object is perceived (observed), projected
onto the "internal space," and this
representation is then used as an observed
input to generate another internal
representation in a cycle, until the perception
act terminates. It is important to note that
this process is typically internal and not
visible externally. However, our model allows
for the exposure of its internal representation,
illustrating its progression. For instance,
\figureref{fig:gradual} demonstrates the
enhancement in the quality of the internal
representation.

The process of converting an "observed" input into  
something "seen" is not restricted to specific  
modalities; rather, it can occur across different  
modalities, such as text and image, or across  
multiple modalities simultaneously. For instance,  
when sound and image interact to form a unified  
perception, a person might hear barking, later see  
a dog, and infer that the dog they now observe is  
the one responsible for the barking. Importantly,  
the internal, personal representation of this  
process remains concealed and inaccessible to  
others. 
Instead, a higher-level representation    
emerges, serving as a shared basis for  
communication between individuals or systems. 
This example illustrates the
framework of our model. 
The  
key idea is that, regardless of the input modality,  
once information enters the system, it is  
transformed into an internal representation that  
propagates through the system in an observe-to-seen  
cycle. This internal representation also enables  
the combination of information from different  
modalities or sensors, allowing for a more  
integrated and holistic understanding. Such a framework accommodates multiple modalities and typically concludes with retranslating the internal representation into its original modality, though this step is not always necessary.

In recent years, attention mechanisms have been effectively integrated into the field of computer vision, with transformer-based architectures outperforming their predecessors. The attention mechanism enables parallel processing and leverages context, but it comes with significant computational demands and often lacks interpretability. In this work, we introduce the CONN mechanism---a lightweight attention module designed to focus on specific, known examples. It operates iteratively, mimicking the sequential behavior of multiple attention layers in a more interpretable and resource-efficient manner. Additionally, one can halt the process at any stage and obtain a result that, while potentially less accurate, still aligns with the desired direction. The longer the mechanism operates and revisits the example, the more reliable and confident the outcome becomes, reflecting the model's increasing certainty.

Recently, Large Language Models (LLMs) 
have garnered significant attention 
within the research community, 
emerging as the primary tools for diverse tasks 
(\citet{NEURIPS2020_1457c0d6,pmlr-v139-radford21a}). 
Notably, the advent of 
multi-modality models has 
expanded their capabilities, enabling them to 
engage with various modalities within their 
internal space 
(\citet{10386743}). Our model aligns 
with this trend, leveraging both internal 
and external representations to 
facilitate communication and perception. 
Consequently, CONNs may be useful for 
analysis of LLMs and other multi-modality models.

The  model of human communication
presented in this article was developed
to represent the existence of
the objects
seen by a person, as well as the existence of objects that the person is aware are being seen
(\sectionref{sec:rikuz:semiotics:person-object}).
Further, the mathematical relations have been obtained describing
other semiotic phenomena of the inter-person communication
(\sectionref{sec:rikuz:semiotics:person to person}).
It worth to note that initially, these new aspects
were not the focus of our attention.
The  ability
to describe supplementary phenomena testifies to the effectiveness of the model.

Awareness, as defined, encompasses the "knowledge or perception of a situation or fact"
(\citet{awareness}).
In this paper, however, 
using our "observed-to-seen"
functional model,
we employ the term "awareness" in a more restricted sense. Here, it signifies the expectation that certain concepts will align with specific instances, occasionally manifesting as particular perceptions. For example, the sound of barking and the image of a dog are anticipated to converge in the recognition of a dog. It is important to note that this use of "awareness" does not inherently extend to emotional (or other) responses, though it can. For instance, an image of a menacing dog might evoke fear, while seeing one’s own dog could elicit affectionate feelings. Throughout this paper, "awareness" will be used with this limited connotation.

Specifically, the "awareness" considered in
 the paper refers to the state of being conscious of perceiving an object in an act
  of object perception by a single person, or in the inter-person dialog as described
   above. For this reason, we call our model Consciousness Networks (CONNs).

In our model, the awareness of perceiving
 an object by a single person and in inter-person dialogue is represented as the fixed
  point functionality of operators in metric spaces. These operators represent
   person-to-object and person-to-person communication, respectively.

In the paper, we introduce techniques for analyzing and interpreting visual information in a social context. By integrating person-to-person communication cues and object perception capabilities, our approach aims to model social perception of objects.

Furthermore, the model can be applied to   computer vision classification tasks. By leveraging our observed-to-seen model, we have created an image classifier that exhibits high visualizability and performs well with small training datasets.

The contributions of this paper are as follows:
\begin{itemize}
    \item Up to our understanding our research is the first attempt to model image
visual perception jointly with the  derived inter-person  communication.
    \item
We model human perception using a sequence of "observed" and "seen"
personalized images. This provides  interpretability of the states of
the modeling network.
    \item
Through the paper we consider communication
either between person
or internally "in the person".
However, "person"
should be interpreted
in general sense,
meaning to be any sort of system
including computer system.
On the same note,
the model described in this paper
supports both internal and external communication
through a unified equations. 
The details
for implementing in different
systems (for example, internal representation
of object in a person, or modality of communication between
two persons) can differ.
    \item
We model the "observed-to-seen" operation as composition of encoder and
decoder operations of convolutional autoencoders. This allows to represent
an act of the object perception as a sequence of iterations converging to
attractor.
    \item
Up to our understanding we introduce the notion of bipartite orbits in
dynamics systems.
    \item
We develop an attractor
based  classifier for classical computer vision classification tasks. The classifier is visualizable and its 
stochastic version
outperforms a standard baseline classifier when dealing with limited training datasets.
   \item
Our model describes several semiotic phenomena of person-to-object and person-to-person communication.
\end{itemize}

The glossary of terms used in this paper is provided in \citet{SI}~A.

\section{Related Work}
\label{sec:Related_Work}

Interestingly, there is limited research on
simulating person-to-person communication.

The Osgood-Schramm model (\citet{Osgood_Schramm_model})
is a cyclic encoder-decoder framework for human 
interactions. There, the encoder outputs are the 
transmitted images.
In contrast, our model
takes a different approach by  employing encoder
outputs as an internal representation of an
individual's input perception.

A few years ago, Google introduced the DeepDream network (\citet{DeepDream}), which bears some resemblance to our work in terms of the notions of the observed and seen images and the cycle between them. In their work, the images representing what a person sees in the input image are treated as input to the network. In our work, on the other hand, we simulate the seen images as the network's output. This fundamental difference accounts for the fact that while delving deep into DeepDream often produces unrealistic "dream" images, our approach tends to generate more realistic "normal" images.

Large language models (LLMs) are central to AI research, with much work addressing their challenges, including hallucinations (\citet{liu2024,tonmoy2024}). Our approach aims to mitigate this issue by aligning outputs with predefined internal knowledge. This resembles using an internal Retrieval-Augmented Generation (RAG) (\citet{RAG}) method, restricting results to domain-specific knowledge and ensuring closer alignment with the intended field.

Many works deal with interpreting and understanding
deep neural networks (for example \citet{Montavon2018-nk}).
In contrast to  methods where we interpret
what a given network "sees"
(for example \citet{Gat2022,xu2018interpreting}),
we explore a different approach. Specifically,
we equip a network with
certain functionality of  perceptual inference.
This also  allows visualization of the obtained network.

Our network is implemented using the encoding-decoding operations of an autoencoder. We rely on the work of \citet{Belkin}, where it has been empirically shown that for overparameterized autoencoders, such sequences converge to attractors. Another basic finding of this work is that an overparameterized autoencoder stores input examples as attractors. We make use of these results when designing our attractor-based classifier (\sectionref{sec:SNRA_classifier}).

The key difference between our classifier and approaches that employ  denoising autoencoders (for example\citet{adv_autoenc}) lies in the iterative nature of the encoding and decoding operations, which leads to convergence to attractors.

In \citet{paper_17}, attractors were applied to classification in the field of speech recognition. In \citet{B2_paper}, the attractor-based classifier is employed to estimate the normalized entropy [34] of the probability vector. This approach is used to detect novel sceneries, such as out-of-distribution or anomaly samples, rather than performing the classification task.

In \citet{B2_paper}, a sample is represented as a series of convergence points in the latent space, obtained during recursive autoencoder operations using Monte Carlo (MC) dropout. Our classifier comes in two forms: vanilla and stochastic, with the latter built upon the former. The vanilla version represents an input sample as a single convergent point in the image space, resulting from encoder and decoder operations.
In our stochastic classifier, an input sample is represented by a set of attractors that are in close proximity to the sample, thereby augmenting the informativeness of the representation. The construction of the attractor sets involves randomized iterative alternations of the samples in the image domain.

Additionally, a meaningful distinction arises between our representation and the dropout approach of \citet{B2_paper}. The dropout mechanism generates outputs that represent known samples with similar representations and unknown ones with dissimilar representations. However, our stochastic
classifier typically assigns different attractors to all examples, including the training ones, in this sense ignoring the novelty of the samples.

Technically, our representation may resemble the SIFT approach (\citet{SIFT}).
There, instead of considering a specific pixel, SIFT considers the neighboring area
of the pixel, known as the "vicinity",
where the histogram
representations for the predefined
gradient directions are
calculated. In our approach, the "histogram bins" are generally associated with the training examples, whereas the constructed "histogram" depends on the convergence of the stochastic algorithm.

Our work has some common ground with RNN networks. In both, an internal state is preserved and is used and updated when new inputs are being processed. In this light, our model can be examined as a few RNN networks, each representing one person, that communicate with each other. In classical RNN networks, the internal state can receive any value (with some implementation detail limitations). On the other hand, our model attempt to preserve its internal model within a certain "pre-defined" set.

In the subsequent sections, we will provide a detailed description of our model and discuss 
how it represents the semiotics of 
human perception and communication.

\section{Modeling Person-to-Person  Communication using Semiotics Networks}
\label{sec:Modeling}

In this section, we introduce the Conscious Neural Network (CONN) for modeling communication between persons perceiving visual images.
We will describe a two-person communication model, the model may be easily generalized to 
a multiperson case.

\begin{figure}[H]
\centering
{\fbox{\includegraphics[width=0.75\linewidth]{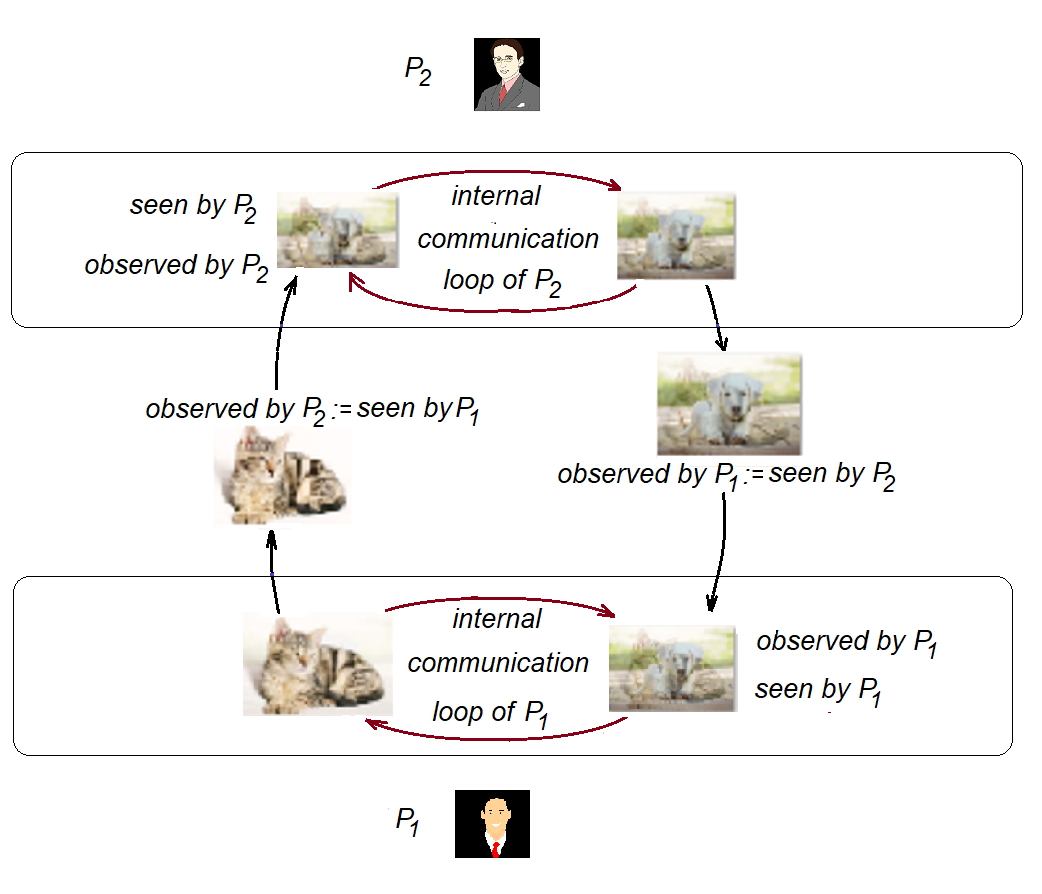}}}    
    \caption{
Person-to-person CONN.
The internal communication loops
    associated with the 
    persons
    are comprised of the observed-to-seen transformations and are denoted by rounded rectangles. 
The persons interchange their seen images, resulting in the internal communication loops,
using the external communication loop 
(denoted by black arrows).
    The flowchart of the implementation of the CONN using autoencoder operations is shown in
\figureref{fig:scheme_full_b}    
     \protect\footnotemark
    }
\label{fig:scheme_full_a}
\end{figure}

Consider two persons, $P_1$ and $P_2$ (refer to \figureref{fig:scheme_full_a}). The first person consistently tends to see cats in all input images, and the second person tends to see dogs. Specifically, the first person performs a sequence of iterations trying to see "catness" in the observed image: at the first iteration it converts an observed input image $Im$ to an image with some features of the cat, at the second iteration  converts the obtained image to a new image with more features of the cat etc. 
This process continues, gradually incorporating more cat features.
At every iteration the currently observed image is converted to the "seen" which becomes  the observed for the next iteration. After a finite number of iterations the person sends the resulted image to the second person and waits its response. Similarly, person $P_2$ tends to see "dogness" in the perceived images: it performs a sequence of iterations with more features of the dog appearing  at each iteration. 
The resulting image is then sent to $P_1$,
while $P_2$ begins waiting for a response from
$P_1$. The whole cycle then continues.
We will refer to the flow of data sent from person to person in  CONN as the external communication loop.

The process is expressed as:
\begin{equation} 
\label{eq:1}
\begin{split}
&Im = obs_{1,1} \xrightarrow{O2S_{P_1}} seen_{1,1} =
obs_{2,1} \xrightarrow{O2S_{P_1}} 
\cdots 
\xrightarrow{O2S_{P_1}} 
\\
&seen_{nsteps_1,1}= obs_{1,2} \xrightarrow{O2S_{P_2}} seen_{1,2} =
obs_{2,2} \xrightarrow{O2S_{P_2}} 
\cdots 
\xrightarrow{O2S_{P_2}} 
\\
&seen_{nsteps_2,2}= obs_{1,3} \xrightarrow{O2S_{P_1}} seen_{1,3} =
obs_{2,3} \xrightarrow{O2S_{P_1}} 
\cdots 
\xrightarrow{O2S_{P_1}} 
\\
&seen_{nsteps_1,3}= obs_{1,4} \xrightarrow{O2S_{P_1}} seen_{1,4} =
obs_{2,4} \xrightarrow{O2S_{P_1}} 
\cdots 
\xrightarrow{O2S_{P_1}} 
\\
&\cdots
\\
&seen_{nsteps_{od(iter)},iter}= obs_{1,iter+1} \xrightarrow{O2S_{P_1}} seen_{1,iter+1} =
obs_{2,iter+1} \xrightarrow{O2S_{P_1}} 
\\
&\cdots
\;
,   
\hspace{25em}
\end{split}
\end{equation}
where the seen images obtained at each iteration, except the last, in every internal communication loop become the observed images for the following iteration of the loop. The seen images from the last iteration become the initial observed images in the subsequent iteration of the external communication loop.

Here, $nsteps_i$ denotes the internal communication loop length, 
$O2S_{P_i}$ denotes the observed-to-seen transformation, 
both for the $i$-th person, $iter$ denotes the index of the 
external communication loop,
and
  \begin{equation}
  \label{eq:i_iter}
    od(iter)=
    \begin{cases}
      1, & \text{if } iter \text{ is odd} \\
      2, & \text{if } iter \text{ is even}.
    \end{cases}
  \end{equation}

\begin{figure}[H]
\centering
{\fbox{\includegraphics[width=0.75
\linewidth]{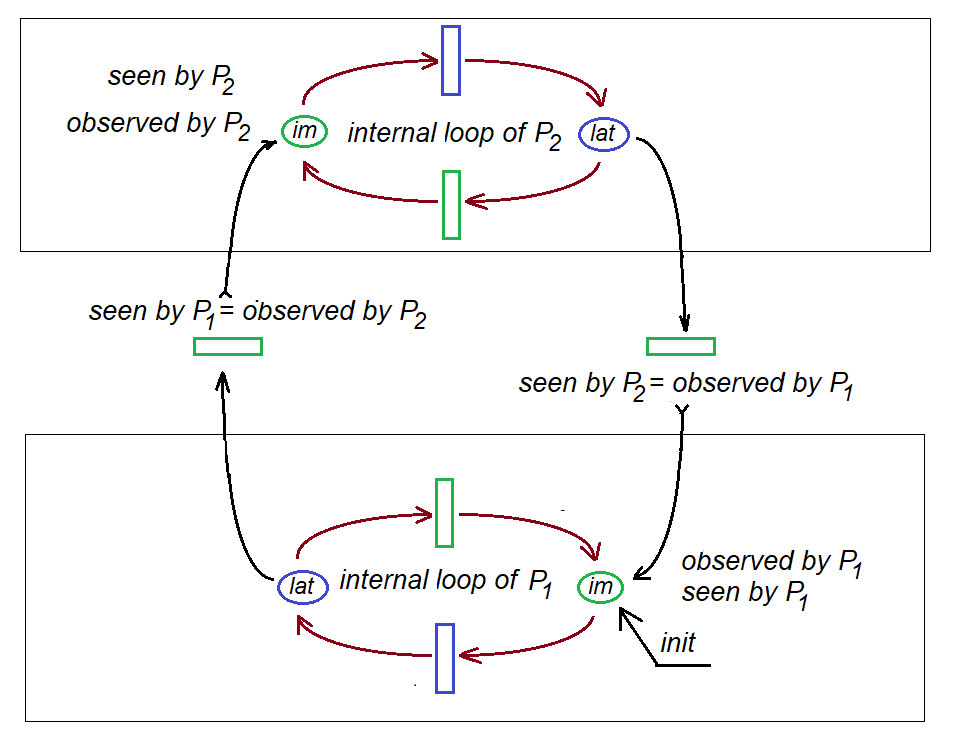}}}
\caption{The  figure shows an
implementation of the person-to-person CONN from \figureref{fig:scheme_full_a}, with the observed-to-seen transformations implemented as the composition of encoder (shown as blue rectangles) and decoder (shown as green rectangles) operations. This implementation is described in 
\algorithmref{alg:CONN}. The external communication loop (denoted by black arrows) is represented by step \ref{nitm:3}, while the internal communication loops (denoted by rounded rectangles) are represented by step \ref{nitm:3.15} of the algorithm\protect\footnotemark
}
\label{fig:scheme_full_b}
\end{figure}

In general, the representations in
\equationref{eq:1}
do not have to be of image modality; they may be, for example, textual descriptions.
We will refer to the modalities of these
representations as raw modalities. Meanwhile, we confine ourselves to the case where 
these representations are themselves the images.\footnote{See the footnote to \autoref{fig:scheme_full_b}.}

The internal communication loops associated with the persons
may be considered as
the PAS (person aligned stream)
loops (\citet{AE_paper}).

CONNs can be implemented in various ways. One approach is to implement the observed-to-seen transformations, which are an essential part of CONN, using convolutional autoencoders. The $enc$ and $dec$ operations of the autoencoders perform transformations from the image space to a latent space and back:
\begin{equation} 
\label{eq:2}
obs\rightarrow seen:
\,\,
obs \xrightarrow{enc}
 enc(obs)
 \xrightarrow{dec}
 seen  = dec(enc(obs))
 .
\end{equation}

Note that both "observed" and "seen" representations here are of the raw modality and not in the latent space. The transition from "observed" to "seen" is through the latent-based autoencoder representations.
Using these operations, the 
CONN  
is implemented, as illustrated in
\figureref{fig:scheme_full_b}.
Its functionality is described by
\algorithmref{alg:CONN}.

\footnotetext{The figures in this paper are best viewed in color.}


\footnotetext{In \citet{SI}~B, we present the flowchart of the CONN operating with several raw modalities. In \figureref{fig:scheme_full_b}, the two blocks that perform transformations from latent to raw representations in the external communication loop appear redundant compared to those located within the internal communication loops. This redundancy arises from the construction of the flowchart in \figureref{fig:scheme_full_b} as a specific instance of the general scheme presented in \citet{SI}~B.}

We employ the autoencoder-based implementation in our modeling of object perception (\sectionref{sec:pers_one_obj:attractors}) and in the construction of the CONN-based classifiers (\sectionref{sec:SNRA_classifier}). Another implementation of CONN involves a more general mathematical representation of observed-to-seen transformations as continuous functions in complete metric spaces. We use this  representation in \sectionref{sec:pers_two_obj}, where the person-to-person communication is considered. Also, in \citet{SI}~F, we examine a simplified computer implementation of  CONN not based on autoencoders.

\begin{algorithm}[htbp]
  \centering
  \caption{A conscious  neural network for  communication between two persons.
The network is comprised
of autoencoders $A_{P_1}$ and
$A_{P_2}$ associates with
persons $P_1$ and $P_2$
	\newline
	{\bf Input}:
	An image $Im_1$ which is related to person
	$P_1$
	\newline
	{\bf Output}:  A sequence of interchange images
	$Im_1, Im_2, \dots Im_k, \dots$
	}
	{
    \begin{enumerate}
        \item
        \label{nitm:1}
            Set $iter=1$;         set $person\_id=1$

        \item
        \label{nitm:2.5}
        Initialize the output queue to an empty list
        
        \item
        \label{nitm:3}
        While $iter \leq  n_{iters}$ do:
        \begin{enumerate}
        \item
        \label{nitm:3.1}
        Use $person\_id$ parameters ($nsteps_{person\_id}$)
        \item
        \label{nitm:3.15}
        Perform
        $nsteps_{person\_id}$
        encoding/decoding iterations
        (\equationref{eq:2})
        of the autoencoder associated
        with person $person\_id$ on
         $Im_{iter}$
        to receive the image representation
        (current $Im$).
        \label{nitm:3.2}

        \item
        \label{nitm:3.3}
        Decode the previous
        encoding result
        $lat$
        ($\text{current } Im=dec(lat)$) to
        receive $Im_{iter+1}$
        ($Im_{iter+1} = \text{current } Im$
        after this operation)

        \item
        Increase $iter$ by 1 and
        change $person\_id$ to other $person\_id$
        \label{nitm:3.4}

        \item
        Send $Im_{iter}$ to the updated person
        and add to the output queue
        \label{nitm:3.5}
        \end{enumerate}
        
        \item
        \label{nitm:4}
        
        Return the output queue

    \end{enumerate}
    }
    \label{alg:CONN}
\end{algorithm}

\section{
Use of CONNs for Modeling  Object
 Perception and Inter-Personal Communication}
\label{sec:conv_autoenc}

In this section, we will study how CONNs model the perception of an object by a person, as well as the perception of an object in a dialogue between persons.

In \sectionref{sec:pers_one_obj:attractors}
we delve into the object
perception by one person.
Additionally, we consider several well known 
attractor 
related notions 
(\citet{Belkin})
and give them perception-related interpretation.
These will serve as the basis for defining the  "perceptualization of a classifier" in Section \ref{sec:SNRA_classifier}.

In \sectionref{sec:pers_two_obj}
we will consider person-to-person communication and introduce bipartite orbits, which may be regarded as the "fixed points"
of interpersonal communication.

The material of 
this section
will allow us
to analyze, 
in \sectionref{sec:rikuz:semiotics},
how CONNs 
represent the semiotics
of
object perception 
and person-to-person
communication.

\subsection{Perception of an Object by One Person: Attractors}
\label{sec:pers_one_obj:attractors}

Below, we will model the interaction between a person and an object as a specific case of CONN modeling, which was introduced for person-to-person communication.
We will rely on the autoencoder-based implementation
of the observed-to-seen
transformation
(\sectionref{sec:Modeling}).

In \citet{SI}~C, we show that, in the CONN model, the perception of an object by a person can be considered a particular case of person-to-person communication.
In 
this scenario, each internal communication cycle of 
images associated with a person begins with the same 
observed image. Assuming an autoencoder-based 
implementation of the observed-to-seen transformation 
and using the notation from \equationref{eq:2}, we can 
write this cycle as:
\begin{equation*}
Im \rightarrow dec(enc(Im)) \rightarrow \dots \rightarrow 
[dec(enc)]^{nsteps}(Im),
\end{equation*}
where $[dec(enc)]^{nsteps}$ denotes $nsteps$ 
compositions of the $dec(enc)$ function.

The process takes an input image $Im$, encodes it into the latent space using $enc(Im)$, and then decodes it back to the image space. This encoding/decoding procedure is repeated $nsteps$ times, resulting in an image representation in the original modality. 
It has been empirically shown 
that for overparameterized autoencoders,
as $nsteps$ approaches infinity,
such sequences converge to attractors
(\citet{Belkin}). We have observed a similar phenomenon in autoencoders which are 
not necessarily overparameterized. Additionally, we observed convergence to cycles. See \sectionref{sec:ExpResults} 
and 
\citet{SI}~K for details.

For an input image $Im$ 
we call the final representation of $Im$ in the image space
\begin{equation} 
\label{eq:3}
\widehat{F}(Im) = \lim_{n\to\infty} [dec(enc)]^n(Im)
,
\end{equation}
if such limit exist,
the {\em percept image} of $Im$.

For the percept image there holds the fixed point property:
\begin{equation} 
\label{eq:fixed}
dec(enc)( \widehat{F}(Im) ) = \widehat{F}(Im).
\end{equation}
The equation indicates that applying the encoding and decoding operations to the percept image results in the same image.

\subsection{Person-to-Person Communication: Bipartite Orbits}
\label{sec:pers_two_obj}

Below, we will delve into inter-person communication and study the asymptotic characteristics of the image sequence exchanged within our CONN model (\sectionref{sec:Modeling}). These properties will play a key role in our exploration of interpersonal communication in \sectionref{sec:rikuz:semiotics:person to person}.

What periodicity is being referred to?
One may assume that the sequence of
the images "perceived" by the person
converges to "attractors".
For example, for a a "dog-like"
person, the sequence converges to
a dog image.
However, when more than one person is involved, this
assumption may not  hold
anymore for the whole sequence
of intertrasmitted images, because
there is no guarantee
that both persons
share the same "attractors".
For example, if one is a "dog-like"
person (i.e., the "attractors"
are comprised of dogs images only) and the other is a "cat-like" person, then a joint "attractor" is of a low choice.
A "dog-like" person is unllikely
"to see" a cat image
and vise versa for
the "cat-like" person.

We will identify two types of periodicity in the sequence of transmitted images between the persons. Both types are observed when the external communication parameter of \algorithmref{alg:CONN} (the number of information exchanges between the persons) tends to infinity. The difference lies in whether the internal communication parameters (the numbers of observed/seen transformations 
as expressed by $nsteps_1$, 
and $nsteps_2$ 
in \equationref{eq:1}) also tend to infinity. 
These two types of periodicity are studied in 
Sections~\ref{sec:subsub:Bipartite Orbit 1} and \ref{sec:subsub:Bipartite Orbit 2}
respectively.

\subsubsection{Attractor-Related Notions for Person-to-Person Communication}\label{sec:subsub:Attractor-Related Notions for Person-to-Person Communication}

In \sectionref{sec:pers_two_obj}, we consider CONNs represented as operations in a complete metric space, which are not necessarily implemented via encoding/decoding operations. For such CONNs, we define the notions from \sectionref{sec:pers_one_obj:attractors} in a more general form.

The definitions of attractors, fixed points, and basins, as provided for Euclidean space by \citet{Belkin}, are applicable to any complete
metric space $X$, and we will adopt them in the following.

Let $F_{P}$ be a continuous function $X \rightarrow X$. For $x \in X$, if the limit
\begin{equation}
\label{eq:F_widehat}
\widehat{F}(x) = \lim_{n\to\infty} [F_{P}]^n(x)
\end{equation}
exists, we refer to this mapping as the "perceptualization operator," and the limit value as the "percept image" (see \equationref{eq:3}).
If, for $x \in X$, the fixed-point equation
\begin{equation}
\label{eq:fixed_points_F_P}
F_{P}(x) = x,
\end{equation}
holds, we refer to this as the "awareness property."
It can be easily shown that if
$x$ is a percept image with respect to $F_P$, it satisfies the awareness property. 
An explanation of these terms will be provided in \sectionref{sec:rikuz:semiotics}.

\subsubsection{Bipartite Orbits of the First Type}
\label{sec:subsub:Bipartite Orbit 1}

In \sectionref{sec:pers_one_obj:attractors}, the fixed points of autoencoders' mappings were considered as modeling the perception of an object by one person. Interestingly, when human {\em communication} is simulated, an asymptotically periodic sequence of inter-person transmitted images has been identified. We will study this property in the this
section.

Formally, let $F_{P_1}$ and $F_{P_2}$ be
two continuous functions $X \rightarrow X$, where
$X$ is a complete metric space with 
distance function $d$,
and $Im\in X$
be an initial point 
("an image").
Consider
a sequence
$W(Im)$
starting with
$Im$
and consisting of subsequent
application
of $nsteps_1$ times of $F_{P_1}$,
then $nsteps_2$ times of $F_{P_2}$,
then $nsteps_1$ times of $F_{P_1}$
etc.\footnote{The
representation of $W$
in terms of encoding and decoding operations is considered in \citet{SI}~D.}
Here, $nsteps_1$ and $nsteps_2$ are given numbers representing the "internal" number of steps for convergence, as in \algorithmref{alg:CONN}.

If we denote
\begin{equation}
\label{eq:T}
\begin{array}{l}
S_{1,P_{1}} =
Im, \, F_{P_1}(Im), \dots , [F_{P_1}]^{nsteps_1}(Im);
\\[0.3em]
\;\;T_{1,P_{1}} = [F_{P_1}]^{nsteps_1}(Im);
\\[0.45em]
S_{2,P_{2}} =
T_{1,P_{1}}, \,   F_{P_2}(T_{1,P_{1}}), \dots ,
[F_{P_2}]^{nsteps_2}  (T_{1,P_{1}})  ;
\\[0.3em]
\;\;T_{2,P_{2}} = [F_{P_2}]^{nsteps_2}(T_{1,P_{1}});
\\[0.45em]
S_{3,P_{1}} =
T_{2,P_{2}}, \,
F_{P_1}(T_{2,P_{2}}),
\dots , [F_{P_1}]^{nsteps_1}(T_{2,P_{2}});
\\[0.3em]
\;\;T_{3,P_{1}} = [F_{P_1}]^{nsteps_1}(T_{2,P_{2}});
\;
\\
\dots \; ,
\\
\end{array}
\end{equation}
then $W(Im)$ can be expressed as the concatenation:
\begin{equation}
\label{eq:W1}
\begin{aligned}
	W(Im)=
	\operatorname{concat}(
	S_{1,P_1}, \, S_{2,P_2}, \, S_{3,P_1},
	\dots
	,
	S_{iter,P_{od(iter )}},
	 \dots)
	 \;.
\end{aligned}	 
\end{equation}

Here,
$iter $ is the "external"
counter of communication,
similarly to
\algorithmref{alg:CONN},
and 
$od$ is defined  in 
\equationref{eq:i_iter}.

Now focus on the elements
$T_{1,P_{1}}$, $T_{2,P_{2}}$,
$T_{3,P_{1}} \ldots$
in  \equationref{eq:T}.
They
represent the final image of each person
at the $iter$-th iteration, which is later sent to the other person.
They
comprise a sub-sequence
$U$ of $W$:
\begin{equation} \label{eq:U0}
\begin{split}
&U(Im,nsteps_1,nsteps_2)=
Im
\xrightarrow{  [F_{P_1}]^{nsteps_1} }
T_{1,P_1}
\xrightarrow{  [F_{P_2}]^{nsteps_2} }
T_{2,P_2} \xrightarrow{  [F_{P_1}]^{nsteps_1} }
\dots
\\
&
\;\;\;\;\;\;\;\;
\xrightarrow{  [F_{P_{od(iter)}}]^{nsteps_{od(iter)}} }
T_{iter,P_{od(iter)}}
\xrightarrow{  [F_{P_{od(iter+1)}}]^{nsteps_{od(iter+1)}} }
\dots
\;.
\end{split}
\end{equation}

Denote
\begin{equation}
\label{eq:F1F2}
F_1 = [F_{P_1}]^{nsteps_1}, 
\; \text{and} \; F_2 = [F_{P_2}]^{nsteps_2}.
\end{equation}

\begin{figure}[htbp]
\centering
    {\fbox{
    {\includegraphics[width=0.95\linewidth]{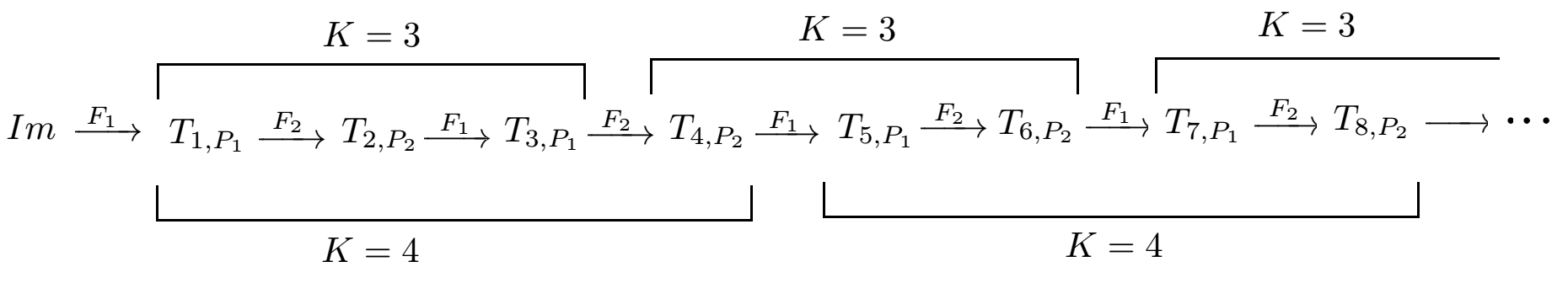}}}}
    \caption{
    Partitioning
    $U$ by segments
    of $K=3$ and $K=4$}
    \label{fig:splashing_by_K}
  \end{figure}

For any $m>0$, $K>0$ one may
partition $1+(m+1)*K$ members of the sequence by
$Im$, followed by the matrix of
$(m+1)$ rows and $K$ columns
(see \citet{SI}~E).
Similarly,
for any $K>0$
we may partition the whole sequence by
$Im$, followed by subsequent segments
of length $K$, see
\figureref{fig:splashing_by_K}.
The bipartite convergence
of the first type 
will be defined
by way of
columns of
the infinite matrix whose
lines
are the $K$-length
segments of such partitioning.

Specifically, for any $K>0$ the sequence may be written as follows:
\begin{equation*}
\begin{array}{@{}cccccccc@{}}
\multicolumn{4}{l}{U(Im,nsteps_1,nsteps_2)= Im \xrightarrow{F_1} } &&&&
\\[0.3em]
 T_{1+0\cdot K,P_1} &\xrightarrow{F_2} & T_{1+0\cdot K+1,P_2} &
            \xrightarrow{F_1}  & \cdots            &\xrightarrow{F_2}       & T_{1+0\cdot K +K-1,P_2} &\xrightarrow{F_1}\\

 T_{1+1\cdot K,P_1} &\xrightarrow{F_2} & T_{1+1\cdot K +1,P_2} &
            \xrightarrow{F_1}  & \cdots            &\xrightarrow{F_2}       & T_{1+1 \cdot K + K-1),P_2} &\xrightarrow{F_1}\\

\dots &&&&&&& \\

 T_{1+m\cdot K,P_1} &\xrightarrow{F_2} & T_{1+m\cdot K+1,P_2} &
            \xrightarrow{F_1}  & \cdots            &\xrightarrow{F_2}       & T_{1+ m \cdot K + K -1,P_2} &\xrightarrow{F_1}\\

\dots \;\;. &&&&&&&   \\
\end{array}
\end{equation*}

For
$j=0,\dots, K-1$
the $j$'s
column of this representation
is
written as
\begin{equation}
\label{eq:colomn}
  C_{j} (nsteps_1,nsteps_2) =
  \begin{pmatrix}
        T_{1+0\cdot K+j,P_{od(j+1)}}
        \\[0.45em]
        T_{1+1\cdot K +j,P_{od(j+1)}}
        \\
        \dots \\
        T_{1+m\cdot K+j,P_{od(j+1)}} \\
        \dots \\
  \end{pmatrix}
  ,
\end{equation}
where $od()$ is defined in \equationref{eq:i_iter}
and the column, treated as the sequence,
is indexed by $m$.

\begin{definition}[Bipartite Convergence of the First Type to Orbit]
\label{def_bipartite_orbit_1}
A sequence
\\
$U(Im,nsteps_1,nsteps_2)$
(\equationref{eq:U0})
is a bipartite
convergent sequence
of the first type
converging to the orbit
$(b_j \;|\; j=0,\dots, K-1)$, $K >0$,
if all $b_j$ are different and for every
$j=0,\dots, K-1$,
the column
$C_j(nsteps_1,nsteps_2)$
as a sequence
converges to $b_j$.
\end{definition}
We will refer to the orbits in this definition as the bipartite 
orbits of the first type.

The next remark is obvious.
\begin{remark}
 \label{remark_assympt_periodic_1}
A sequence $U = U(Im,nsteps_1,nsteps_2)$
 is a bipartite
sequence
of the first type if and only if
it is
an asymptotically $K$-periodic
sequence (\citet{asymptotically_periodic}).
In this case the sequence comprising the
period of $U$ is the bipartite orbit of $U$.
\end{remark}

Under what conditions
a sequence $U(Im,nsteps_1,nsteps_2)$
 is a bipartite
convergent sequence
of the first type?
Although the existence of such orbits
was not formally proven,
in our experiments
\sectionref{sec:Bipartite Orbits}
with the autoencoders' generated images,
we  observed convergence to
the bipartite orbits
for every initial $Im$, $nsteps_1$, and 
$nsteps_2$.
Note that the autoencoders were not overparameterized in these experiments.
In addition, we developed a simplified computational model
for simulating inter-person communication (\citet{SI}~F).
 The running of the model consistently demonstrates convergence
 to what can be referred to as the first type orbit of the
 simplified model.

Given a bipartite 
sequence $U = U(Im,nsteps_1,nsteps_2)$
with period length $K$,
we may denote
\begin{equation}
\label{eq:G1G2}
\begin{gathered}
G_1 = [\,F_1 \cdot F_2\,]^{K/2},
 \\
G_2 = [\,F_2 \cdot F_1\,]^{K/2},
\end{gathered}
\end{equation}
where
$F_1$, $F_2$ are defined in
\equationref{eq:F1F2}.
Then we may write the sequence
as:
\begin{equation}
\label{eq:matrix}
\begin{array}{@{}cccccccc@{}}
\multicolumn{4}{l}{U(Im,nsteps_1,nsteps_2)= Im \xrightarrow{F_1} } &&&&
\\
T_{1+0\cdot K,P_1} &\xrightarrow{F_2} & T_{1+0\cdot K+1,P_2} &
        \xrightarrow{F_1}  & \dots  &\xrightarrow{F_2}  & T_{1+0 \cdot K +K-1 ,P_2}  &
        \xrightarrow{F_1}
\vspace{.2cm}
\\
G_1 \downarrow & & G_2 \downarrow &&&& G_2 \downarrow & 
\\
T_{1+1\cdot K,P_1} &\xrightarrow{F_2} & T_{1+1\cdot K+1,P_2} &
        \xrightarrow{F_1}  & \dots &\xrightarrow{F_2} & T_{1+1 \cdot K +K-1,P_2}  &
        \xrightarrow{F_1}
\vspace{.2cm}
\\
G_1 \downarrow & & G_2 \downarrow &&&& G_2 \downarrow & 
\\
\dots & & \dots &&&& \dots & 
\\
G_1 \downarrow & & G_2 \downarrow &&&& G_2 \downarrow & 
\\
T_{1+m\cdot K,P_1} &\xrightarrow{F_2} & T_{1+m\cdot K+1,P_2} &
        \xrightarrow{F_1}  & \dots &\xrightarrow{F_2} & T_{1 + m \cdot K +K-1,P_2} &
        \xrightarrow{F_1}
\vspace{.2cm}
\\
G_1 \downarrow & & G_2 \downarrow &&&& G_2 \downarrow & 
\\
\dots & & \dots &&&& \dots & 
\\
b_0            & & b_1            &&&& b_{K-1}        & 
\\
.
\end{array}
\end{equation}
In this notation, the next lemma holds.

\begin{lemma}
\label{lemma_att1}
The elements of a bipartite orbit
$(b_0,b_1,\dots,b_{K-1})$
of the first type
satisfy the properties:

    \begin{enumerate}

        \item
        \label{att1:1}
They form a loop
with respect to alternating
$F_1$, $F_2$ operations:
\begin{equation}
\label{eq:att1}
b_0
\xrightarrow{  F_2 }
b_1
\xrightarrow{  F_1 }
\dots
\xrightarrow{  F_2 }
b_{K-1}
\xrightarrow{  F_1 }
b_0
,
\end{equation}

        \item
        \label{att1:2}

These elements
are also
alternating fixed points of functions
$G_1$, $G_2$:

\begin{equation}
\label{eq:G1G2forFixed}
\begin{aligned}
&G_1(b_h) = b_h, \text{ for even } h, \\
&G_2(b_h) = b_h, \text{ for odd } h.
\end{aligned}
.
\end{equation}

\end{enumerate}

\begin{proof}
See \citet{SI}~G. 
\end{proof}
\end{lemma}

It should be noted that the elements
comprising the bipartite orbits of the first
type are not necessarily the fixed points
of $F_1$ or $F_2$.
This is due to "non-deep" character
of the internal communication 
(the $nsteps_1, nsteps_2$ are not tending to infinity).

We consider semiotic  interpretation 
of operators $G_1$ and $G_2$
in \sectionref{sec:rikuz:semiotics:person to person}.

\subsubsection{Bipartite Orbits of the Second Type}
\label{sec:subsub:Bipartite Orbit 2}

In this section, we will 
continue exploring the periodic properties of the 
sequences 
of inter-person transmitted images. These 
properties
will be further interpreted in \sectionref{sec:rikuz:semiotics:person to person}.

The bipartite convergence studied in  \sectionref{sec:subsub:Bipartite Orbit 1}, describes the behavior of the sequences
as the parameter $iter$ in
 \equationref{eq:T}  tends to infinity. 
The convergence considered in this 
section describes the  behavior of the sequences as the "internal" persons' parameters, $nsteps_1$ and $nsteps_2$  in \equationref{eq:U0}, also tend towards infinity. 
These parameters represent the steps involved in converging to the fixed points of $F_{P_1}$ and $F_{P_2}$. These points may be treated as the person-dependent representations, independent of another person.

 As before, our assumption regarding $F_1$ and $F_2$ is that they are continuous functions operating in complete metric spaces.

\begin{figure}[h!]
\centering
    {\includegraphics[width=0.6\linewidth]{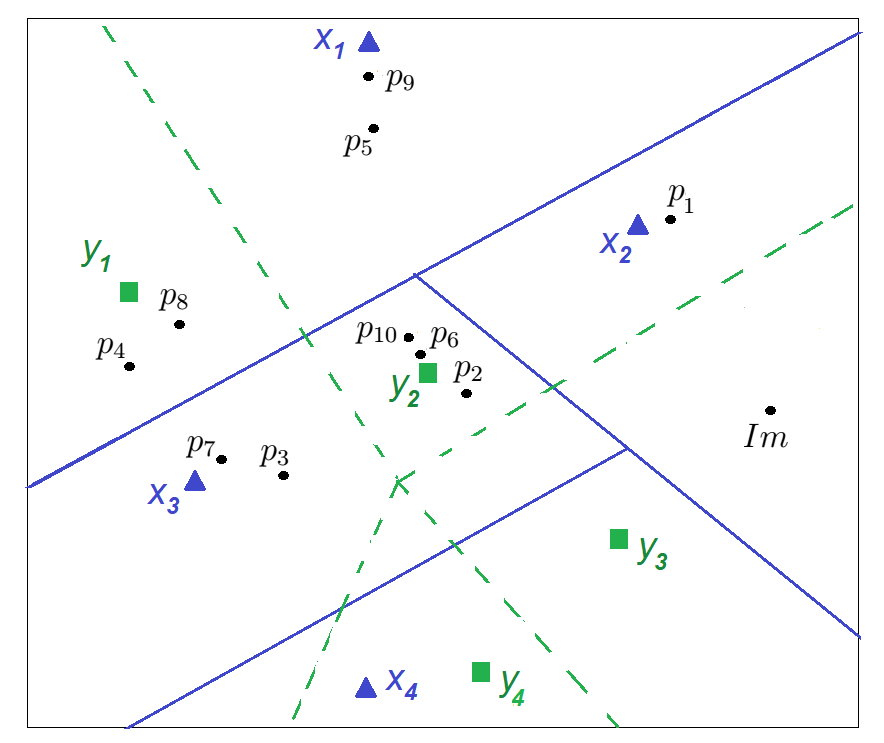}}
\caption{
Illustration of the bipartite convergence of the second type, claimed in \theoremref{theorem_L_A}.
        The space $X$ contains 4 
        basins for $F_{P_1}$ with 
        attractors $x_1, \ldots ,x_4$ depicted 
        as blue triangles.
        The borders between the basins are denoted by blue lines.
         Analogously, 
        $X$ contains  3 
        basins for $F_{P_2}$ with 
        attractors $y_1$, $y_2$,
        $y_3$, and $y_4$ depicted 
        as green rectangles. 
        The borders between the basins are denoted by dashed green lines.
        Alternating mappings to the
        attractors
        of $F_{P_1}$ and
        $F_{P_2}$,
        starting with $Im$,        
        yield sequence
        $H=Im\rightarrow x_2
        \rightarrow y_2
        \rightarrow x_3        
        \rightarrow y_1        
        \rightarrow x_1        
        \rightarrow y_2
        \rightarrow x_3        
        \dots$,\, terminated by  cycle
        $(x_3,y_1,x_1,y_2)$.
For a selected proximity, 10 sequential elements 
$p_1, p_2,\dots, p_{10}$
from a subsequence $(p_i)$ of sequence $U(Im,nsteps_1,nsteps_2)$ (\equationref{eq:U0}) are shown. 
The parameters $nsteps_1$ and  $nsteps_2$, and $iter$, the position of  $p_1$
in $U$, are chosen sufficiently large, so that the elements $p_i$ fall within the predefined proximity to the respective attractors:
$p_1$ is close to $x_2$,
$p_2$ to $y_2$,
$p_3$ to $x_3$,
\dots,
$p_{10}$ to $y_2$, etc.}
\label{fig:theorem_L_A}
\end{figure}

Consider the following example. In \figureref{fig:theorem_L_A}, the image space 
$X$ is depicted, partitioned by basins corresponding to the finite sets of attractors of $F_{P_1}$ and $F_{P_2}$. Let us examine the sequence $H$ consisting of the elements shown in the picture. The sequence starts with the images $Im$. Each subsequent element $y$ of the sequence is defined by assigning it the attractor of the basin to which the previous element $x$ belongs.
These basins correspond to alternating functions $F_{P_1}$ and $F_{P_2}$: $Im$ converges to $x_2$, the attractor of $F_{P_1}$. Further, $x_2$ converges to $y_2$, the attractor of $F_{P_2}$. Then $y_2$ converges to $x_3$, the attractor of $F_{P_1}$, etc. Since the number of attractors is finite, starting from a certain index, the sequence becomes cyclic: $H = Im \rightarrow x_2 \rightarrow y_2 \rightarrow x_3 \rightarrow y_1 \rightarrow x_1 \rightarrow y_2 \rightarrow x_3 \dots \,$.

The bipartite convergence of sequences $U$ studied below describes their behavior as they become infinitesimally close to cycles of elements, 
like $(y_2,x_3,y_1,x_1)$, in \figureref{fig:theorem_L_A},
with the values of $nsteps_1$, $nsteps_2$, and $iter$ tending to infinity.

\begin{definition}[Bipartite Convergence of the Second Type to Orbit]
\label{def_bipartite_orbit_2}
A sequence
\\
$U(Im,nsteps_1,nsteps_2)$
(\equationref{eq:U0})
is a bipartite
convergent sequence
of the second type
converging to the orbit
$(b_j \;|\; j=0,\dots, K-1)$, $K>0$,
if all $b_j$ are different and for every
$j=0,\dots, K-1$
column
$C_j(nsteps_1,nsteps_2)$ (\equationref{eq:colomn})
as a sequence
converges to $b_j$ at $m$,
$nsteps_1$, and $nsteps_2$  tending to infinity.
\end{definition}

In other words, for sufficiently large
$iter$, $nsteps_1$, and $nsteps_2$ the elements
of $U$ at positions beyond $iter$
fall within arbitrary small vicinities
around the orbit's elements.

We will refer to the orbits in this definition as the bipartite 
orbits of the second type.\footnote{The orbit elements in the 
definition are not necessarily
attractors.}

In our experiments described in
\sectionref{sec:Bipartite Orbits},
we observed
convergence to the bipartite
orbits of the second type
for every initial $Im$.
We also observed similar phenomenon in a simplified model of inter-personal communication
(see \citet{SI}~F).

The questions that arise are:
\begin{enumerate}
    \item \label{itm:zero}
When is $U$ a bipartite sequence of the second type?
    \item \label{itm:sec}
    What are the properties of
    the bipartite sequence
    of the second type?
\end{enumerate}

Theorems~\ref{theorem_L_A} and \ref{thr}
below answer these questions under certain natural conditions, characterizing the behavior of the sequences of inter-transmitted images in  metric and Euclidean spaces, respectively.

Let $X$ be a complete metric space, and let $r=1,2$. For each $r$, let $F_{P_r}$ be a continuous function $X \rightarrow X$, and let $\mathcal{A}_r$ be a subset of the set of attractors of $F_{P_r}$. The function $\widehat{F_r}: X \rightarrow X$ (\equationref{eq:F_widehat}) denotes the mappings to attractors of $F_{P_r}$.

One may see that 
the awareness properties 
of \equationref{eq:fixed_points_F_P}
hold:
\begin{equation}
\label{eq:fixed_points_F_P_1and2}
\begin{aligned}
&F_{P_r}(x) = x, \; \text{for } x \in \mathcal{A}_r.
\end{aligned}
\end{equation}

Define $\alpha(r)$ as
  \begin{equation*}
    \alpha (r)=
    \begin{cases}
      2, & \text{if } r=1 \\
      1, & \text{if } r=2.
    \end{cases}
  \end{equation*}

For $x\in X$
and 
$\epsilon>0$
$B_{\epsilon}(x)$
denotes
the open ball 
\begin{equation*}
B_{\epsilon}(x) = 
\{
x' 
\;| \;
 d(x,x') < \epsilon
\}
.
\end{equation*}

Also, given a function 
$f: X \rightarrow X$, define 
$f(B_\epsilon(x))$ 
as the image of the $\epsilon$-ball under $f$.

The theorem below states the bipartite convergence of the second type for continuous functions in metric spaces under several natural conditions. These conditions are related to the arrangement of the attractors, implying that the attractors in $\mathcal{A}_r$ do not belong to the borders of the basins of the attractors in $\mathcal{A}_{\alpha(r)}$. Another condition for the bipartite convergence is the local uniform convergence of the sequences of functions $F_{P_r}^{[n]}$ to the attractors in 
certain open neighborhoods of their respective attractors.
\begin{theorem}

  \label{theorem_L_A}

If for $r=1,2$, the following conditions hold:

\begin{enumerate}[(a)]

\item 
\label{A1}

Sets $\mathcal{A}_r$  are finite and disjoint.

  \item 
  \label{A5}
  An $Im \in X$ belongs to
  basin $\mathcal B (a)$
of  some   
$a \in \mathcal A _1$.

\item 
\label{A2-new}
Every  $a \in \mathcal A_r$ belongs to the basin $\mathcal B(a')$ of some  $a' \in \mathcal A_{\alpha (r)}$
together 
with an open ball of a certain radius 
${\delta}(a)$ around $a$:
$B_{\delta (a)}(a) \subset
\mathcal B(a')
$.

\item 
\label{A2.5-new}

For every  $a \in \mathcal A_{r}$
convergence of 
the sequence of functions $(F_{P_r}^{[n]})$
to $a$ is locally uniform at $a$:
there exists $\delta (a) > 0$ such that for any $\epsilon > 0$, there exists $n_0$ such that
\begin{equation*}
F_{P_r}^{[n]}(B_{\delta(a)}(a))
 \subseteq
 B_{\epsilon}(a)
\end{equation*}
for any $n \geq  n_0$.

  \end{enumerate}
  
Then  
  the sequence of
\equationref{eq:U0}    
\begin{equation*} 
\begin{split}
U(Im,nsteps_1,nsteps_2)=
Im
\xrightarrow{  [F_{P_1}]^{nsteps_1} }
T_{1,P_1}
\xrightarrow{  [F_{P_2}]^{nsteps_2} }
\\
T_{2,P_2} \xrightarrow{  [F_{P_1}]^{nsteps_1} }
T_{3,P_1}
\xrightarrow{  [F_{P_2}]^{nsteps_2} }
T_{4,P_2}
\xrightarrow{  [F_{P_1}]^{nsteps_1} }
\dots
\;.
\end{split}
\end{equation*}
  is a bipartite
  convergent
 sequence of the   
  second type,
  converging to the orbit
  consisting of alternating
  attractors
  of
$F_{P_1}$ and $F_{P_2}$.

The orbit elements form a loop
with respect to alternating
$\widehat{F_1}$, $\widehat{F_2}$ operations:
\begin{equation}
\label{eq:att2}
{b_0}
\xrightarrow{  {\widehat{F_2}} }
{b_1}
\xrightarrow{  {\widehat{F_1}} }
\dots
\xrightarrow{  {\widehat{F_2}} }
\,
{b_{K-1}}
\,
{\xrightarrow{  \widehat{F_1} }}
\,
{b_0}
.
\end{equation}

These orbit elements
are also
alternating fixed points of functions
$\widehat{G_1}$ and $\widehat{G_2}$:
\begin{equation}
\label{eq:G1G2hatsforFixed}
\begin{aligned}
&\widehat{G_1}(b_h) = b_h, \text{ for even } h,
\\
&\widehat{G_2}(b_h) = b_h, \text{ for odd } h,
\end{aligned}
\end{equation}
where
\begin{equation}
\label{eq:hatG1hatG2}
\begin{aligned}
&\widehat{G_1} = [\,\widehat{F_1} \cdot \widehat{F_2}\,]^{K/2},
\\
&\widehat{G_2} = [\,\widehat{F_2} \cdot \widehat{F_1}\,]^{K/2}
.
\end{aligned}
\end{equation}

\begin{proof} 
  See \citet{SI}~H.
\end{proof}
  \end{theorem}

Equations~\ref{eq:att2} and \ref{eq:G1G2hatsforFixed} 
are the counterparts
of  
Equations~\ref{eq:att1} and \ref{eq:G1G2forFixed} in
\sectionref{sec:subsub:Bipartite Orbit 1}.
  
The theorem is illustrated in \figureref{fig:theorem_L_A}.  

The following statement is
well known (for example \citet{Belkin}):
  \begin{lemma}
\label{lemma_B}
If $a$ is a fixed point of a differentiable map $F: X \rightarrow X$ in Euclidean space $X$, and all eigenvalues of the Jacobian  of $F$ at $a$ are strictly less than 1 in absolute value, then $a$ is an attractor of $F$.
  \end{lemma}

The operator norm of the Jacobian  of an operator $F$ satisfying the lemma is strictly less than 1.
Considering approximation
of $F$ by the differential of
$F$ at $a$, 
one may show 
that
for certain $\lambda$, $0 < \lambda < 1$, and $\delta > 0$, the following holds: 
\[ \|F(x) - F(a)\| < \lambda \|x - a\| \]
for any $x \in B_{\delta}(a)$. This ensures  local uniform convergence of the sequence of functions $(F^{[n]})$ to the attractor $a$
in an open neighborhood  of $a$. Therefore, the following lemma holds:

  \begin{lemma}
\label{lemma_L_aux_A}
  
The conditions of
\lemmaref{lemma_B}
guarantee
locally uniform 
convergence of the sequence of
functions $(F^{[n]}) $
to the attractor $a$
in an open neighborhood of   $a$.
\end{lemma}


%
Now we obtain the theorem which asserts the bipartite convergence of the second type for differentiable maps under well-established conditions regarding the existence of the attractors and their natural arrangement  (see the related statement preceding \theoremref{theorem_L_A}):

\begin{theorem}
\label{thr}

Let $r=1,2$. Let
$\mathcal{F}_r$ be a subset
of the set of fixed points of a differentiable map
$F_{P_r}:
X \rightarrow X
$ in Euclidean space $X$.

If the following conditions hold:
\begin{enumerate}[(a)]

\item 
\label{eigens}
For any 
$a$ in  $\mathcal F_{ r}$
all eigenvalues of the
Jacobian of $F_{P_r}$  at $a$ are strictly less than 1 in absolute value.

\item 
\label{A1F}

Sets $\mathcal{F}_r$  are finite and disjoint.

  \item 
  \label{A5F}
  An $Im \in X$ belongs to the
  basin $\mathcal B (a)$ of some
  $a \in \mathcal F _1$.

\item 
\label{A2-newF}
Every $a \in \mathcal F_r$ belongs to the basin $\mathcal B(a')$ of some   $a' \in \mathcal F_{\alpha (r)}$
together 
with an open ball of a certain radius 
${\delta}(a)$ around $a$:
$B_{\delta (a)}(a) \subset
\mathcal B(a')
$.

\end{enumerate}

Then hold conclusions of
\theoremref{theorem_L_A}.

\begin{proof} 

By \lemmaref{lemma_B}, every
$\mathcal F_{r}$ 
consists of attractors
of $F_{P_r}$
and therefore hold conditions
\ref{A1}, \ref{A5}, and 
\ref{A2-new} of 
\theoremref{theorem_L_A}.
By \lemmaref{lemma_L_aux_A},
from condition \ref{eigens}
follows condition 
\ref{A2.5-new}
of \theoremref{theorem_L_A}.
\end{proof}
  \end{theorem}

The properties of the sequences of interchanged images considered in this section will receive a semiotic interpretation in
\sectionref{sec:rikuz:semiotics}.
Finally, \tableref{tab:att}
summarizes the
properties of bipartite
orbits.

\begin{table}[h]
    \centering
    \caption{Properties of bipartite orbits}
    \vspace{1.em}
    {\begin{tabular}{|l|p{2.9cm}|p{3.2cm}|p{3.2cm}|p{2.6cm}|}    
        \hline
        
        \abovestrut{3.ex}
        \bfseries \#  & \bfseries Property &  \bfseries First Type& \bfseries Second  Type & \bfseries References \\\hline
        
        1 &
        Infinity limit parameters&
        $nsteps_1$,  $nsteps_2$ &
        $niter$, $nsteps_1$, $nsteps_2$  
        & 
        \\\hline
        
\abovestrut{3.0ex}        
2 &
Alternating cyclic transition functions
&
$F_1,F_2$
&
perceptualization operators
$\widehat{F_1}, \widehat{F_2}$
& 
Equations~\ref{eq:F1F2},\,\ref{eq:F_widehat},  and \ref{eq:att2}
\\\hline

        \abovestrut{3.ex}
        3 &
Consists of the percept images
        &
                Typically not
        &
        Yes
        & 
\equationref{eq:F_widehat}               
        \\\hline

        \abovestrut{3.ex}
        4 &
        Attractors of $F_{P_1},$ $F_{P_2}$
        &
        Typically not&
        Yes (Theorems~\ref{theorem_L_A} and \ref{thr})
        & \equationref{eq:fixed_points_F_P_1and2}
        (awareness properties)
        \\\hline

        \abovestrut{3.ex}
        5 &
        Consists of the percept
        images of the dialogue
        &
                Yes
        &
        Yes
        & 
        See 
        \sectionref{sec:rikuz:semiotics:person to person}
        \\\hline

\abovestrut{3.0ex}
        6 &
        Fixed points identities functions &
        $G_1, G_2$
        & 
        $\widehat{G_1}, \widehat{G_2}$
        & 
        Equations~\ref{eq:G1G2},\,\ref{eq:G1G2forFixed},  and \ref{eq:G1G2hatsforFixed}
        \\\hline
        
\abovestrut{3.0ex}        
        7 &
        Validation of existence &
        Observed experimentally&
        Proven under certain natural conditions. Observed experimentally
        & Theorems~\ref{theorem_L_A} and \ref{thr}
        \\\hline

    \end{tabular}
    }
    \label{tab:att}
\end{table}

\section{The CONN Classifiers}
\label{sec:SNRA_classifier}

In this section, we introduce the conversion 
of a given baseline image classifier into 
vanilla and stochastic attractor-based classifiers. 
The conversion is implemented as the addition 
of a new "perceptual" layer that precedes the input 
to the baseline classifier. The obtained classifiers 
are visualizable, enabling us to observe the images 
"perceived" by the network and associate them with 
the training examples. The stochastic classifier 
demonstrates effectiveness for classification tasks 
with small training datasets. However, the 
effectiveness and visualizability come at the cost 
of longer inference time, as input samples take 
longer to converge to attractors.

Given a baseline classifier $M$ and a training dataset 
$TR$, the conversion to a CONN classifier (which can be 
either vanilla or stochastic) proceeds according to the 
following framework.
First, we train an overparameterized 
autoencoder on $TR$. Using the autoencoder, we transform 
input images into the respective images "perceived" 
by a CONN classifier (the use of this term is explained 
in \sectionref{sec:rems_classifiers}). This transformation 
is based on constructing image sequences that converge 
to the attractors of the autoencoder.

The transformation proceeds for every training image, 
as well as for the image used in the inference. In 
both cases, the baseline classifier treats the 
transformed images as if they were the original inputs.

The flowchart of 
the CONN classifiers is shown
in 
\figureref{fig:image-a9}.
The transformation 
$F$ to the 
"perceived" images
converts the training 
set $TR$ and  the test set $TE$ into  new sets $ATR$ and $ATE$ respectively. The latter are used as the new training and test datasets for the baseline classifier.
The notation 
$TE \xrightarrow{F} ATE$
is used for the analysis of the classifier; calculation of the 
classifier value during inference
proceeds independently on other image samples.

In the upcoming sections, we describe two types of attractor-based classifiers: vanilla and stochastic.
The stochastic classifier demonstrates improved classification performance at the cost of a larger inference time.

\subsection{Vanilla Classifier}
\label{sec:atts_classifier_vanilla}

In this section, we introduce the conversion of a given image 
classifier $M$ into a vanilla CONN classifier. The images 
"perceived" by the CONN classifier  consist of 
the attractors of the autoencoder, which is trained on 
the training set of the baseline classifier $M$.

For a given image $Im$, consider the limit of the transformation defined in \equationref{eq:3}. We reproduce this formula as follows:
\begin{equation}
\label{eq:nakaonetc}
\widehat{F}(Im) =
\lim \limits_{n\to\infty} [dec(enc)]^n(Im)
.
\end{equation}
where the limit is taken over successive applications of the encoder-decoder pair.

Empirical evidence by \citet{Belkin} demonstrates that, for an arbitrary image $Im$, the sequence of \equationref{eq:nakaonetc} typically converges to an attractor 
$a$, which can be a memorized example or a spurious attractor. In the case of the vanilla classifier, 
the data samples 
converge to attractors following 
\equationref{eq:nakaonetc} and then 
passed to classifier $M$, 
trained on $ATR$, for prediction.

\begin{figure}[h!]
    \centering
          {\fbox{\includegraphics[height=0.4\linewidth]{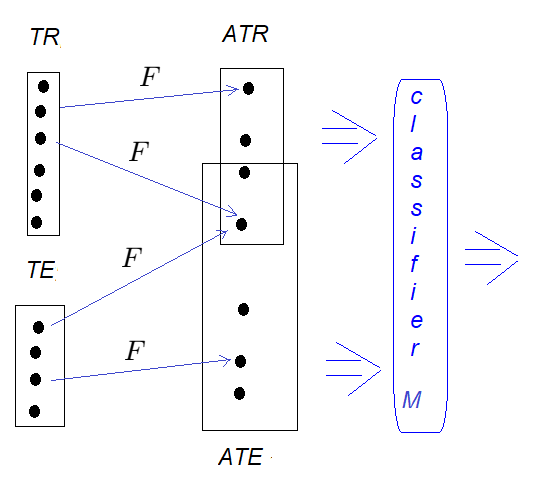}}}
      \caption{
Representation of the work of the CONN classifiers as 
the transformation $F$ of a training set $TR$ (resp. test 
set $TE$) to a new training set $ATR$ (resp. test set 
$ATE$) consisting of the images "perceived" by the 
classifier. For the vanilla classifier, the transformation 
$F$ denotes the transformation $\widehat{F}$ to the attractor 
(\equationref{eq:nakaonetc}). For the stochastic 
classifier, $F$ denotes the transformation $F^*$ to the averaged 
randomized ensemble of attractors 
(Equation~\ref{eq:ARA})      }
          \label{fig:image-a9}
        \end{figure}

Specifically, given a training image dataset $TR$, we first train an overparameterized autoencoder $A$ to memorize 
examples of $TR$ (without using the labels of $TR$). We then construct a new training dataset comprised of attractors:
\[
ATR = \{ \widehat{F}(Im) \,|\, Im \in TR    \}
.
\]
We assign the same labels to the images
$\widehat{F}(Im)$ as to $Im$.
Assuming  the memorization
of the images
 from $TR$, dataset $ATR$ is a twin of $TR$.\footnote{
We follow 
the framework shown in 
\figureref{fig:image-a9}.
For the stochastic classifier considered in the next section, $ATR$ typically differs from $TR$.}
Dataset $ATR$ is then used to train the baseline classifier $M$.

At  inference, an input 
$Im$ is first converged to 
$\widehat{F}(Im)$. Further,
the inference value of the CONN classifier
is defined as  
the value of the trained $M$ at 
$\widehat{F}(Im)$.
\footnote{
We assign an arbitrary label to the images $Im$ 
for which the attractor $\widehat{F}(Im)$ does not 
exist. Although the existence of an attractor for 
an arbitrary  $Im$ is not guaranteed (see \citet{SI}~H, 
Figure~4), the number of such images is negligibly 
small. We did not observe any such images in our 
experiments with overparameterized autoencoders
(\sectionref{sec:ExpClassifiers}).
}

From this, it follows that the vanilla CONN classifier assigns the same label to all images within to the basin $\mathcal{B}(a)$ of an attractor $a$.

Let an image $Im$ belong to the basin of
a training example $a$ memorized as the attractor. It can be seen that, 
assuming 
the baseline classifier $M$ properly classifies the training examples from $ATR$, the vanilla classifier assigns to $Im$ the ground truth label of $a$. 
In this sense, 
the vanilla classifier function is similar to a 1-nearest neighbor classifier, where 
the attractor
$a$ serves  as the "closest" training example to $Im$.

Experimental results for the vanilla CONN  classifier are presented in \sectionref{sec:ExpResults}.

\subsection{Stochastic Classifier}
\label{sec:atts_classifier_multi}

Below we introduce the stochastic CONN classifier. It provides better classification results than its vanilla counterpart, albeit with increased inference computational time.

The rationale behind it may be explained as follows. As seen in
\sectionref{sec:atts_classifier_vanilla},
given an image $Im$, the inference of the vanilla CONN classifier is equivalent to selecting the attractor $d$ to whose basin $Im$ belongs and assigning to $Im$ the ground truth label of $d$. This approach  leads to misclassification when $Im$ and $d$ have different ground truth labels. However, the ground truth labeling of several elements in the neighborhood of $Im$ may better characterize the ground truth labeling of $Im$ than that of a single element $d$. In this sense, representing $Im$ via several neighboring attractors may be more informative
(see \citet{SI}~J.)
Actually, we apply here the idea of transitioning from 1-NN to k-NN to our vanilla classifier.

Our approach is as follows.
Instead of representing $Im$ solely by a sequence of elements converging to an attractor, we construct $J>0$ sequences that start with $Im$ and converge to attractors. 
Similarly to the vanilla classifier,
these sequences are built following 
\equationref{eq:nakaonetc}, while also incorporating random augmentations. As a result, we obtain an ensemble of $J$ attractors that represent $Im$. (The ensemble may contain repetitions of attractors).  
Finally, we derive the final attractor representation $F^*(Im)$ by averaging the ensemble in the image domain.

Specifically, given an autoencoder and an input image $Im$,  the
average of  the ensemble of attractors
is defined as:
\begin{equation} 
\label{eq:ARA}
F^*(Im) =
\frac{1}{J} \sum_{j=1}^{J} a_j,
\end{equation}
where  the ensemble
\begin{equation} 
\label{eq:AS}
\{ a_j \mid j=1,\ldots, J \}
\end{equation}
is comprised of $J$ attractors
\begin{equation} 
\label{eq:AA}
a_j =
\lim_{i \to \infty} x_{i,j},
\end{equation}
where
\[
x_{0,j} = Im \,,
\]
and
\begin{equation}
\label{eq:i-iter}
x_{i+1,j} = dec(enc( \tau _{i, \gamma _i} (x_{i,j})))
.
\end{equation}
for $i\geq 0$.

The term $\tau _{i, \gamma _i} (x_{i,j})$ denotes a sampling of random augmentation $\tau _{i, \gamma _i}$ 
applied to images $x_{i,j}$, where the magnitude of augmentation is denoted by $\gamma _i$. 

When $\gamma_i = 0$, no augmentation is applied to the image. 
The assignment $\gamma_i = 1$ corresponds to  the maximum  level of augmentation.
The value of $\gamma_i$ is determined using the formula:
\begin{equation}
\label{eq:beta}
\gamma _ i= \beta ^{\frac{1}{i+1}} \, ,
\end{equation}
where the parameter $\beta > 1$ controls the relaxation of the augmentation amplitude as $i$ increases.

The experimental results in \sectionref{sec:ExpClassifiers} demonstrate that the stochastic classifier outperforms its vanilla counterpart.

\subsection{Remarks on Classifiers}
\label{sec:rems_classifiers}

It is worth noting that although the stochastic CONN
classifier explores augmentations,
the approach itself is not an augmentation of the training examples.
In fact, the number of training examples in the stochastic
classifier remains the same as in the vanilla version.

The transformation in \equationref{eq:nakaonetc}
that turns $Im$ into an attractor represents the final form of the observed-to-seen transformations in \equationref{eq:2}. 
 Therefore, it is natural to refer to attractor $\widehat{F}(Im)$,
as the image "perceived" by the vanilla classifier 
given
an "observed" 
image $Im$. 
This justifies the notation
\begin{equation} \label{eq:seen_vanilla}
perc_{V}(Im) = \widehat{F}(Im).
\end{equation}

Similarly, we will refer to $F^*(Im)$ as the image
"perceived" by a stochastic classifier $C_S$:
\begin{equation} \label{eq:seen_stoch}
perc_{S}(Im) = F^*(Im).
\end{equation}

Currently, the memorization of training data was demonstrated for autoencoders trained on data sets consisting of up to several hundred examples (\citet{Belkin}). This limitation restricts the effective usage of the CONN-based classifiers to situations where the training data is limited in size.

\iftrue 
In the stochastic CONN classifier, we perform a series of converging sequences, where each sequence is terminated by an attractor. The attractors in the series may vary, but they demonstrate consistency throughout the series. For example, the set of attractors obtained
for $j$ ranging 
from $0$ to $50$
is similar to that
for $j$ ranging 
from $51$ to $100$. Additionally, the terminating elements (attractors) are predefined, meaning they are determined solely by the training examples.

This allows us to view the stochastic CONN classifier from the perspective of visual perception, particularly in relation to multistable perception (\citet{Gage2018}). Multistable perception, as demonstrated by the Rubin's face-vase illusion and similar phenomena (\citet{Ittelson}), involves the perception of different patterns. These patterns are typically consistent and predefined for individuals over time, although different individuals may perceive different patterns.
For instance, in the Rubin's vase/face illusion, the perceived patterns typically consist of either a vase or a face.

In this regard, the stochastic CONN classifier mimics the properties of consistency and predefinency observed in human multistable perception. 
\fi

\section{Semiotic Interpretation of the Model}
\label{sec:rikuz:semiotics}

In this section, we will explore how our model describes the phenomena of human perception and communication. We begin by discussing the perception of a visual object by a single person, followed by an exploration of two-person communication.

\subsection{Perception of a
Visual Object by a Person}
\label{sec:rikuz:semiotics:person-object}

The goal of this section is to specify the  relations that describe human perception of visual objects and demonstrate how the communication model introduced in \sectionref{sec:Modeling} incorporates these relations.
We proceed as follows:
first, we will
formalize some properties
of human visual object perception, to derive  relevant mathematical relationships.
Then, we consider
how these relations are
represented in our model.

We focus on the "atomic" perception, which involves the process of identifying a specific object within a specified period of time. Note that the perception of objects in different times and spaces, which is related to object perception in a general sense, is beyond the scope of the current work.

Persons see and "see" objects.
In other words, they are doing two separate actions. First, they see, namely perceive objects using their designated devices -- usually their eyes. Then, they become {\em aware} of that
object.  Further
actions may be taken
based on the perception to accomplish
specific tasks. For example,
imagine a situation in which
a car is coming fast towards
you. First, you see the car
("see"), then you identify
the car approaching you
("aware"), and finally,
you step onto the sidewalk
("action").
Here, we formalize
the first two steps -- "see" and "aware".

The process of seeing the physical image is complex and involves various stages of image processing, feature extraction, and visual perception mechanisms in the human visual system. It encompasses the physiological and cognitive processes through which the visual information from the image  is interpreted and translated into the perceived image. This includes the extraction of relevant visual features, the integration of contextual information, and the interpretation of the visual scene based on the individual's cognitive processes and prior knowledge.

It is important to note that the process of seeing the image is subjective and may vary among individuals due to differences in their visual perception abilities, cognitive processes, and prior experiences. Environmental factors such as lighting conditions and viewing distance also influence  the perceiving process.

In our model, we conceptualize the process of "seeing" the image as a series of successive image processing steps that generate a new image. Note that the seen object is in the same modality.

Let $x$ represent a specific image that is observed by a person. For example, $x$ could be a digital image. The seeing process involves the conversion of $x$ into a seen image denoted as $seen(x)$. We can treat $seen(x)$ as a new image, of similar modality, which represents the image that the person perceives. This conversion can be represented mathematically as:        \[
        x  \rightarrow seen(x)
        .
        \]

For example, $x$ is a given image of
a dog, and $seen(x)$ be
an initial visual representation
of the dog. Note that the latter 
visual representation
may differ
from the initial one,
but it is still an image.

The $seen$ operator might be
slightly distinct
for different persons.
For example, people may
see
dissimilar details in an observed image.
Note that attention
is only a part
of the internal representation.
This reflects the
phenomenon that people perceive objects differently.

\begin{figure}[htbp]
\centering
  {
    \subfigure[]{\label{fig:image10-a}
      \includegraphics[width=0.18\linewidth]{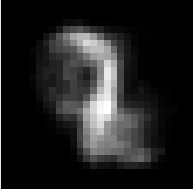}}
    \subfigure[]{\label{fig:image10-b}
      \includegraphics[width=0.18\linewidth]{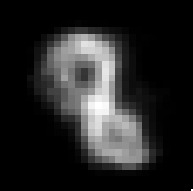}}
    \subfigure[]{\label{fig:image10-c}
      \includegraphics[width=0.18\linewidth]{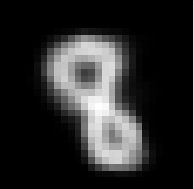}}
    \subfigure[]{\label{fig:image10-d}
      \includegraphics[width=0.18\linewidth]{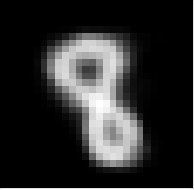}}
    \subfigure[]{\label{fig:image10-r}
      \includegraphics[width=0.18\linewidth]{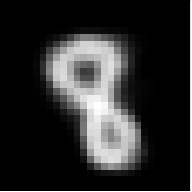}}
  }
    \caption{From left to right: the result of subsequent
    application of $seen$ operator
    observed in the experiments.
    Simulated is the visual perception of a person
    with the seen functionality biased
    to perception
    of the even digits. 
    (a): An observed image $x$,
    (b): $seen(x)$, $\ldots$\,,
    (e): $seen(seen(seen(seen(x)))))$
    }
  \label{fig:gradual}
\end{figure}

        Further, we  formalize
        the
        process of seeing
        as sequential application
        of $seen$ function:
        \begin{equation} 
        \label{eq:seen}
        x  \rightarrow seen(x) \rightarrow
        seen(seen(x))
        \rightarrow \dots
        \;.
        \end{equation}
        For example,
        during perception process of an image, its details
        may become clearer
        in a gradual fashion, this is illustrated in
        \figureref{fig:gradual}.

What are the
relations that reflect
the
image perception
awareness?
Direct access to awareness
metrics is hard
(as it may involve operations procedures or
require dedicated equipment,
that is expensive), therefore
we employ a mediated method
of semantic analysis.

Consider the statement "I see this image".
In this sentence, "this image" has two meanings.
First, it refers to the object itself (in the relevant modality). For example,
an image of a dog.
Second, it refers to an internal perceived image
which is a translation
of the original image.
An inherent property of a consistent communication system
is to make these two meanings close to each other,
namely, to make
$\widehat{x}$ equal to $seen(\widehat{x})$:
        \begin{equation}
        \label{eq:awareness0}
        seen(\widehat{x}) = \widehat{x}
        \,
        ,
        \end{equation}
where $\widehat{x}$ is the final
image representation
referred as "this image".
We will refer to this fixed-point equation as the {\em awareness property}.
A detailed exploration of the  above sentence with respect to \equationref{eq:awareness0} is given in \citet{SI}~I.

Final representation of
the image perception 
\equationref{eq:seen}
may be formalized
as convergence
        \begin{equation}
        \label{eq:awareness}
        \lim
        \limits_{n \to \infty}
        seen^{[n]}(x) = \widehat{x}
        \
        .
        \end{equation}
We refer to the limit value, if such exists, as the "percept image" of $x$.

Overall, the seeing of visual objects can be formalized as follows:
\begin{itemize}
\item
An operator
$seen()$ acting in the image domain;
\item
Sequential application
of
$seen()$
to the
initial image
(\equationref{eq:seen}).
\end{itemize}

In addition,
awareness in perceiving of visual objects
is formalized
as:
\begin{itemize}
\item
Convergence equation
(\equationref{eq:awareness});
\item
The awareness property
(\equationref{eq:awareness0}).
\end{itemize}

How are these
properties 
described by the model
of \sectionref{sec:Modeling}?
\equationref{eq:2}
in \sectionref{sec:Modeling}
represents the
$seen$ operator:
\begin{equation*}
obs\rightarrow seen:
\,\,
obs \xrightarrow{enc}
 enc(obs)
 \xrightarrow{dec}
 seen  = dec(enc(obs))
 .
\end{equation*}

Step~\ref{nitm:3.2}
of \algorithmref{alg:CONN}
represents
\equationref{eq:seen}
at
$iter$
going to infinity.
Equations~\ref{eq:fixed} and \ref{eq:3}
in \sectionref{sec:pers_one_obj:attractors}
are related to convergence to the fixed points
and represent respectively
Equations~\ref{eq:awareness0} and \ref{eq:awareness}.

In summary,  our model
reflects the following
phenomena
of human visual objects
perception:
the existence of the objects
{\em seen} by a person, as well as the existence of the objects the person {\em is aware of } as such.

\subsection{Person-to-Person Communication}
\label{sec:rikuz:semiotics:person to person}

In this section, we will explore how the properties
of the person-to-person communication are described by the communication model of \sectionref{sec:Modeling}.
We will 
consider  sequences of
 images 
observed, seen, and 
exchanged during communication
and study,
using these sequences as an illustration, 
how the mathematical properties
of the bipartite orbits 
express
the key properties of the communication.

Consider  a sequence $U(Im)$ of the images transmitted during a dialogue, as described in \equationref{eq:U0}
in
\sectionref{sec:subsub:Bipartite Orbit 1}, which we rewrite as follows:
\begin{equation*}
\begin{split}
&U(Im,nsteps_1,nsteps_2)=
Im_1
\xrightarrow{  [F_{P_1}]^{nsteps_1} }
Im_2
\xrightarrow{  [F_{P_2}]^{nsteps_2} }
Im_3
\xrightarrow{  [F_{P_1}]^{nsteps_1} }
\\
&
\;\;\;\;\;
Im_4
\xrightarrow{  [F_{P_2}]^{nsteps_2} }
Im_5
\xrightarrow{  [F_{P_1}]^{nsteps_1} }
\cdots
\;
,
\end{split}
\end{equation*}
where $Im_1=Im$.

In our model, the "internal depth" of communication  depends on the $nsteps$ parameters. This reflects  the fact that communication 
may proceed  in a way where persons delve more or less profoundly into processing the information received during the interaction. This is the 
first
phenomenon of interpersonal communication
modeled by our representation.

Interpersonal dialogue can, after a certain point, become repetitive. 
In our representation, the process of interpersonal communication is typically convergent to an orbit — 
 a repetitive loop of images (\sectionref{sec:subsub:Bipartite Orbit 1}). In this way, the CONN model captures the phenomenon of converging dialogue to a cycle.

May we recognize the functionality
similar to the "seen" and 
"aware" of \sectionref{sec:rikuz:semiotics:person-object} in the inter-person  dialogue?
Here, these notions pose greater challenges for examination compared to the person-object communication. 
Indeed, the perceived content of the 
dialogue is harder to reproduce than 
perception of objects.
While we may feel the entities of the dialogue,
they possess an elusive quality that may evade our conscious recognition.
Similarly to the
person-to-object
communication considered
\sectionref{sec:rikuz:semiotics:person-object},
our awareness  may be limited to the  ultimate form of these entities
in the inter-person 
communication.

Are "seen" and "observed"
in dialogue represented in our model?
To answer this question,
assume 
that the sequence 
$U$ consisting of the 
images transmitted between the persons
converges to a bipartite orbit
$(b_0,b_1,\dots,b_{K-1})$
of the first type. 
We refer to 
\equationref{eq:matrix} in
\sectionref{sec:subsub:Bipartite Orbit 1}.

Consider how 
operator $G_1$
from \equationref{eq:G1G2} 
acts on the elements of
the sequence $U$.

Likewise the observed-to-seen transformation
expressed in \equationref{eq:2} of \sectionref{sec:Modeling}, $G_1$ converts the image to a similar image by passing through the internal representations. 
However, here, the conversion proceeds through a sequence of typically different images constructed using the internal representations of both persons. Therefore, it is natural to consider $G_1$ as the operator
transforming
 the images observed  {\em in the dialogue} to those seen  {\em in the dialogue}. The same holds to $G_2$, as well as to $\widehat{G_1}$ and $\widehat{G_2}$ from \equationref{eq:hatG1hatG2}.

Applying  reasoning similar to that
in \sectionref{sec:rikuz:semiotics:person-object},
we refer 
to the fixed point property 
of bipartite elements $b_h$,
expressed by \equationref{eq:G1G2forFixed}
in \sectionref{sec:subsub:Bipartite Orbit 1}:
\begin{equation*}
\begin{aligned}
&G_1(b_h) = b_h, \text{ for even } h, \\
&G_2(b_h) = b_h, \text{ for odd } h,
\end{aligned}
\end{equation*}
as  representing
the person's awareness that the element $b_h$ is seen in the dialogue. Here,
operator $G_i$ represents
the awareness of the 
$i$-th person.
A similar interpretation applies to the fixed point properties of the second-type orbits of \equationref{eq:G1G2hatsforFixed} in
\sectionref{sec:subsub:Bipartite Orbit 2}.
This allows us to refer to the images satisfying
the fixed point relations
Equations~\ref{eq:G1G2forFixed} and \ref{eq:G1G2hatsforFixed}
as the "percept images of $Im$ in the dialogue".

In such a way, the existence of both types of objects -- those that are seen 
in the dialogue and those that the person is aware of as the seen  is the property of inter-person communication represented  
by our model.

Further, according to the model, the observed content of a dialogue varies for different persons (odd and even positions of the elements
in \equationref{eq:matrix}). The structure of $G_1$ and $G_2$ operators 
(and their ultimate counterparts
$\widehat{G_1}$ and $\widehat{G_2}$
) reveals another non-obvious aspect of dialogue. Namely, not only does the content observed by
a person in a dialogue depend on the other participant ("what"), but also the way in which the person perceives it differs for different participants, being also influenced by the other participant ("how").

At times, the images that we see in the dialogue may be twofold. On one hand, we experience them as reflecting the view of the second person, as discussed above. In this sense, they are "imposed" on us. On the other hand, upon closer inspection, we may begin to feel that these images are actually our own, pre-existing before the start of the communication, with no connection to the other person. In this sense, our dialogue merely served as a pretext for their manifestation.
Our model provides a representation of this phenomenon.

Indeed, as we observed in our experiments 
(\sectionref{sec:Bipartite Orbits}), for large values of $nsteps_1$, 
the elements $b_h$ at even positions $h$ of the first 
type orbits (in
\sectionref{sec:subsub:Bipartite Orbit 1}) became indistinguishable from the fixed points of $F_{P_1}$. As discussed previously, 
the $b_h$s represent the entities perceived in the dialogue by 
the first person. In other words, while the person became more aware 
in perceiving the dialogue entities (as reflected by the increase 
of $nsteps_1$), the perceived entities became indistinguishable 
from the fixed points of $F_{P_1}$.

These points are predetermined before the dialogue and are independent of 
the other person, as well as of the starting image. They encapsulate internal 
image representations inherently associated 
with the person. The specific 
fixed point to which the sequence converges depends on the starting image and the 
other person involved in the communication.

In this way, our model captures   the phenomenon described above: sometimes, the communication dialogue  merely acts as a signal to "wake up" one of the predefined internal representations. And this is another aspect of inter-person communication described by our model.

\section{Experimental Results}
\label{sec:ExpResults}

\subsection{Attractors}
\label{sec:Attractors_P2P}

Our visualization of attractors in the autoencoder latent space is presented in \citet{SI}~K.
The results demonstrate that the memorization of training examples is not necessary for  convergence of sequences of encoding-decoding operations to attractors. 
In our experiments, the sequences initiated from random samples converge to attractors, with approximately 6\% of the cases exhibiting convergence to cycles.

\subsection{CONN  
for person-to-person communication
}

\label{sec:ExpResults_Algo1}

In our implementation of \algorithmref{alg:CONN}
the autoencoders $A_{P_1}$ and
$A_{P_2}$
were 
trained at odd and even digits
(30508 and
29492 images) from the MNIST database (\citet{mnist})
respectively.
The autoencoders are
multi-layer perceptrons consisting of 6 hidden layers, with a depth of 512 units, and the latent space 2, trained at 20 epochs.

\subsubsection{Bipartite Orbits}
\label{sec:Bipartite Orbits}

In the experiments with \algorithmref{alg:CONN}, we varied the parameters $nsteps$ and the initialization images. For each configuration, we observed convergence to the first type orbits: starting from a certain
number the sequence of  images
transmitted between $P_1$
and $P_2$ becomes cyclic.
For $nsteps_1$ and $nsteps_2$ greater than 50, the sequence
of \equationref{eq:U0} 
in
\sectionref{sec:subsub:Bipartite Orbit 1}
did not depend on the specific values of $nsteps_1$ and $nsteps_2$, thus demonstrating 
convergence to 
the second type orbits.
Refer to \figureref{fig:exp_results_ag}.

\begin{figure}[h]
  \centering
  \begin{tabular}{ccc}
    \includegraphics[width=0.13\linewidth]{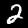} & 
    \includegraphics[width=0.13\linewidth]{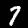} & 
    \includegraphics[width=0.13\linewidth]{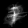} \\
    '2' & '7' & '2' \\
    [6.pt] 
    \includegraphics[width=0.13\linewidth]{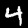} & 
    \includegraphics[width=0.13\linewidth]{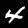} & 
    \includegraphics[width=0.13\linewidth]{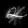} \\
    '4' & '4' & '4' \\
    [6. pt] 
    \includegraphics[width=0.13\linewidth]{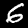} & 
    \includegraphics[width=0.13\linewidth]{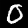} & 
    \includegraphics[width=0.13\linewidth]{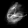} \\
    '6' & '0' & '6' \\
  \end{tabular}
  \caption{
    Simulation of perceptual inference in the vanilla and stochastic classifiers.
    Shown are examples from the test set used in the experiments and the corresponding images "perceived" by the vanilla and stochastic CONN classifiers.
    Left column: Original "observed" images from the MNIST test set, each annotated with its ground truth label.
    Middle column: The respective images "perceived" by the vanilla classifier, annotated with the labels assigned by the classifier.
    Right column: The respective images "perceived" by the stochastic classifier, annotated with the labels assigned by the classifier.
  }
  \label{fig:obs_seen_both_test}
\end{figure}

\subsection{Classifiers}
\label{sec:ExpClassifiers}

\begin{figure}[htbp]
    \centering
    {
         \subfigure[]{
         \label{fig:exp_results_a}
         \includegraphics[width=0.12\linewidth,height=0.12\linewidth]{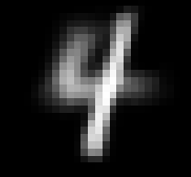}}
         \qquad
         \subfigure[]{
         \label{fig:exp_results_b}
         \includegraphics[width=0.12\linewidth,height=0.12\linewidth]{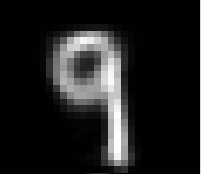}}
         \qquad
         \subfigure[]{
         \label{fig:exp_results_c}%
         \includegraphics[width=0.12\linewidth,height=0.12\linewidth]{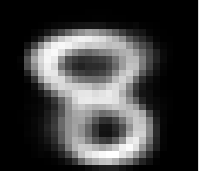}}
        \\\vspace{2ex}
         \subfigure[]{
         \label{fig:exp_results_d}%
         \includegraphics[width=0.12\linewidth,height=0.12\linewidth]{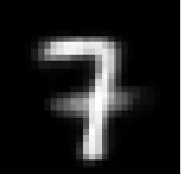}}
         \qquad
         \subfigure[]{
         \label{fig:exp_results_e}%
         \includegraphics[width=0.12\linewidth,height=0.12\linewidth]{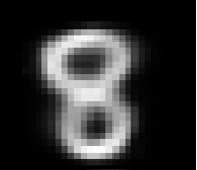}}
         \qquad
         \subfigure[]{
         \label{fig:exp_results_f}
         \includegraphics[width=0.12\linewidth,height=0.12\linewidth]{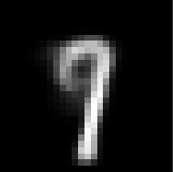}}
         \qquad
         \subfigure[]{
         \label{fig:exp_results_g}
         \includegraphics[width=0.12\linewidth,height=0.12\linewidth]{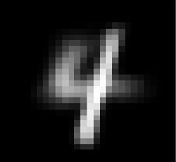}}
    }
    \caption{
    A bipartite orbit of the second type with a period length $K=6$.
The CONN  consists 
of autoencoders $A_{P_1}$ and
$A_{P_2}$ trained at odd and even digits
from the MNIST training data respectively.
    $(a)$: 
    The image transferred from $P_2$ to $P_1$   
    at $iter=9$ (step~\ref{nitm:3.5} of \algorithmref{alg:CONN}
    in \sectionref{sec:Modeling}),
    $(b)$:
    the image transferred from $P_1$ to $P_2$   
    at $iter=10$, $\cdots$ ,
    $(g)$:
    the image transferred from $P_2$ to $P_1$  at $iter=15$.
    Note the difference between "8"s depicted in 
    $(c)$
    and 
    $(e)$
      }
    \label{fig:exp_results_ag}
  \end{figure}

\begin{figure}[]
\centering
  {
    \subfigure[]{\label{fig:image-a5}
      \includegraphics[width=0.55\linewidth]{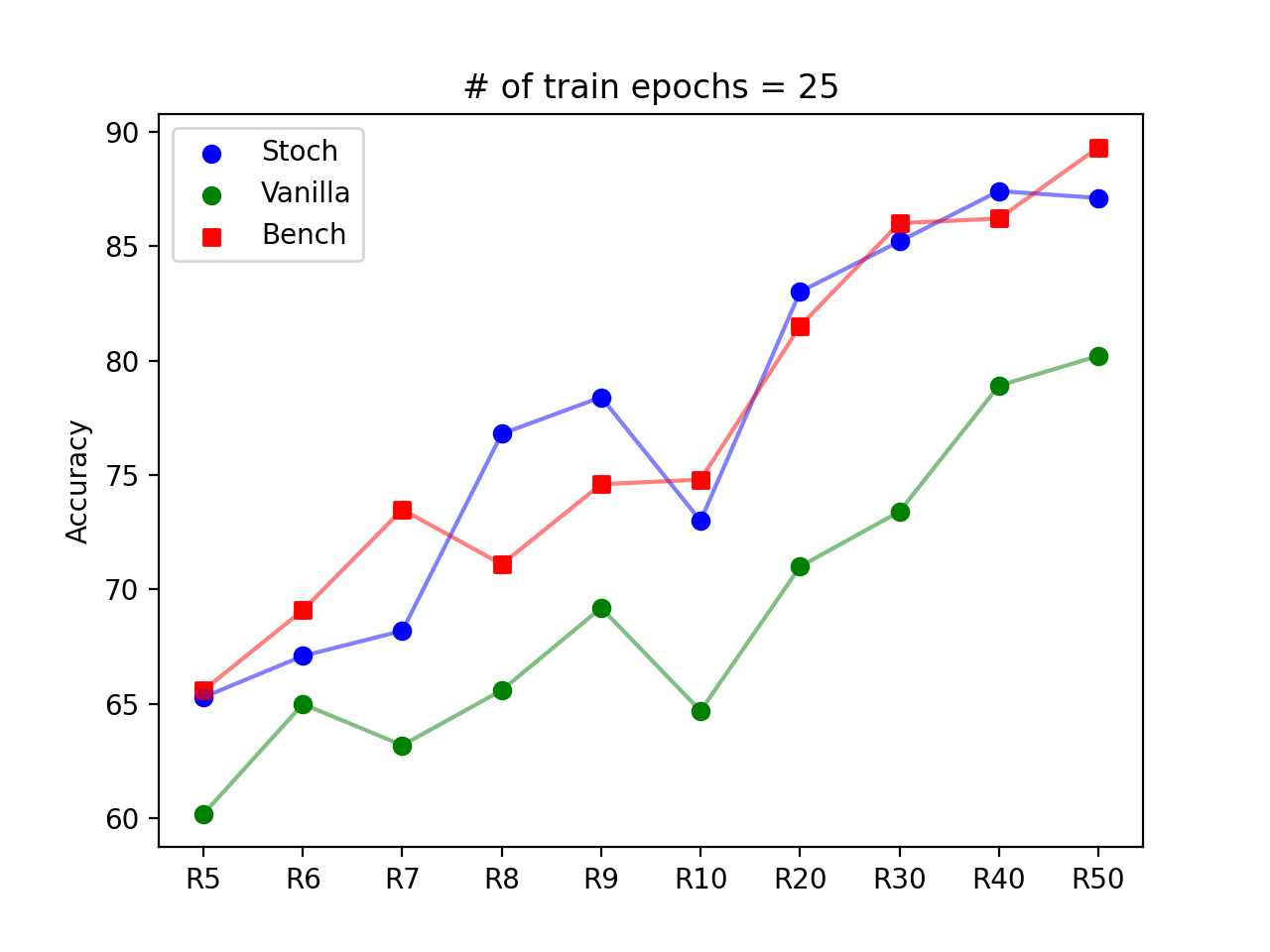}}%
    \subfigure[]{\label{fig:image-b5}
      \includegraphics[width=0.55\linewidth]{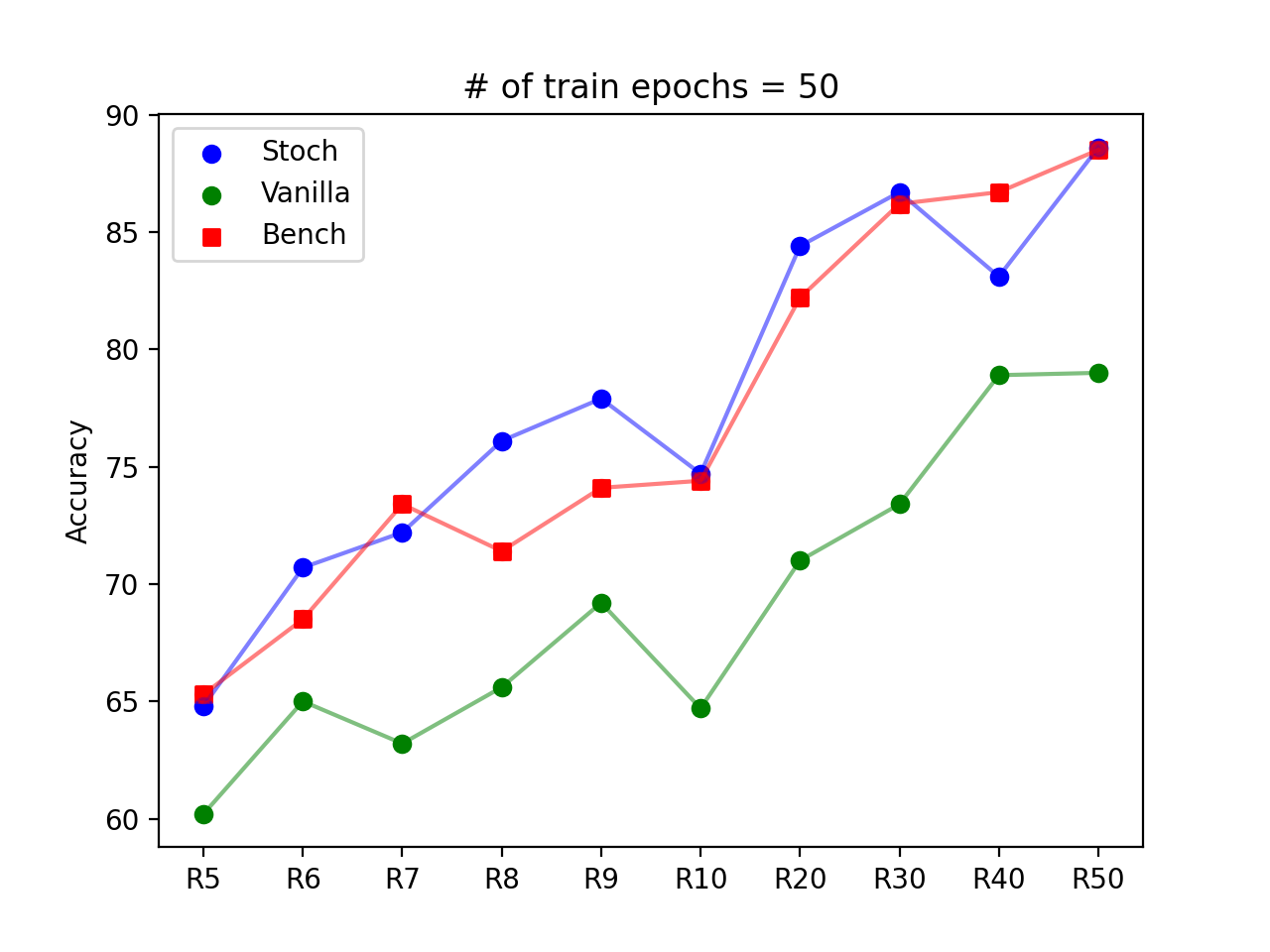}}%
    \\ \vspace{2ex}
    \subfigure[]{\label{fig:image-e5}
      \includegraphics[width=0.55\linewidth]{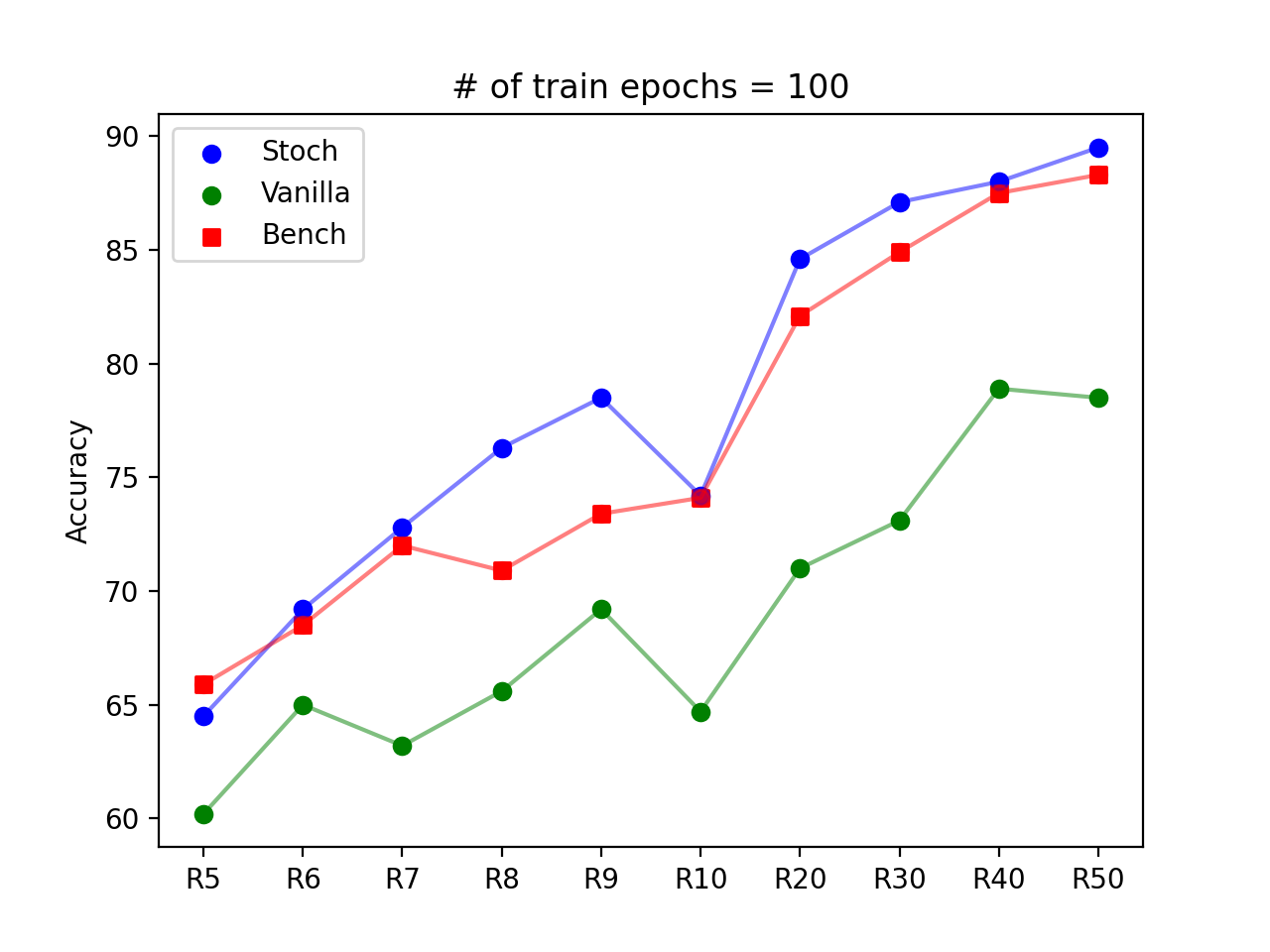}}%
    \subfigure[]{\label{fig:image-f5}
      \includegraphics[width=0.55\linewidth]{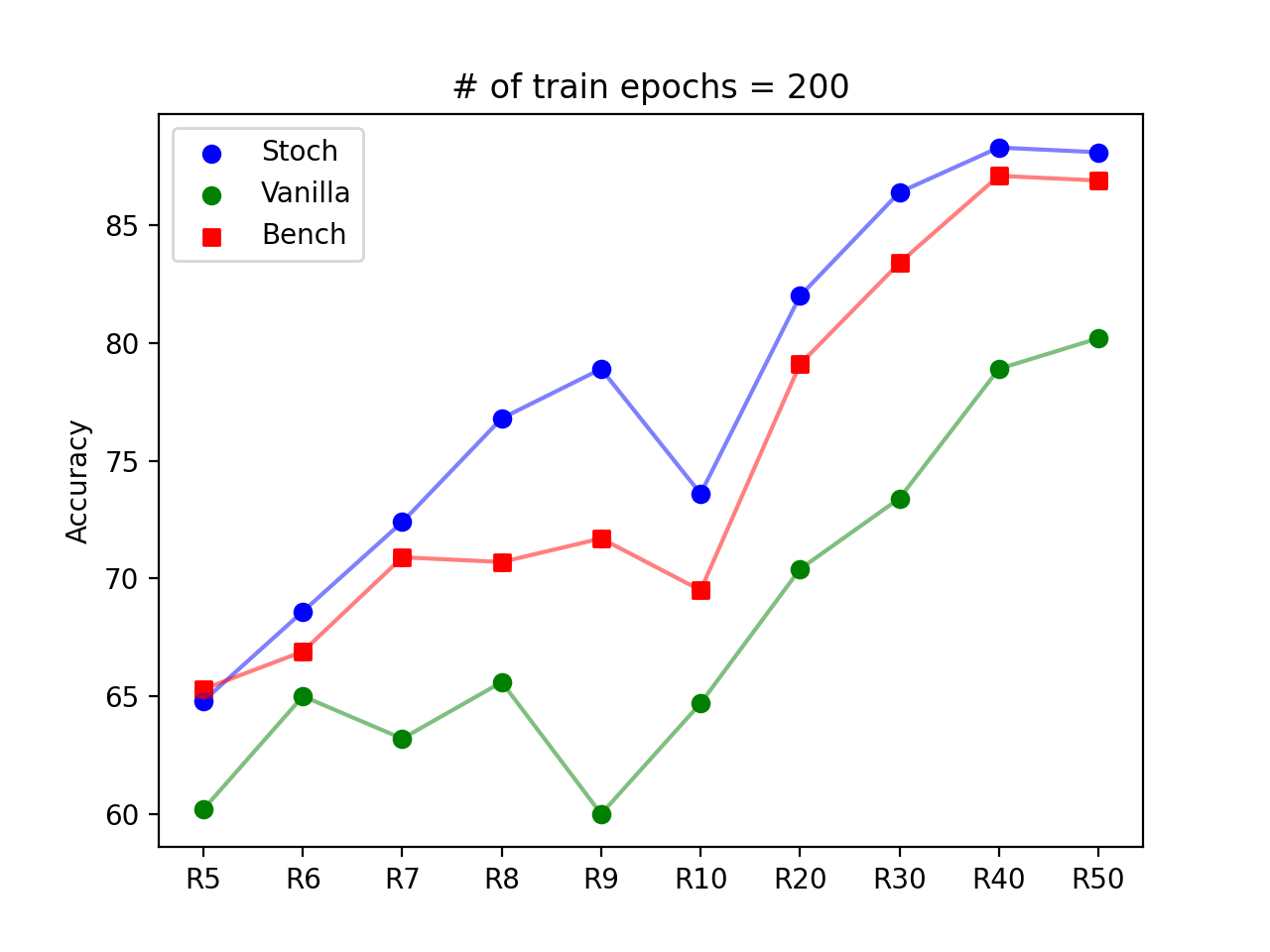}}%
  }
\caption{(a), (b), (c), and (d): Accuracy curves for the stochastic CONN classifier (in blue), the vanilla CONN classifier (in green), and the baseline classifier (in red) with different numbers of training epochs: 25, 50, 100, and 200, respectively. The x-axis tick values represent the size of the restricted MNIST training datasets. For example, R10 corresponds to a training dataset comprising 10 randomly selected examples per digit from the MNIST dataset}
  \label{fig:hconn_results}
\end{figure}

We tested the performance of a standard  MLP classifier $M$ against its CONN vanilla and stochastic counterparts by embedding $M$ within these frameworks, as described in \sectionref{sec:SNRA_classifier}.

Our baseline classifier $M$ is a 3-layer MLP with an input size of 28x28 pixels. It has two hidden layers with 500 and 100 neurons, respectively.

The classifier was trained at 40 training configurations, produced by combinations of 10 training datasets and 4 numbers of training epochs. The 10 training datasets $TR$ were constructed by randomly selecting 5, 6, 7, 8, 9, 10, 20, 30, 40, and 50 examples respectively for every digit from the MNIST training dataset
following \citet{mnielsen2}. The numbers of training epochs were selected as 25, 50, 100, and 200.

The test set $TE$ was constructed by randomly selecting 1000 examples from the MNIST test set.

Our vanilla and stochastic CONN classifiers preprocess the data 
using fully convolutional autoencoders, similar to the autoencoder 
with the Cosid nonlinearity
(\citet{BelkinSI}). We trained the autoencoders with the same architecture on the 10 training datasets $TR$, carefully tuning the hyperparameters to minimize the training error. For detailed information, refer to \citet{SI}~L.

Further, we mapped all pairs of the training and test datasets ($TR$, $TE$) to new pairs of training and test datasets ($ATR$, $ATE$) for exploring the baseline classifier (refer to \figureref{fig:image-a9} in
\sectionref{sec:SNRA_classifier}). As a result, we constructed 10 pairs of datasets ($ATR$ and $ATE$) for the vanilla classifier, and another 10 pairs for the stochastic classifier.

For the vanilla classifier, the mapping was done following \equationref{eq:seen_vanilla}
in 
\sectionref{sec:rems_classifiers}, and for the stochastic classifiers, following \equationref{eq:seen_stoch}.
For the vanilla classifier, the construction of the attractors (\equationref{eq:nakaonetc}) was completed at $n=100$, when subsequent members of the iterative sequence become indistinguishable. In the case of the stochastic classifier, the corresponding parameter $i$ in \equationref{eq:AA} was set to 30.

For the stochastic classifier, 
the geometric and image processing augmentations of  \equationref{eq:i-iter} were generated using the library of \citet{imgaug}. 
The ensemble length $J$ in \equationref{eq:ARA} was set to 500, and the relaxator $\beta$  in \equationref{eq:beta} was set to 2.6.

The construction of the images forming 
the $ATE$ sets for the vanilla and 
stochastic 
CONN classifiers is illustrated in 
\figureref{fig:obs_seen_both_test}. 
Additionally, refer to \citet{SI}~M for details on the construction of the image "perceived" by the stochastic classifier, shown in the middle row's right column of the figure.

The performance results for the baseline classifier $M$ were obtained by training $M$ on 10 sets of $ATR$ over 4 training epochs (see above), followed by testing the trained models on the $TE$ set. The results for the vanilla and stochastic CONN classifiers were obtained by training on the respective 10 sets of $ATR$ over 4 training epochs, followed by testing the trained models on the 10 respective sets of $ATE$.

\begin{figure}[] 
    \centering
      {
        \subfigure[]{\label{fig:image-a7}
          \includegraphics[width=0.5\linewidth]{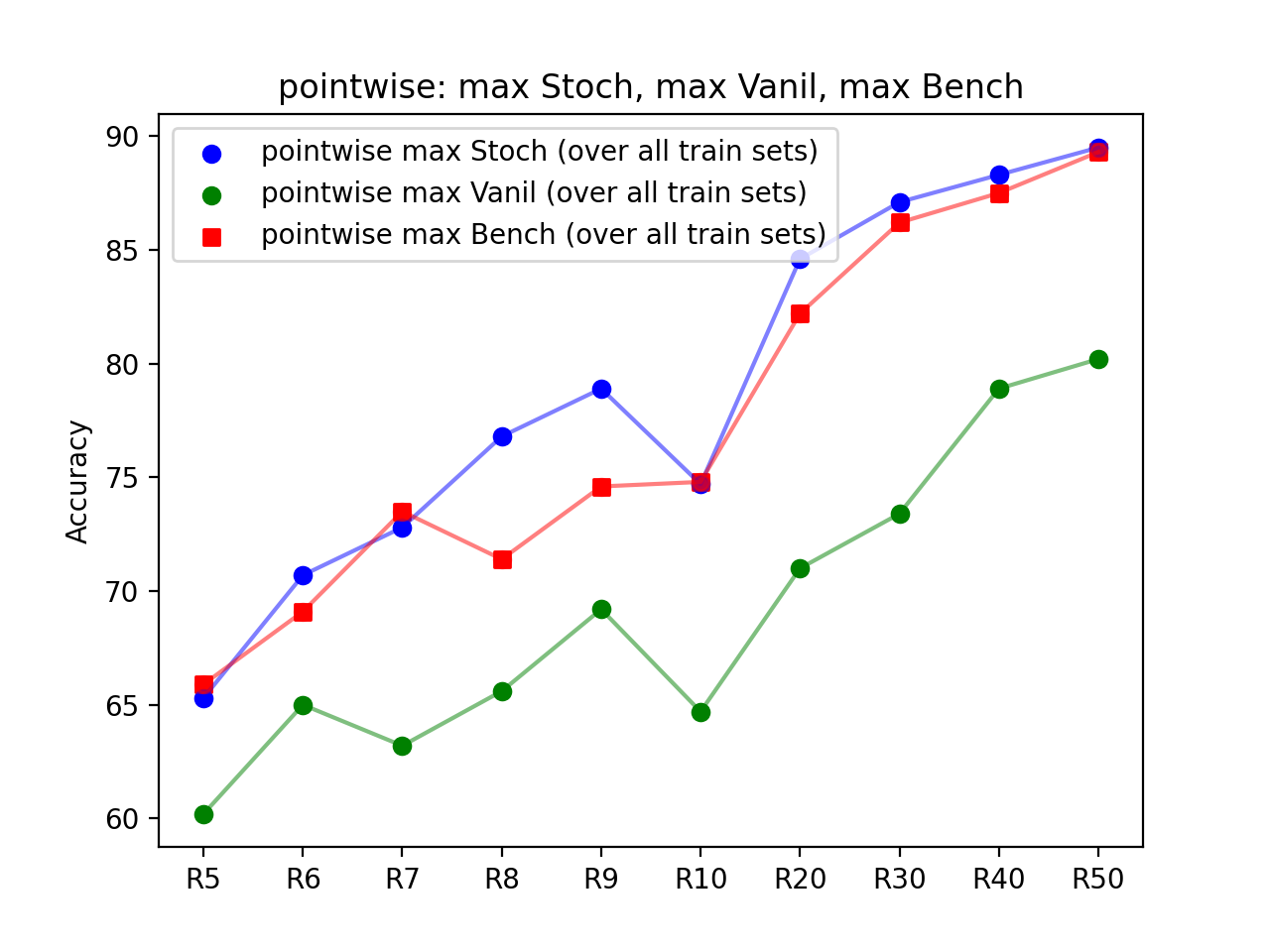}}%
        \subfigure[]{\label{fig:image-b7}
          \includegraphics[width=0.5\linewidth]{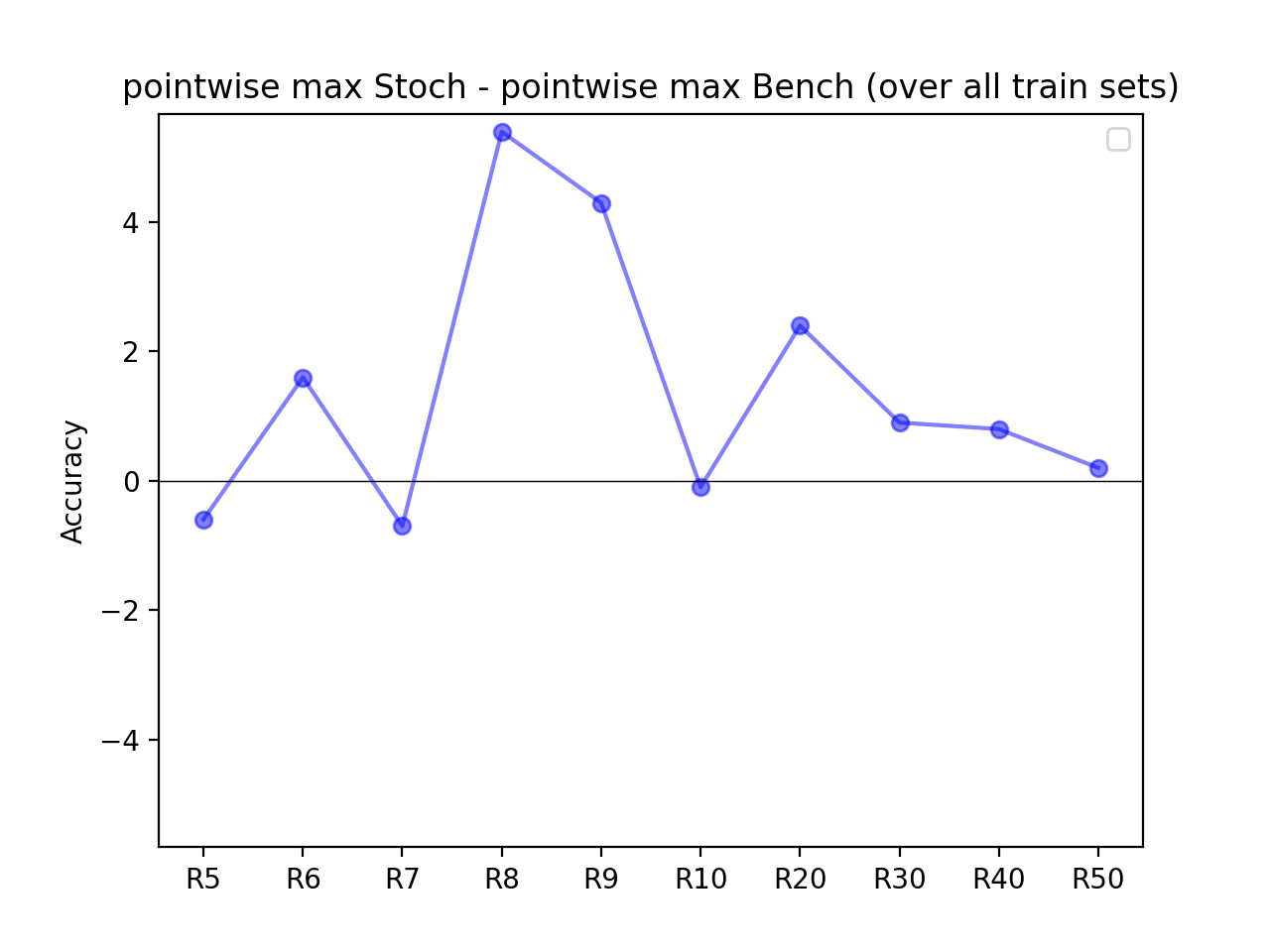}}%
      }
      \caption{Aggregated  accuracy curves for the CONN and the baseline
classifiers, along with their difference.        
      The x-axis tick values correspond to the size of the training databases, as in \figureref{fig:hconn_results}.
      (a): Pointwise maxima of the accuracy functions 
from \figureref{fig:hconn_results}      
      for the stochastic CONN classifier (in blue), the vanilla CONN classifier
      (in green), and the baseline classifier (in red) across the number of training epochs: 25, 50, 100, and 200.
      (b): Difference between the
pointwise maxima functions
for  the stochastic CONN classifier
      and the baseline classifier, shown in (a)
      }
      \label{fig:hconn_results_comulat}
    \end{figure}

The obtained results 
for the baseline, the vanilla and the stochastic CONN
classifiers are
shown in \figureref{fig:hconn_results}.
In \figureref{fig:image-a7}
the maximum accuracy scores over the 4 numbers of 
training epochs for the classifiers are shown.
In \figureref{fig:image-b7} the 
difference between 
the accuracy values of the stochastic and 
baseline classifiers is shown.

\begin{figure}[h!]
\centering
  {\includegraphics[width=0.65\linewidth]{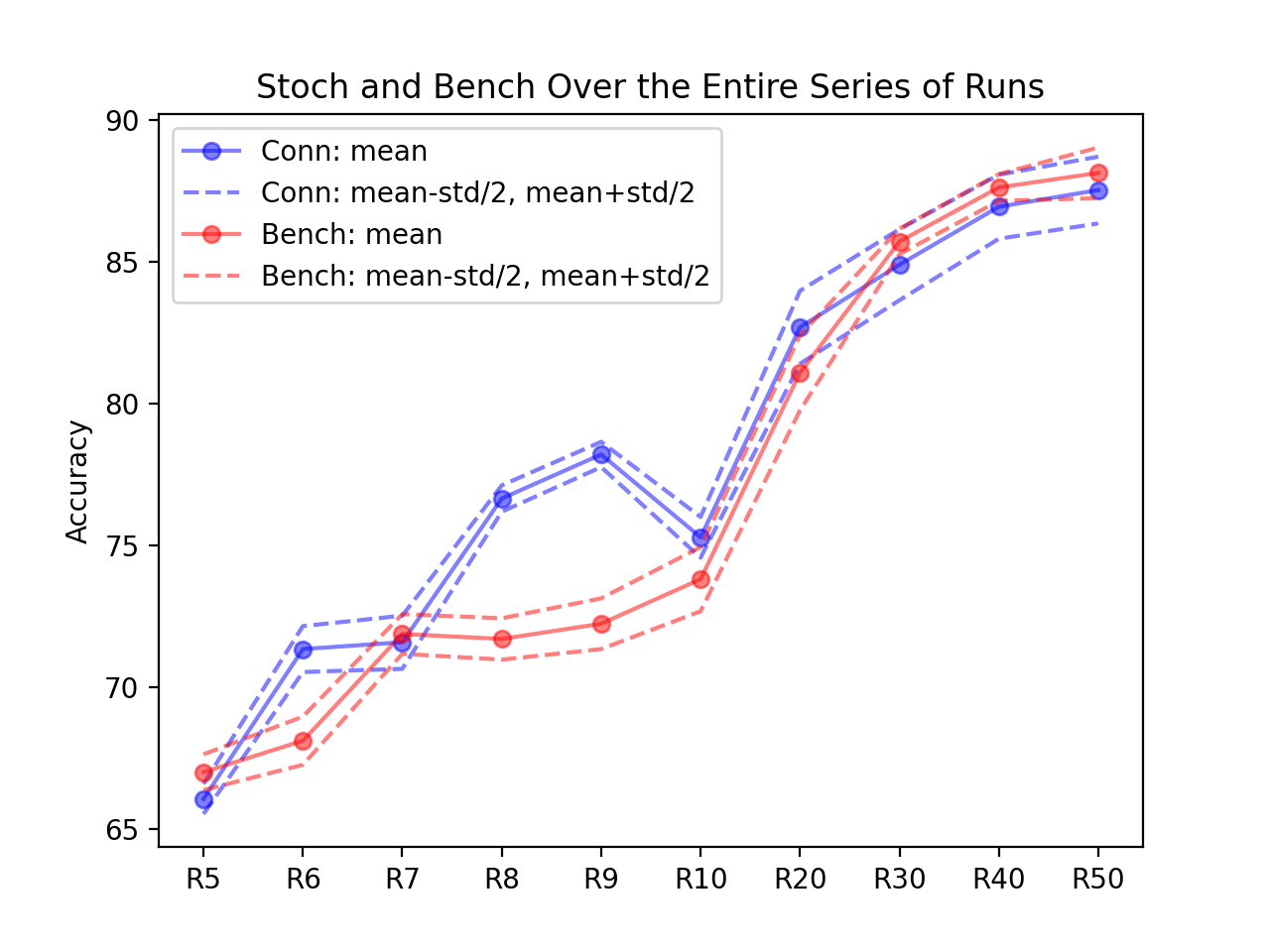}}
    \caption{Mean and standard deviation curves
for the
stochastic CONN classifier 
(in blue) and
the baseline classifier
(in red) 
 calculated from 100 maximum accuracy score curves for these classifiers, analogous to those depicted in \figureref{fig:image-a7} 
    }
  \label{fig:hconn_results_series_comulat}
\end{figure}

Furthermore, our experiments with the CONN classifiers were extended to reflect a certain dependence of the obtained accuracies on setting the seeds for random number generation.\footnote{In NumPy and PyTorch environments.} For each training configuration discussed above, we conducted 100 training sessions with randomly selected seeds for random number generation. This provided us with 100 maximum accuracy score curves, similar to those shown in
 \figureref{fig:image-a7}. The obtained mean and standard deviation curves, shown in \figureref{fig:hconn_results_series_comulat}, demonstrate  demonstrate the superior accuracy of the stochastic classifier compared to the baseline.

The source code of our experiments is available at \citet{conn_repo}.

\section{Discussion}
\label{sec:Disc}

The CONN model describes communication between persons, where participants receive information in 
an external 
communication loop
and process it 
using internal communication loops.
Additionally, the participants are partially or fully aware
of the received information and exchange this perceived
information with each other in the external loop.
The model is structured as a sequence of observed-to-seen
operations and may employ subject-associated autoencoders
for the implementation.

In a wide sense we may consider our model as decision-make. The model is composed of internal and external phases and can cope both short and prolonged decision-making processes. The internal process is iterative and an inaccurate decision (but still valid) may result if the number of iterations is too small. Yet a valid decision can be returned at any time (iteration). This process can thus incorporate both fast and long decision-making procedures and can explain both reflexes and regular decisions, under the same procedure.

Our work addresses the perception of one person (internal perception) and communication between two persons, but this model can be extended to involve more than two persons. Additionally, it is not limited to persons. The work may be applied to any system that involves processing from "latent" to "raw" representations.

Under our model, the flow of information involved in perceiving an object by a person converges to a fixed point, which can be treated as a single-element cycle. This convergence characterizes the awareness of perceiving an object.
Similarly, in the two-person communication model, we have experimentally observed and proven, under certain natural conditions, that the modeled flow of information between the participants exhibits the property of converging to a bipartite cycle
(\theoremref{thr}). In this sense, the bipartite orbits,
when considered as a whole, can be seen as the
"attractors of interpersonal communication",
representing what can be referred to as the "collective
consciousness" within this communication.

In cognitive science, perceptual inference is considered the brain's process of interpreting sensory
information by combining predictive processing,
Bayesian inference, top-down and bottom-up processing, and contextual cues to resolve
ambiguities and make sense of the environment.
It enables us to recognize objects and understand scenes by integrating prior knowledge and expectations with sensory data, ensuring coherent perception despite noisy and ambiguous inputs.

Our observed-to-seen functional model allows us to simulate some aspects of perceptual inference. The construction of the "percept" image via attractor basins provides a method for resolving ambiguity, potentially reducing noise and enhancing perceptual clarity. However, we do not claim that the internal representations are necessarily the "correct" representations. For example, the percept images from the middle and the right column in \figureref{fig:obs_seen_both_test}
 do not coincide with the ground truth images from the first column.

Furthermore, the CONNs simulate perceptual awareness in two aspects. Firstly, they model
the observed/seen functionality of the  visual perceptual awareness (\sectionref{sec:rikuz:semiotics:person-object}).
Additionally, they emulate the phenomenon of multistable human perception, which is elicited by ambiguous images such as the Rubin face-vase (\citet{RubinFaceVase}). As discussed in \sectionref{sec:atts_classifier_multi},
stochastic CONN classifier specifically emulates
the properties of consistency and predefinency
observed in human multistable perception.
On the other hand, the importance of multistable perception for perceptual awareness has long been recognized (\citet{Leopold_multistable, Lumer}).
Recent neuroscience research  establishes a connection between multistable phenomenon and perceptual awareness, suggesting that multistability can play a crucial role in understanding the  process of perceptual inference (\citet{Saracini}). Thus, CONNs
mimics multistable perception, which is recognized as essential for awareness. This represents the second aspect of CONN's functionality in simulating awareness.

The consistency and predefinency of human perception in interpreting ambiguous visual stimuli mentioned above
reflects the robustness and generalization abilities of the human visual system.
Another manifestation of these
abilities is resilience to adversarial attacks.
It is widely acknowledged that human perception exhibits greater resilience against adversarial
attacks compared to neural networks
(for example, \citet{Adversarial_examples_in_the_physical_world, limitations_deep_learning}).
Are the CONN classifiers, which mimic certain properties of human perception, also resilient to adversarial attacks?

We explore this question in \citet{SI}~N. 
There we provide a rationale for the assumption that vanilla CONN classifiers, trained 
on small datasets of examples with sufficiently large distances between the
examples, 
possess
intrinsic resilience to perturbation attacks. We show
that the perceptual layer
hinders the 
attacks within the basins of the attractors associated with
the training example. 

Concerning the stochastic CONN classifier, one may notice
that it possesses
additional defensive measures
such as ensembling (see \citet{adv_autoenc, adv_Lin}) and introducing augmentation noise during both the training and testing phases (see \citet{adv_noise_train_only,
adv_Lin,adv_noise_train_test}).

Our ongoing research focuses on exploring and assessing the resilience of CONN classifiers against various adversarial attacks. Additionally, while our current analysis uses the MNIST database, future work will extend to other datasets.

\acks{
We are grateful to Victor Halperin, Andres Luure, and Michael Bialy for their valuable contributions. We also acknowledge the Pixabay image collection (\citet{pixabay}) for the images used in this paper.
}

\bibliographystyle{unsrtnat}
\bibliography{isn_arxiv} 

\end{document}


\title{Supplementary Information for \,
\newline
 Semiotics Networks Representing  Perceptual Inference}

\author{
\textnormal{David Kupeev}\thanks{Corresponding author}\\
Independent Researcher, Israel\\
kupeev@gmail.com \\ \\
\and 
Eyal Nitzany\\
Independent Researcher, Israel\\
eyalni@gmail.com \\ \\
}

\maketitle

\section{Glossary}
\label{SI:Glossary}

\begin{itemize}

\item \textbf{Semiotic Networks:} These are computational structures designed to simulate the processes of human perception and communication, particularly focusing on how external objects are observed and processed into internal representations.

\item \textbf{Perceptual Inference:} A method
that
simulate the interpretation and processing of perceived objects, transforming external stimuli into internal representations.
This method can be used by human or
semiotics networks.

\item \textbf{Observed vs. Seen:} The terms "observed" and "seen" represent two stages of perception. "Observed" refers to the initial sensory input, while "seen" is the processed version of that input, reflecting a deeper interpretation or internal transformation. These two steps are cyclically repeated until convergence is achieved or a predefined stop condition is met, simulating the continuous refinement of perception in the network.

\item \textbf{Perceptualized Classifier:} A neural network model that includes a perceptual layer. This layer simulates the process of perception in the network, helping improve the classifier's ability to interpret and handle input data, especially with limited training datasets.

\item \textbf{Consciousness Networks (CONN):} A novel model introduced in the paper that represents the flow of information between agents (for example, people or systems) during communication. It captures both internal processing loops and external data exchanges, reflecting how individuals perceive and interpret shared information.

\item \textbf{Attractor-based Classifier:} A classification method using attractor dynamics, where the network iteratively processes an input until it converges to a stable representation (the attractors), which aids in the final classification.

\item \textbf{Bipartite Orbits:} Describes the asymptotic behavior of sequences of exchanged information between communicating agents. It involves periodic sequences of representations as participants converge on stable shared interpretations.

\item \textbf{Awareness Property:} A concept introduced in the paper that links the idea of a perceived object with the internal representation reaching a fixed point in the network, marking the subject's conscious awareness of an object.

\item \textbf{Stochastic Classifier:} A classification model that uses random augmentations of input data to build multiple attractors, enhancing the classifier's robustness and accuracy, especially in tasks involving limited data.

\item \textbf{Autoencoder:} A type of neural network used for unsupervised learning, where the network is trained to encode the input into a compressed latent space and then decode it back to the original space, often used for tasks such as dimensionality reduction and feature learning.

\item \textbf{Latent Space:} The internal representation of data in a reduced-dimensionality space, often generated by autoencoders during the process of encoding input data.

\end{itemize}

\section{Multy-Modal CONN}
\label{SI:Multy-Modal CONN}

\begin{figure}[H]
\centering
    {\fbox{\includegraphics[width=1.0\linewidth]{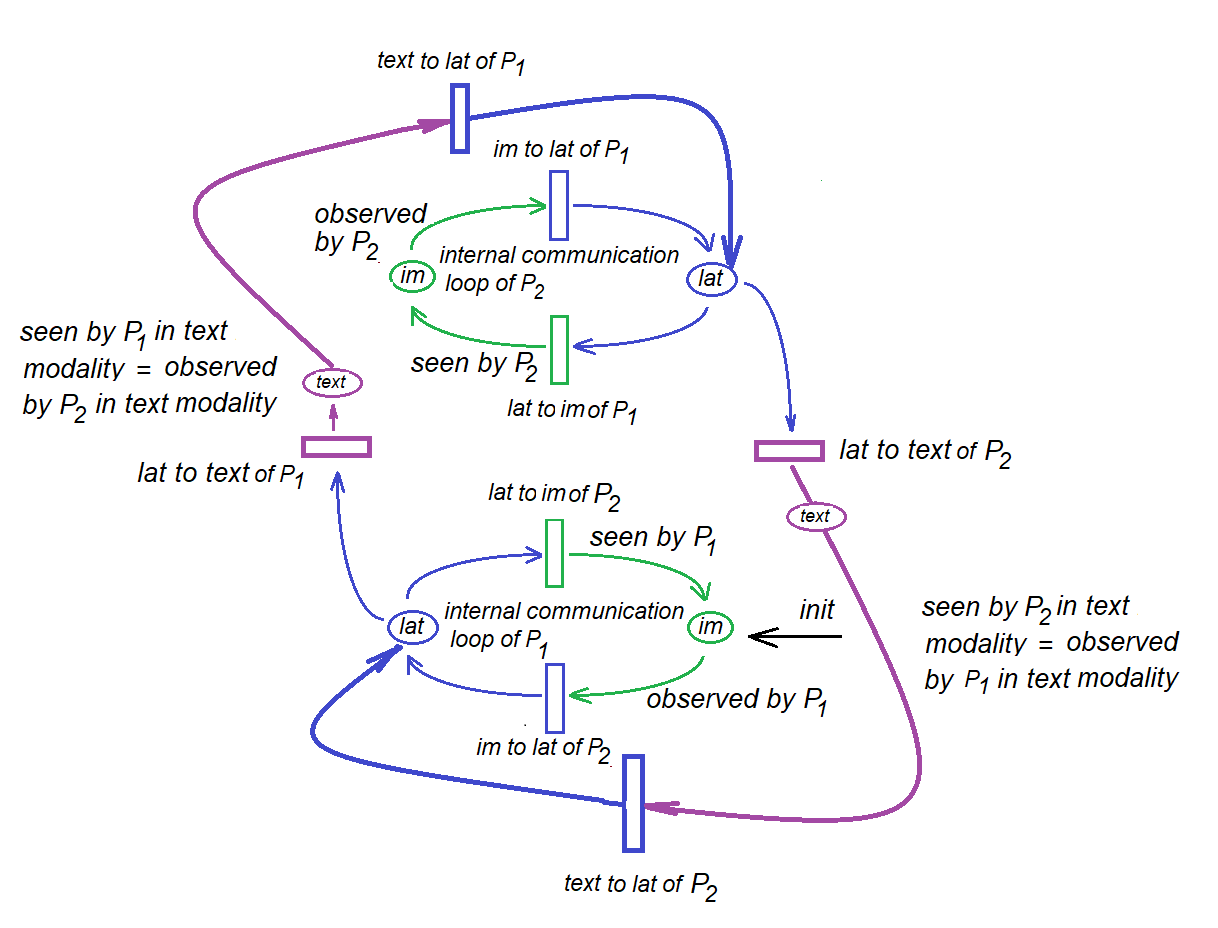}}}
\caption{
The hypothetical CONN operating with two raw modalities. The internal communication loop operates with image modality (shown in green), while the external communication loop operates with  text modality (shown in violet). The dataflow of the latent modality is shown in
blue\protect\footnotemark
}
    \label{fig:scheme_full_b2_textual}
  \end{figure}

The Consciousness Networks (CONNs)
could  potentially operate with several raw modalities. For instance, a hypothetical CONN, as illustrated in \figureref{fig:scheme_full_b2_textual}, operates with two raw modalities: image modality for the internal communication loop and text modality for the external communication loop. The latent modality of the loops is represented by a joint latent space for text and image embeddings (\citet{clip}).


\addtocounter{footnote}{-1}
\footnotetext{The figures in SI are best viewed in color.}

\begin{figure}[H]
    \centering
    {\fbox{\includegraphics[width=0.8\linewidth]{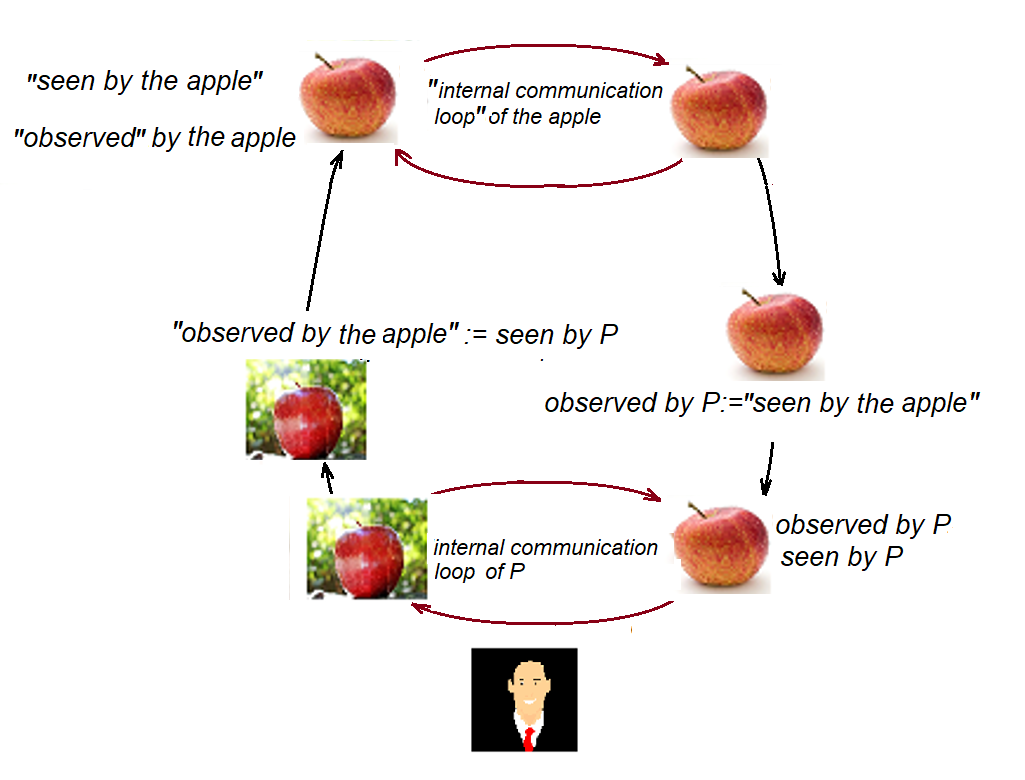}}}
\caption{The CONN model for inter-person communication, representing the process of perceiving an object by a person}
\label{fig:apple_as_face2}
  \end{figure}

\section{Modeling Object Perception as a Form of Communication}
\label{SI:Modeling}

Below, we will model the interaction between a person and an object as a  specific case of CONN modeling, which was introduced for person-to-person communication in \citet{isn}.
In \figureref{fig:apple_as_face2}, we illustrate the perception of an object (an apple) by a person.
The "observed-to-seen"
CONN transformation associated with the apple does not depend on the input and always returns the same image of the apple, which is then further observed by a person.
Further, as in the
person-to-person  CONN
the image becomes the observed input for the person and is further personalized in the observed-to-seen transformations, altering the image.

\section{The Sequences of Images Generated by CONN Implemented Using Autoencoders}
\label{SI:W_enc_dec}

\begin{algorithm}[htbp]
  \centering
  \caption{A conscious  neural network for  communication between two persons.
The network is comprised
of autoencoders $A_{P_1}$ and
$A_{P_2}$ associates with
persons $P_1$ and $P_2$
	\newline
	{\bf Input}:
	An image $Im_1$ which is related to person
	$P_1$
	\newline
	{\bf Output}:  A sequence of interchange images
	$Im_1, Im_2, \dots Im_k, \dots$
	}
	{
    \begin{enumerate}
        \item
        \label{nitm:1}
            Set $iter=1$;         set $person\_id=1$

        \item
        \label{nitm:2.5}
        Initialize the output queue to an empty list

        \item
        \label{nitm:3}
        While $iter \leq  n_{iters}$ do:
        \begin{enumerate}
        \item
        \label{nitm:3.1}
        Use $person\_id$ parameters ($nsteps_{person\_id}$)
        \item
        Perform
        $nsteps_{person\_id}$
        encoding/decoding iterations
        (\equationref{eq:F_P_i})
        of the autoencoder associated
        with person $person\_id$ on
         $Im_{iter}$
        to receive the image representation
        (current $Im$).
        \label{nitm:3.2}

        \item
        \label{nitm:3.3}
        Decode the previous
        encoding result
        $lat$
        ($\text{current } Im=dec(lat)$) to
        receive $Im_{iter+1}$
        ($Im_{iter+1} = \text{current } Im$
        after this operation)

        \item
        Increase $iter$ by 1 and
        change $person\_id$ to other $person\_id$
        \label{nitm:3.4}

        \item
        Send $Im_{iter}$ to the updated person
        and add to the output queue
        \label{nitm:3.5}
        \end{enumerate}

        \item
        \label{nitm:4}

        Return the output queue

    \end{enumerate}
    }
    \label{SI:alg:CONN}
\end{algorithm}

Algorithm\,1 in \citet{isn}  represents the implementation of the CONN using encoder/decoder operations of the autoencoders. Below, we describe the sequences of the images generated by the algorithm.

Let $S_{iter,P_{od(iter)}}$ be a sequence of images generated in
step\,\ref{nitm:3.2} of the algorithm:
\begin{equation*}
\begin{split}
&S_{iter,P_{od(iter)}} =
Im_{iter}, \,
[dec(enc)]^{1}(Im_{iter}), \,
\dots ,
[dec(enc)]^{2}(Im_{iter}), \,
\dots ,
\\
&\;\;\;\;[dec(enc)]^{nsteps_{od(iter)}}(Im_{iter})=Im_{iter+1}
\;,
\end{split}
\end{equation*}
where $dec$ and $enc$ are encoding and decoding
operations of $P_{od(iter)}$,
$iter = 1,\dots,
n_ {iters}$, and
\begin{equation}
  \label{eq:i_iter}
    od(iter)=
    \begin{cases}
      1, & \text{if}\ iter \text{ is odd} \\
      2, & \text{if}\ iter \text{ is even,}
    \end{cases}
  \end{equation}
and every member of the sequence is obtained from the previous one by applying a composition of decoder and encoder operations.
The $iter$
refers to "external"
iteration and $od(iter)$
to the person's index.

For example,
$S_{5,P_1}$ refers to
the sequence of images
in the fifth interchange
of Person one.

The entire sequence of images
generated by the algorithm is represented as
	\begin{equation*}
\begin{aligned}
	W(Im)=
	\operatorname{concat}(
S_{1,P_1}, \, S_{2,P_2}, \, S_{3,P_1},
	\, S_{4,P_2},
	 \dots)
	 \;,
\end{aligned}
	\end{equation*}
where $Im=Im_1$.

The sequence $U$ of the images
$(Im_{iter})$
transmitted in
inter-person
communication
is represented as:
\begin{equation*}
\begin{split}
U(Im,nsteps_1,nsteps_2)=
Im_1
\xrightarrow{  [F_{P_1}]^{nsteps_1} }
Im_2
\xrightarrow{  [F_{P_2}]^{nsteps_2} }
\\
Im_3
\xrightarrow{  [F_{P_1}]^{nsteps_1} }
Im_4
\xrightarrow{  [F_{P_2}]^{nsteps_2} }
Im_5
\xrightarrow{  [F_{P_1}]^{nsteps_1} }
\dots
\;\;,
\end{split}
\end{equation*}
where $Im_1=Im$, and
\begin{equation}
\label{eq:F_P_i}
F_{P_i} \text{ is a composition
of decoder and encoder
operations applied to }
P_i.
\end{equation}

\section{Example of Partitioning the Sequence of the Images Transmitted Between the Persons}
\label{SI:part_of_U}

Let the sequence $U$ of the images transmitted between the persons
is represented  as

\begin{equation}
\label{eq:U0}
\begin{split}
U(Im,nsteps_1,nsteps_2)=
Im
\xrightarrow{  [F_{P_1}]^{nsteps_1} }
T_{1,P_1}
\xrightarrow{  [F_{P_2}]^{nsteps_2} }
\\
T_{2,P_2} \xrightarrow{  [F_{P_1}]^{nsteps_1} }
T_{3,P_1}
\xrightarrow{  [F_{P_2}]^{nsteps_2} }
T_{4,P_2}
\xrightarrow{  [F_{P_1}]^{nsteps_1} }
\dots
\;.
\end{split}
\end{equation}

We can partition  the first $1+(m+1)\times K$ members of  $U$
by the first element of $U$
($Im$), followed by the matrix of
$(m+1)$ rows and $K$ columns.
The partitioning proceeds similarly to
function {\em reshape} of Python.
In our example, $K=2$ and
$m=2$. Using the Python notation for indexing
the sequence, the partitioning
may be written as:
\begin{equation*}
\begin{array}{@{}cccc@{}}
\multicolumn{4}{l}{U \,
[0: 1+(m+1)\times K +1 ]
= Im \xrightarrow{F_1} }
\\[0.3em]
 T_{1+0\cdot K,P_1} &\xrightarrow{F_2} & T_{1+0\cdot K+K-1,P_2} &
            \xrightarrow{F_1}\\

 T_{1+1\cdot K,P_1} &\xrightarrow{F_2} & T_{1+1\cdot K +K-1,P_2} &
            \xrightarrow{F_1}\\

\dots &&& \\

 T_{1+m\cdot K,P_1} &\xrightarrow{F_2} & T_{1+m\cdot K+K-1,P_2} \; .&
            \\
\end{array}
\end{equation*}

\section{An Implementation
of CONN 
Not Based on Autoencoders}
\label{SI:1}

In this section, we describe a simple implementation of a CONN. The "observed-to-seen" transformations of the network are modeled as non-continuous operators in a 2-dimensional space, unlike the CONNs considered in \citet{isn}.

Let $N_1$ and $N_2$ be given numbers of "attractors"
associated with persons
$P_1$ and $P_2$, respectively.
We first define the "basins of attractors" and then the "attractors" itself.
Let $R$ be a rectangle $ [0,1] \times [0,1]$.

We randomly select
$N_1-1$ numbers $0 < a_1 < \dots < a_{N_1-1} < 1$ and
partition $R$
by $N_1$ basins of attractors
$R_1^i$
for $P_1$:

\begin{equation*}
\begin{split}
&R_1^1 = [a_0, a_1) \times [0,1], \\
&R_1^2 = [a_1, a_2) \times [0,1], \\
&\dots \\
&R_1^{N_1-1} = [a_{N_1-2}, a_{N_1-1}) \times [0,1], \\
&R_1^{N_1} = [a_{N_1-1}, 1] \times [0,1].
\end{split}
\end{equation*}
Then randomly select
a point in
each
$R_1^i$, the point will serve as
the attractor of
the  basin.

Similarly, select
$N_2-1$ numbers $0 < b_1 < \dots < b_{N_2-1} < 1$ and
partition $R$
by $N_2$ basins of attractors for $P_2$:
\begin{equation*}
\begin{split}
&R_2^1 = [0,1] \times [b_0, b_1)    , \\
&R_2^2 = [0,1] \times [b_1, b_2)    , \\
&\dots \\
&R_2^{N_2-1} = [0,1] \times [b_{N_2-2}, b_{N_2-1}) , \\
&R_2^{N_2} = [0,1] \times [b_{N_2-1}, 1] .
\end{split}
\end{equation*}
Then randomly select an attractor in
each basin $R_2^i$.

\begin{figure}[]
\centering
    {\fbox{\includegraphics[width=0.45\linewidth]{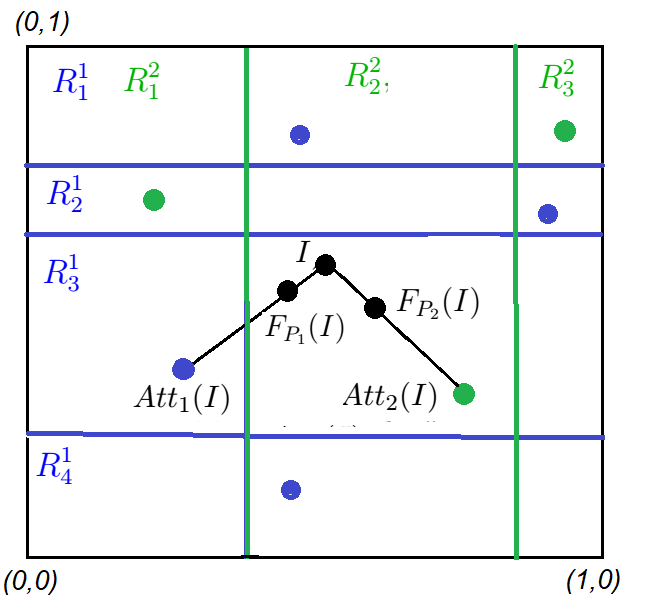}}}
\caption{
Illustration of the non-autoencoder-based implementation of CONN.
Basins of "attractors" of person
$P_1$ (resp. $P_2$) are  the
rows (resp. columns) of rectangle
$R =  [0,1] \times [0,1]$.
"Attractors" associated with persons $P_1$ and $P_2$ are denoted by blue and green circles, respectively}
    \label{fig:non_autoencpng}
  \end{figure}

For a point $I \in R$
and $i=1,2$,
define
$Att_{i} (I)$ as the
attractor
of the basin $R_i^j$,
such that $I \in R_i^j$.

Also, for a $k$ with $0<k<1$,
define
the
"observed-to-seen"
transformation
of $I$
as the shift of $I$ to
the point
$F_{P_{i}}(I)$
such that
\begin{equation}
\label{eq:F_P for simpliofiied}
        \overrightarrow{(F_{P_i}(I),Att_{i} (I) )} =
        k \;
        \overrightarrow{(I, Att_{i} (I)   )}.
\end{equation}
The transformation
shifts $I$ in the
direction  of
attractor $Att_{i} (I)$.
See \figureref{fig:non_autoencpng},
which illustrates the construction of
$F_{P_i}$.

Note that $F_{P_{1}}$ and $F_{P_{1}}$ are not continuous
(unlike the autoencoder functions
in \citet{isn}).
Also, in the discussed model,
every $I \in R$ belongs to a certain basin for
$P_1$ and $P_2$.\footnote{Unlike to
the scenario
for continuous
$F_{P_{i}}$
depicted in \figureref{fig:uniform}.
}

Our implementation of this CONN is described in \algorithmref{algo:isn2}.

\begin{algorithm}[]
\centering
	\caption{The
	CONN not Based on Autoencoders
\newline
	{\bf Parameters:}
	$0 < k < 1$, $nsteps_{1}$, and
	$nsteps_{2}$
\newline
	{\bf Input}:
	A point
	$I$ in $R = [0,1]\times [0,1]$
which is related to person
	$P_1$
\newline
	{\bf Output}:  
	A sequence of interchange
	points
	$I_1, I_2, \dots I_k, \dots$.
		}
{
    \begin{enumerate}
        \item
        \label{nitm:1 isn2}
            Set $iter=1$, $I_1=I$,
            $person_{id}=1$

        \item
        \label{nitm:2.5 isn2}
        Initialize the output queue to
        an empty list

        \item
        \label{nitm:3 isn2}
        While $iter \leq n_{iters}$ do:
        \begin{enumerate}
        \item
		\label{nitm:3.4 isn2}
        Construct
        $I_{iter+1}
        =
        F_{P_i}^{[nsteps_{i}]}(I_{iter})$,
        where $i=person_{id}$
        \item
        \label{nitm:3.4.5 isn2}
        Increase $iter$ by 1 and
        change $person_{id}$ to other $person_{id}$

        \item
         \label{nitm:3.4.6 isn2}
        Add
        $I_{iter}$
        to the output queue
        \end{enumerate}
        \item
    \label{nitm:4 isn2}
    Return the output queue
    \end{enumerate}
}
\label{algo:isn2}
\end{algorithm}
In \ref{nitm:3.4 isn2} of the algorithm,
$F_{P_i}^{[nsteps_{i}]}(I)$
denotes
$nsteps_{i}$ compositions of
operator $F_{P_i}$
(determined in \equationref{eq:F_P for simpliofiied}).

The definitions of the first and second type bipartite orbits
in \citet{isn} are
easily transferred to the sequences of interchange points
output by the algorithm.

In our experiments with
the implementation of
 \algorithmref{algo:isn2}, we varied
parameters
$N_1, N_2, k, nsteps_1$, $nsteps_2$,
and
initial point $I$.
For each configuration, we observed convergence to the first type orbits: starting from a certain
number the sequence of  interchange points
transmitted between $P_1$
and $P_2$ becomes cyclic.
For $nsteps_1$ and
$nsteps_2$   greater than 20,
the sequence
did not depend on the values
of $nsteps_1$ and
$nsteps_2$,
thus demonstrating
convergence to
the second type orbits.

\section{Lemma Concerning Bipartite Orbits of the First Type}
\label{SI:Lemma_bipart_1_type}

Let $X$ be a complete metric space with distance function $d$.
For the first type
bipartite orbits
defined for
continuous
functions
$F_{P_1}$ and $F_{P_2}$ from $X$ to $X$
(\citet{isn}, Section 4.2),
the
following lemma holds.
\begin{lemma}
\label{SI:lemma_att1}
The elements of a bipartite orbit
$(b_0,b_1,\dots,b_{K-1})$
of the first type
satisfy the following properties:
    \begin{enumerate}

        \item
        \label{att1:1}

They form a loop with respect to alternating $F_1$ and $F_2$ operations:
\begin{equation}
\label{eq:att1}
b_0
\xrightarrow{  F_2 }
b_1
\xrightarrow{  F_1 }
\dots
\xrightarrow{  F_2 }
b_{K-1}
\xrightarrow{  F_1 }
b_0
\;,
\end{equation}
where
\begin{equation}
\label{eq:F1F2}
\begin{aligned}
&F_1 = [F_{P_1}]^{nsteps_1},\\
&F_2 = [F_{P_2}]^{nsteps_2}
,
\end{aligned}
\end{equation}

        \item
        \label{att1:2}

They are alternating fixed points for the functions $G_1$ and $G_2$:

\begin{equation}
\label{SI:eq:G1G2forFixed}
\begin{aligned}
&G_1(b_h) = b_h, \text{ for even } h, \\
&G_2(b_h) = b_h, \text{ for odd } h\,,
\end{aligned}
\end{equation}

where

\begin{equation*}
\begin{gathered}
G_1 = [\,F_1 \cdot F_2\,]^{K/2},
 \\
G_2 = [\,F_2 \cdot F_1\,]^{K/2}\,.
\end{gathered}
\end{equation*}

\end{enumerate}

\begin{proof}

\equationref{SI:eq:G1G2forFixed}
follows from
\equationref{eq:att1}
thus it is sufficient to prove the latter.

Consider the bipartite sequence of the first type converging to an orbit $(b_0, b_1, \dots, b_{K-1})$
(\citet{isn}, Equation (14)):

\begin{equation*}
\begin{array}{@{}cccccccc@{}}
\multicolumn{4}{l}{U(Im,nsteps_1,nsteps_2)= Im \xrightarrow{F_1} } &&&&
\\
T_{1+0\cdot K,P_1} &\xrightarrow{F_2} & T_{1+0\cdot K+1,P_2} &
        \xrightarrow{F_1}  & \dots  &\xrightarrow{F_2}  & T_{1+0 \cdot K +K-1 ,P_2}  &
        \xrightarrow{F_1}
\vspace{.2cm}
\\
G_1 \downarrow & & G_2 \downarrow &&&& G_2 \downarrow &
\\
T_{1+1\cdot K,P_1} &\xrightarrow{F_2} & T_{1+1\cdot K+1,P_2} &
        \xrightarrow{F_1}  & \dots &\xrightarrow{F_2} & T_{1+1 \cdot K +K-1,P_2}  &
        \xrightarrow{F_1}
\vspace{.2cm}
\\
G_1 \downarrow & & G_2 \downarrow &&&& G_2 \downarrow &
\\
\dots & & \dots &&&& \dots &
\\
G_1 \downarrow & & G_2 \downarrow &&&& G_2 \downarrow &
\\
T_{1+m\cdot K,P_1} &\xrightarrow{F_2} & T_{1+m\cdot K+1,P_2} &
        \xrightarrow{F_1}  & \dots &\xrightarrow{F_2} & T_{1 + m \cdot K +K-1,P_2} &
        \xrightarrow{F_1}
\vspace{.2cm}
\\
G_1 \downarrow & & G_2 \downarrow &&&& G_2 \downarrow &
\\
\dots & & \dots &&&& \dots &
\\
b_0            & & b_1            &&&& b_{K-1}        &
\\
.
\end{array}
\end{equation*}

Rewrite
this sequence in a more convenient
for our proof form:

\begin{equation}
\label{eq:matrix_mh}
\begin{array}{@{}cccccccc@{}}
\multicolumn{5}{l}{U(Im,nsteps_1,nsteps_2)= Im \xrightarrow{F_1} } &&&
\vspace{.3cm}
\\
R_{0,0} &\xrightarrow{F_2} & R_{0,1} &
        \xrightarrow{F_1}  & \dots  &\xrightarrow{F_2}  & R_{0,K-1}  &
        \xrightarrow{F_1}
\vspace{.2cm}
\\
G_1 \downarrow & & G_2 \downarrow &&&& G_2 \downarrow &
\\
R_{1,0} &\xrightarrow{F_2} & R_{1,1} &
        \xrightarrow{F_1}  & \dots &\xrightarrow{F_2} & R_{1,K-1}  &
        \xrightarrow{F_1}
\vspace{.2cm}
\\
G_1 \downarrow & & G_2 \downarrow &&&& G_2 \downarrow &
\\
\dots & & \dots &&&& \dots &
\\
G_1 \downarrow & & G_2 \downarrow &&&& G_2 \downarrow &
\\
R_{m,0} &\xrightarrow{F_2} & R_{m,1} &
        \xrightarrow{F_1}  & \dots &\xrightarrow{F_2} & R_{m,K-1} &
        \xrightarrow{F_1}
\vspace{.2cm}
\\
G_1 \downarrow & & G_2 \downarrow &&&& G_2 \downarrow &
\\
\dots & & \dots &&&& \dots &
\\
b_0            & & b_1            &&&& b_{K-1}        &
\\
.
\end{array}
\end{equation}

Let us show that
\begin{equation}
\label{eq:att1_FIRST}
  F_2(b_0)=
b_1
.
\end{equation}

Denote
$F_2(b_0)$ by $z$ and assume
$z\neq b_1$.
Select an arbitrary $\epsilon > 0$.
Since $F_2$ is continuous,
there exist $\delta >0$
such that if $d( R_{m,0},b_0) \leq
\delta$ then
\begin{equation*}
d( R_{m,1},z) \leq
\epsilon .
\end{equation*}

Since $( R_{m,0} )$
converges to $b_0$,
the latter equation holds
for $m$  greater than some
$M_0$.
On the other hand,
by the convergence of $( R_{m,1} )$
to $b_1$,
for $m$ greater than some $M_1$, it holds that
\begin{equation*}
d( R_{m,1},b_1) \leq
\epsilon .
\end{equation*}

Thus,
for sufficiently large $m$
$d(z,b_1)\leq 2\epsilon$.
This contradicts  the
assumption
$z\neq b_1$ and thus
\equationref{eq:att1_FIRST} holds.
Similarly,
considering the respective
columns of the
matrix representation of
\equationref{eq:matrix_mh},
the entire
\equationref{eq:att1}
is proven.
\end{proof}
\end{lemma}

\section{Theorem  Concerning Bipartite Orbits of the Second Type}
\label{SI:Lemma_bipart_2_type}

Let $X$ be a complete metric space with distance function $d$.
The definitions of attractors, fixed points, and basins, as provided for  Euclidean space by \citet{Belkin}, are applicable to $X$, and we will adopt them in the following.

Let
$r=1,2$. Let $F_{P_r}$ be a continuous function $X \rightarrow X$.
Let $\mathcal{A}_r$ be a subset
of the set of attractors of $F_{P_r}$.

Define the function $\widehat{F_r}: X \rightarrow X$ as follows:
 \begin{equation}
\label{eq:F12_widehat}
\widehat{F_r}(x) = \lim_{n\to\infty} [F_{P_r}]^n(x),
\end{equation}
if the limit exists and is an
attractor for
$F_{P_r}$.

Define  functions $\alpha$ as
  \begin{equation*}
    \alpha (r)=
    \begin{cases}
      2, & \text{if } r=1 \\
      1, & \text{if } r=2,
    \end{cases}
  \end{equation*}
and
$od$ as in \equationref{eq:i_iter}:
  \begin{equation*}
    od(iter)=
    \begin{cases}
      1, & \text{if } iter \text{ is odd} \\
      2, & \text{if } iter \text{ is even}.
    \end{cases}
  \end{equation*}

For $x\in X$
and
$\epsilon>0$
$B_{\epsilon}(x)$
denotes
the open ball
\begin{equation*}
B_{\epsilon}(x) =
\{
x'
\;| \;
 d(x,x') < \epsilon
\}
.
\end{equation*}
For a function
$f: X \rightarrow X$,
$f(B_\epsilon(x))$ denotes
the image of the $\epsilon$-ball under $f$.

We will need the
following lemma:
\begin{lemma}

  \label{lemma_theorem_L_A}

Let $F$: $X \rightarrow X$
be a continuous function
defined at every $x\in X$.
If the following conditions hold:

\begin{enumerate}[(a)]

\item
\label{l_A2-new}
A point $a \in X$ belongs to  basin $\mathcal B(a')$ of some attractor $a' $ of $F$
together
together
with an open ball of a certain radius
${\delta}(a)$ around $a$:
$B_{\delta (a)}(a) \subset
\mathcal B(a')
$.

\item
\label{l_A2.5-new}
Convergence of the sequence
of functions $(F^{[n]})$
to $a'$ is locally uniform at $a'$:
there exists $\delta (a') > 0$ such that for any $\epsilon > 0$, there exists $n_0$ such that
\begin{equation*}
F^{[n]}(B_{\delta(a')}(a')) \subseteq B_{\epsilon}(a')
\end{equation*}
for any $n \geq  n_0$.
See \figureref{fig:uniform}.

  \end{enumerate}

Then
the convergence
of the sequence
of functions
$(F^{[n]})$
to $a'$ is locally uniform at $a$:
there exists $\lambda (a) > 0$ such that for any $\epsilon > 0$, there exists $n_0$ such that
\begin{equation}
\label{eq:lambda_atfar}
F^{[n]}( B_{\lambda(a)}(a)) \subseteq B_{\epsilon}(a')
\end{equation}
for any $n \geq  n_0$.

\begin{proof}

Let $B_{\delta}(a')$ be an open ball
around $a'$  where
uniform
convergence
of the sequence $(F^{[n]})$
to $a'$
holds.
Select, by \ref{l_A2-new},
an $N>0$ such that
$F^{[N]}(a)$
falls to $B_{\delta}(a')$.
Select
$\delta_2 < \delta -
d(a',F^{[N]}(a))$,
$\delta _2>0$.
By the triangle inequality,
\[
B_{\delta _2}(
   F^{[N]} (a)
   )
\subseteq B_{\delta}(a')
.
\]

Since
$F^{[N]}$ is continuous,
for a certain
$\lambda>0$,
\[
F^{[N]} (B_{\lambda}(a))
\subseteq
B_{\delta _2}(
   F^{[N]} (a)
   )
.
\]

Therefore,
\begin{equation}
\label{eq:Nsofsof}
F^{[N]} (B_{\lambda}(a))
\subseteq
B_{\delta}(a')
.
\end{equation}

For an arbitrary $\epsilon > 0$,
using the definition of
$B_{\delta}(a')$,
 select $n_0$ such that
\[
F^{[n]}(B_{\delta(a')}) \subseteq B_{\epsilon}(a') \quad \text{for all } n \geq n_0.
\]

By
\equationref{eq:Nsofsof},
\[
F^{[N+n]} (B_{\lambda}(a))
\subseteq B_{\epsilon}(a')
,
\]
and thus
\[
F^{[n']} (B_{\lambda}(a))
\subseteq B_{\epsilon}(a')
,
\quad \text{for all } n' \geq N+n_0
.
\]

This proves
uniform convergence
of the sequence $(F_{P_r}^{[n ]})$
to $a'$
at $B_{\lambda}(a)$.

\end{proof}

\end{lemma}

Conditions
\ref{l_A2-new} and
\ref{l_A2.5-new}
of the lemma are reflected
in conditions
\ref{A2-new}
and
\ref{A2.5-new}
of the following
theorem.

\begin{theorem}

  \label{theorem_L_A}

If for $r=1,2$, the following conditions hold:

\begin{enumerate}[(a)]

\item
\label{A1}

The attractor sets $\mathcal{A}_r$  are finite and disjoint.

  \item
  \label{A5}
  An $Im \in X$ belongs to
  basin $\mathcal B (a)$
of  some
$a \in \mathcal A _1$.

\item
\label{A2-new}
Every  $a \in \mathcal A_r$ belongs to the basin $\mathcal B(a')$ of some  $a' \in \mathcal A_{\alpha (r)}$
together
with an open ball of a certain radius
${\delta}(a)$ around $a$:
$B_{\delta (a)}(a) \subset
\mathcal B(a')
$.

\item
\label{A2.5-new}

For every  $a \in \mathcal A_{r}$
convergence of
the sequence of functions $(F_{P_r}^{[n]})$
to $a$ is locally uniform at $a$:
there exists $\delta (a) > 0$ such that for any $\epsilon > 0$, there exists $n_0$ such that
\begin{equation*}
F_{P_r}^{[n]}(B_{\delta(a)}(a))
 \subseteq
 B_{\epsilon}(a)
\end{equation*}
for any $n \geq  n_0$.
See \figureref{fig:uniform}.

  \end{enumerate}

Then
  the sequence of
\equationref{eq:U0}:
\begin{equation*}
\begin{split}
U(Im,nsteps_1,nsteps_2)=
Im
\xrightarrow{  [F_{P_1}]^{nsteps_1} }
T_{1,P_1}
\xrightarrow{  [F_{P_2}]^{nsteps_2} }
\\
T_{2,P_2} \xrightarrow{  [F_{P_1}]^{nsteps_1} }
T_{3,P_1}
\xrightarrow{  [F_{P_2}]^{nsteps_2} }
T_{4,P_2}
\xrightarrow{  [F_{P_1}]^{nsteps_1} }
\dots
\end{split}
\end{equation*}
  is a bipartite
  convergent
 sequence of the
  second type,
  converging to the orbit
  consisting of alternating
  attractors
  of
$F_{P_1}$ and $F_{P_2}$.

The orbit elements form a loop
with respect to alternating
$\widehat{F_1}$, $\widehat{F_2}$ operations:
\begin{equation}
\label{eq:att2}
{b_0}
\xrightarrow{  {\widehat{F_2}} }
{b_1}
\xrightarrow{  {\widehat{F_1}} }
\dots
\xrightarrow{  {\widehat{F_2}} }
\,
{b_{K-1}}
\,
{\xrightarrow{  \widehat{F_1} }}
\,
{b_0}
.
\end{equation}

The orbit elements
are also
alternating fixed points of functions
$\widehat{G_1}$ and $\widehat{G_2}$:
\begin{equation}
\label{eq:G1G2hatsforFixed}
\begin{aligned}
&\widehat{G_1}(b_h) = b_h, \text{ for even } h,
\\
&\widehat{G_2}(b_h) = b_h, \text{ for odd } h,
\end{aligned}
\end{equation}
where
\begin{equation*}
\begin{aligned}
&\widehat{G_1} = [\,\widehat{F_1} \cdot \widehat{F_2}\,]^{K/2},
\\
&\widehat{G_2} = [\,\widehat{F_2} \cdot \widehat{F_1}\,]^{K/2}
.
\end{aligned}
\end{equation*}

\begin{figure}[]
\centering
    {
    {\includegraphics[width=1.0\linewidth]{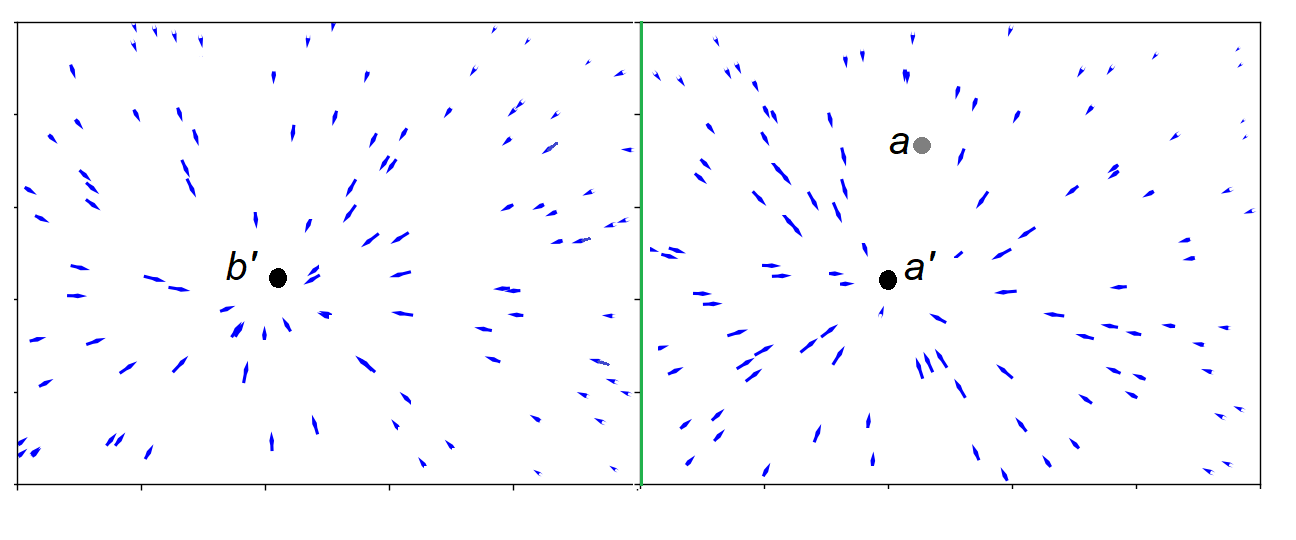}}}
    \caption{
        Illustration to
        the condition of
        local uniform convergence
        considered in
\lemmaref{lemma_theorem_L_A}
and
        \theoremref{theorem_L_A}.
        In the depicted hypothetical scenario,
        an image space $X$ is divided into two basins of attractors, $a'$ and $b'$, respectively,
for
        a mapping
        $F_{P}:X \rightarrow X$.
The basins are open sets, not
        including the border
        between the basins
        (shown as green line).
        The vector field denotes
        the direction of motion
        given by iteration
        $F_{P}$.\protect\footnotemark \,The magnitude of the vector field becomes infinitesimally small near the borders of the basins
        and near the attractors.
In
\lemmaref{lemma_theorem_L_A},
we claim local uniform convergence of the sequence of functions
 $([F_{P}]^n)$
to the attractor $a'$
at the point $a$,
given the
local uniform convergence
of the sequence
to $a'$,
at the point $a'$.
The latter condition is
stipulated in
\ref{A2.5-new} of
\theoremref{theorem_L_A}.
Note that the condition of global uniform convergence does not hold within the basins due to the behavior of the vector field near their borders
       }
    \label{fig:uniform}
\end{figure}

\begin{proof}

The central point in our proof will be  representation of
the sequence of
\equationref{eq:U0}
in the form
\begin{equation*}
\begin{array}{@{}cccccccc@{}}
\multicolumn{4}{l}{U(Im,nsteps_1,nsteps_2)= Im \xrightarrow{F_1} } &&&&
\\[0.3em]
 T_{1+0\cdot K,P_1} &\xrightarrow{F_2} & T_{1+0\cdot K+1,P_2} &
            \xrightarrow{F_1}  & \cdots            &\xrightarrow{F_2}       & T_{1+0\cdot K +K-1,P_2} &\xrightarrow{F_1}\\

 T_{1+1\cdot K,P_1} &\xrightarrow{F_2} & T_{1+1\cdot K +1,P_2} &
            \xrightarrow{F_1}  & \cdots            &\xrightarrow{F_2}       & T_{1+1 \cdot K + K-1,P_2} &\xrightarrow{F_1}\\
\dots &&&&&&& \\
 T_{1+m\cdot K,P_1} &\xrightarrow{F_2} & T_{1+m\cdot K+1,P_2} &
            \xrightarrow{F_1}  & \cdots            &\xrightarrow{F_2}       & T_{1+ m \cdot K + K -1,P_2} &\xrightarrow{F_1}\\
\dots \;\;. &&&&&&&   \\
\end{array}
\end{equation*}
such that
every $j$-th column
\begin{equation*}
  C_{j} (nsteps_1,nsteps_2) =
  \begin{pmatrix}
        T_{1+0\cdot K+j,P_{od(j+1)}}
        \\[0.45em]
        T_{1+1\cdot K +j,P_{od(j+1)}}
        \\
        \dots \\
        T_{1+m\cdot K+j,P_{od(j+1)}} \\
        \dots \\
  \end{pmatrix}
  ,
\end{equation*}
in this representation,
considered as a sequence,
converges to
an element
$b_j$
as $m$,
$nsteps_1$, and $nsteps_2$  tend to infinity.

To derive this representation,  we will show first that
the sequence of alternating
$\widehat{F_1}$, $\widehat{F_2}$ operations
applied to $Im$
terminates with a cycle
consisting of attractors
(step\,\ref{N3} below).
Then, we will
enclose
each attractor $a \in \mathcal A$ by a small
$\epsilon$-ball such that
${\widehat F}_r$
transformations
 map these ball to
inside analogous
balls around
${\widehat F}_{\alpha (r)}(a)$
(\equationref{eq:balls} below).
This will provide the
convergence of
the column sequences.
We proceed through the following steps:

\begin{enumerate}[(A)]

\item
\label{N0}

Applying
\lemmaref{lemma_theorem_L_A}
to
conditions
\ref{A2-new}
and
\ref{A2.5-new}
we obtain the
following
intermediate result:
every attractor $a \in \mathcal A_r$ belongs to  basin $\mathcal B(a')$ of some attractor $a' \in \mathcal A_{\alpha (r)}$.
Convergence of
the sequence of functions
$([F_{P_{\alpha (r)}}]^n)$
to $a'$ is locally uniform at $a$:
there exists $\lambda (a) > 0$ such that for any $\epsilon > 0$, there exists $n_0$ such that
\begin{equation}
\label{eq:lambda_atfar_B}
F^{[n]}(B_{\lambda(a)}(a)) \subseteq B_{\epsilon}(a')
\end{equation}
for any  $n \geq  n_0$.

\item
\label{N1}
Take an arbitrary $\epsilon > 0$.
Without loss of generality,
assume
that
\begin{equation}
\label{eq:wlog}
\epsilon < \min _
{a \in \mathcal A} \{ \lambda(a) \}
,
\end{equation}
where
\[
{\mathcal A} =
{\mathcal A _1}
\cup
{\mathcal A _2}
.
\]

\item
\label{N2 }
Consider the sequence $E$ of the elements
generated by
subsequent application
of alternating ${\widehat{F_1}}$ and
${\widehat{F_2}}$
operations
(\equationref{eq:F12_widehat}):
\begin{equation}
\label{eq:E}
E: Im
\xrightarrow{  {\widehat{F_1}} }
{a_1}
\xrightarrow{  {\widehat{F_2}} }
{a_2}
\xrightarrow{  {\widehat{F_1}} }
\dots
.
\end{equation}

By
\ref{A5} and \ref{A2-new},
$E$ consists of
attractors from $\mathcal A$.
Since $\mathcal A$ is finite,
starting with some
$a_{N_l}$ the
sequence becomes cyclic.  Since   $\mathcal A_1$ and
$\mathcal A_2$ do not intersect,
and the
cycle is generated by alternating
$\widehat{F_1}$ and $\widehat{F_2}$
operations,
it has
an even length which we denote
by $K$.

Select $m, N$ such that
$N\geq N_l$ and
\begin{equation}
\label{eq:N}
N = 1 + m\cdot K
.
\end{equation}
Then,
for $n\geq N$
sequence
$E$
is cyclic:
\begin{equation}
\label{eq:cyclic}
E: Im
\xrightarrow{  {\widehat{F_1}} }
{a_1}
\xrightarrow{  {\widehat{F_2}} }
\cdots
\xrightarrow{  {\widehat{F_1}} }
{a_N}
\xrightarrow{  {\widehat{F_2}} }
{a_{N+1}}
\xrightarrow{  {\widehat{F_1}} }
\cdots
\xrightarrow{  {\widehat{F_2}} }
\,
{a_{N+K-1}}
\,
{\xrightarrow{  \widehat{F_1} }}
\,
{a_N}
\xrightarrow{  {\widehat{F_2}} }
\cdots
.
\end{equation}

\item
\label{N3}
If we denote
\[
{b_0}={a_N}, \;
{b_1}={a_{N+1}}, \;
\ldots
,
{b_{K-1}}={a_{N+K-1}}
\]
then
for ${b_i}$,
Equations~\ref{eq:att2} and \ref{eq:G1G2hatsforFixed}
hold.

\item
\label{N4}

Basing on
\ref{A5},
select $n_0$ such that
\[
F_{P_1}^{[n]}  (Im)
\in
B_{\epsilon}(a_1)
\]
for any
$n \geq  n_0$.

Basing on
Equations~\ref{eq:wlog} and \ref{eq:lambda_atfar_B},
select  $n_1$ such that
for
$n\geq n_1$
\[
F_{P_2}^{[n]}  (B_{\epsilon}(a_1))
\subseteq
B_{\epsilon}(a_2)
  .
\]

Similarly,
select  $n_2$ such that
for
$n\geq n_2$
\[
F_{P_1}^{[n]}  (B_{\epsilon}(a_2))
\subseteq
B_{\epsilon}(a_3)
  ,
\]
etc.

Finally,
select  $n_{N+K-1}$ such that
for
$n\geq n_{N+K-1}$
\[
F_{P_1}^{[n]}  (B_{\epsilon}(a_{N+K-1}))
\subseteq
B_{\epsilon}(a_{N})
  .
\]

Then, for any
\begin{equation}
\label{eq:n}
n \geq  \max \{ n_i \;|\; i = 0, \ldots, N+K-1 \}
\end{equation}
hold

\begin{equation}
\label{eq:balls}
\begin{aligned}
&F_{P_1}^{[n]}(Im) \in B_{\epsilon}(a_1),
\\
&F_{P_2}^{[n]}(B_{\epsilon}(a_1)) \subseteq B_{\epsilon}(a_2),
\\
&\dots ,
\\
&F_{P_2}^{[n]}(B_{\epsilon}(a_N)) \subseteq B_{\epsilon}(a_{N+1}),
\\
& \dots,
\\
&
F_{P_1}^{[n]}(B_{\epsilon}(a_{N+K-1})) \subseteq B_{\epsilon}(a_{N}),
\end{aligned}
\end{equation}

\item
\label{N5}

Select arbitrary 
$n$ satisfying 
\equationref{eq:n},
along with
$nsteps_1, nsteps_2  \geq n$,
and represent the sequence
of
\equationref{eq:U0}
as a matrix with $K$ columns and an infinite number of rows:
\begin{equation}
\label{eq:matrix_0}
\begin{array}{@{}cccccccc@{}}
\multicolumn{4}{l}{U(Im,nsteps_1,nsteps_2)= Im \xrightarrow{F_1} } &&&&
\\[0.3em]
 T_{1+0\cdot K,P_1} &\xrightarrow{F_2} & T_{1+0\cdot K+1,P_2} &
            \xrightarrow{F_1}  & \cdots            &\xrightarrow{F_2}       & T_{1+0\cdot K +K-1,P_2} &\xrightarrow{F_1}\\

 T_{1+1\cdot K,P_1} &\xrightarrow{F_2} & T_{1+1\cdot K +1,P_2} &
            \xrightarrow{F_1}  & \cdots            &\xrightarrow{F_2}       & T_{1+1 \cdot K + K-1,P_2} &\xrightarrow{F_1}\\

\dots &&&&&&& \\

 T_{1+m\cdot K,P_1} &\xrightarrow{F_2} & T_{1+m\cdot K+1,P_2} &
            \xrightarrow{F_1}  & \cdots            &\xrightarrow{F_2}       & T_{1+ m \cdot K + K -1,P_2} &\xrightarrow{F_1}\\

\dots \;\;. &&&&&&&   \\
\end{array}
\end{equation}
where 
$F_1$ and $F_2$ are defined as in
\equationref{eq:F1F2}:
\begin{equation*}
\begin{aligned}
&F_1 = [F_{P_1}]^{nsteps_1},\\
&F_2 = [F_{P_2}]^{nsteps_2}
\,.
\end{aligned}
\end{equation*}

From \equationref{eq:balls},
it follows
that
\begin{equation*}
T_{1+0\cdot K,P_1} \in B_{\epsilon}(a_1),
\;
T_{1+0\cdot K + 1,P_2} \in B_{\epsilon}(a_2),
\cdots \;,
T_{N=1+m\cdot K ,P_1} \in B_{\epsilon}(a_N)
.
\end{equation*}

Also,
since the elements
$a_q$, $q\geq N$, form a loop
(see \equationref{eq:cyclic}),
from \equationref{eq:balls},
it follows that
starting from the $m$-th row of
the matrix of $T$-elements
of \equationref{eq:matrix_0}, the following
holds:
\begin{equation*}
\begin{aligned}
&\hspace{-4em}T_{1+m\cdot K,P_1} \in B_{\epsilon}(a_N), \;\;
T_{1+m\cdot K + 1,P_2} \in B_{\epsilon}(a_{N+1}),
\, \dots \;,
\\
&\hspace{10em}
T_{1+m\cdot K+K-1 ,P_2} \in B_{\epsilon}(a_{N+K-1}),
\\
&\hspace{-4em}T_{1+(m+1)\cdot K,P_1} \in B_{\epsilon}(a_N), \;\;
T_{1+(m+1)\cdot K + 1,P_2} \in B_{\epsilon}(a_{N+1}),
\, \dots \;,
\\
&\hspace{10em}
T_{1+(m+1)\cdot K+K-1 ,P_2} \in B_{\epsilon}(a_{N+K-1})
\\
&\hspace{-4em}T_{1+(m+2)\cdot K,P_1} \in B_{\epsilon}(a_N), \;\;
T_{1+(m+2)\cdot K + 1,P_2} \in B_{\epsilon}(a_{N+1}),
\, \dots \;,
\\
&\hspace{10em}
T_{1+(m+2)\cdot K+K-1 ,P_2} \in B_{\epsilon}(a_{N+K-1})
\\
&\hspace{-4em}\dots
\;\;\;
.
\end{aligned}
\end{equation*}

Therefore, for
 the $j$-th
column:
\begin{equation*}
  C_{j} (nsteps_1,nsteps_2) =
  \begin{pmatrix}
        T_{1+0\cdot K+(j-1),P_{od{(j)}}}
        \\[0.45em]
        T_{1+1\cdot K +(j-1),P_{od{(j)}}}
        \\[0.45em]
        \dots
        \\[0.45em]
        T_{1+m\cdot K+(j-1),P_{od{(j)}}} \\
        \dots \\
  \end{pmatrix}
\end{equation*}
of the matrix,
the elements starting
with
$T_{1+m\cdot K+(j-1),P_{od{(j)}}}$
belong to
$B_{\epsilon}(a_{N+j-1})$,
$j=1,\dots, K$.
This proves
convergence of every $j$-th
column's sequence to
$a_{N+j-1}$ and completes the theorem.
\end{enumerate}
\end{proof}
\end{theorem}

\footnotetext{similarly to \citet{Belkin}.}

\section{Examining the Fixed Point Property Through Analysis of a Statement}
\label{SI:semiotics}

In this section, we will aim to reveal in the statement
"I see this image" the fixed point property,  incorporated  in
our functional modeling of perceptual inference in
\citet{isn}. For this purpose, we will compare different
interpretations of this statement.

One meaning of the statement is "{\em the result} of my seeing is this image". This implies equivalence between two images. The first one is what the person refers to as "I see". The
seeing may be represented as  the application of the $seen$ operator to some image $\widehat{x}$. The second image is "this image". In such a way, the meaning may be expressed as
\[
seen(\widehat{x}) = \text{"this image"}.
\]

What is $\widehat{x}$ to whom the "see" is applied?
Note that another meaning of "I see this image" is "see {\em this image}",
that is, "see" is applied to "this image."
This specifies $\widehat{x}$: \[
\widehat{x} \text{ is "this image"}.
\]
This results in the fixed point property: \[
seen(\widehat{x}) = \widehat{x}.
\]

\section{Remarks on the Image Representations in the CONN  Classifiers}
\label{SI:vanilla_vs_stochastic}

\begin{figure}[]
    \centering
      {
        \subfigure[]{\label{fig:image-a9}
          {\fbox{\includegraphics[width=0.45\linewidth,height=0.45\linewidth]{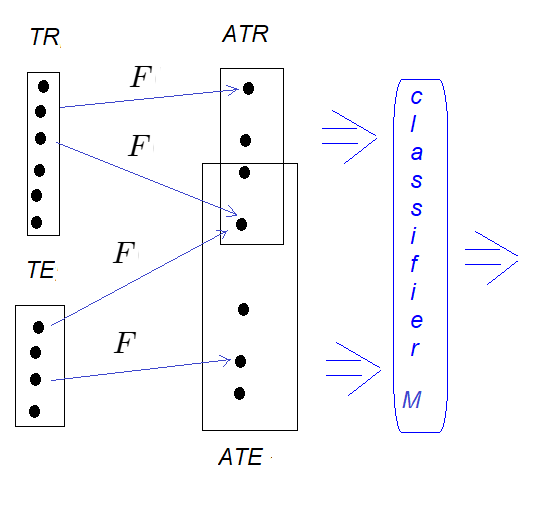}}}}
                \qquad
        \subfigure[]{\label{fig:image-b9}
          {\fbox{\includegraphics[width=0.35\linewidth,height=0.45\linewidth]{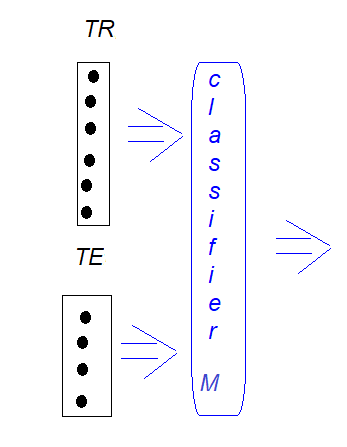}}}}
} 
\caption{
(a): Representation of the operations of CONN
classifiers: 1) Transformations $F$ applied to
the elements of the training set ($TR$) and test
set ($TE$), resulting in new sets $ATR$ and $ATE$,
which represent the images "perceived" by the
classifiers. For the vanilla classifier, $F$
corresponds to the transformation $\widehat{F}$
to attractors (\equationref{eq:nakaonetc}); for
stochastic classifiers, $F$ denotes the
transformations $F^*$ to the averaged
ensembles of attractors. Note that the notation
$TE \xrightarrow{F} ATE$ is used for the
classifier analysis, while the calculation of
the classifier value during inference proceeds
independently for each input image sample.
2) Applications of the baseline classifiers.
(b): Comparisons of the baseline classifiers $M$
with the attractor-based classifiers.
}
      \label{fig:AEBC}
\end{figure}

Our vanilla CONN classifier is illustrated in
\figureref{fig:AEBC}.
It is based on the representation of an input sample $Im$ by  the attractor:
\begin{equation}
\label{eq:nakaonetc}
\widehat{F}(Im) =
\lim \limits_{n\to\infty} [dec(enc)]^n(Im)
,
\end{equation}
where $enc$ and $dec$
are encoding and decoding
operations of an autoencoder
trained at the training dataset
$TR$.

The training set fidelity of the vanilla CONN classifier is inherited from that of the baseline classifier:

  \begin{lemma}
\label{vanilla_properly_at_train}
Suppose the training examples comprising $TR$ are memorized by the autoencoder, and a baseline classifier $M$ trained on $TR$ correctly classifies them. In that case, the vanilla CONN classifier will also correctly classify the training examples from $TR$.

\begin{proof}
Due to the memorization, the function $\widehat{F}$ does not alter the training examples,
that is, $ATR=TR$. Therefore, the classifier $M$
trained at $ATR$ properly classifies the training examples.
\end{proof}
\end{lemma}

\begin{figure}[h]
    \centering
{\fbox
    {\includegraphics[scale=.6]{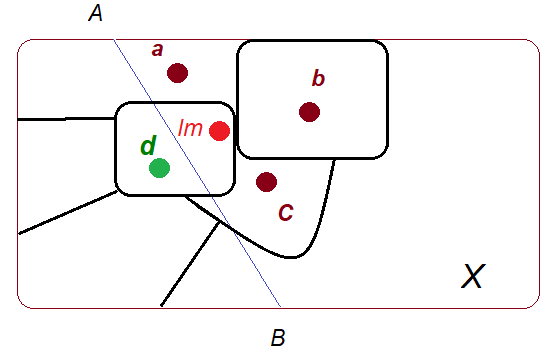}}}
    \caption{
Illustration of image representations
in vanilla and stochastic attractor-based
classifiers, using a two-class classification
problem. Line $(A, B)$ (in blue) marks the ground
truth division of the image space $X$ into
two classes. The partitioning of $X$ by
basins of the attractors (memorized training
examples) is shown. Image $Im$ denotes
a test sample. The examples $a$, $b$, and
$c$ (of the first class) and $d$ (of the
second) are the training examples located
close to $Im$. In the vanilla classifier,
$Im$ is represented by a single attractor $d$. The attractor assembly
used for representing $Im$ in the stochastic
CONN classifier comprises multiple instances
of $a$, $b$, $c$, and $d$, providing a more
informative representation of $Im$ by
attractors from different classes
}
    \label{fig:stoch_motivation}
\end{figure}

The representation $\widehat{F}$,  employed in
the vanilla classifier, may not be sufficiently informative  for individual samples, leading to misclassifications.
This is illustrated in  \figureref{fig:stoch_motivation},  where we depict  a two-class classification problem in the image space $X$.
The vanilla
classifier maps
a test sample $Im$ to
attractor $\widehat{F}(x)=d$,
followed by application of $M$ to $d$. Assuming that the baseline classifier $M$ correctly classifies the training example $d$,
 one may observe that
the vanilla classifier  misclassifies  $Im$.

This example also demonstrates that
the vanilla classifier behaves similarly to the one-nearest neighbor approach. It maps \(Im\) to the "nearest" training example \(d\). This representation may lose information about \(Im\), leading to misclassification.

The stochastic classifier is designed to leverage a more informative image representation.
It generalizes the vanilla classifier by representing $Im$ through a combination of several "nearest" training examples instead of a single one.
As a result, the  classifier  makes more informed decisions when classifying input images.

\section{Visualization of
Attractors of Undercomplete Autoencoders}
\label{SI:Attractors_P2P}

In the experiments described below, we trained two autoencoders,
denoted as $A_{P_1}$ and $A_{P_2}$,
on all odd and even digits from the MNIST training dataset, respectively.

The autoencoders
share the same
architecture:  a neural network with four fully connected layers in both the encoder and decoder, each containing 512 hidden units and ReLU activation functions. The input size is 784, and the bottleneck layer has a size of 2.

\begin{figure}[H]
\vspace{-32pt}
    \centering
    {
      \subfigure[]{\label{fig:image-a}
        \includegraphics[width=0.8\linewidth]{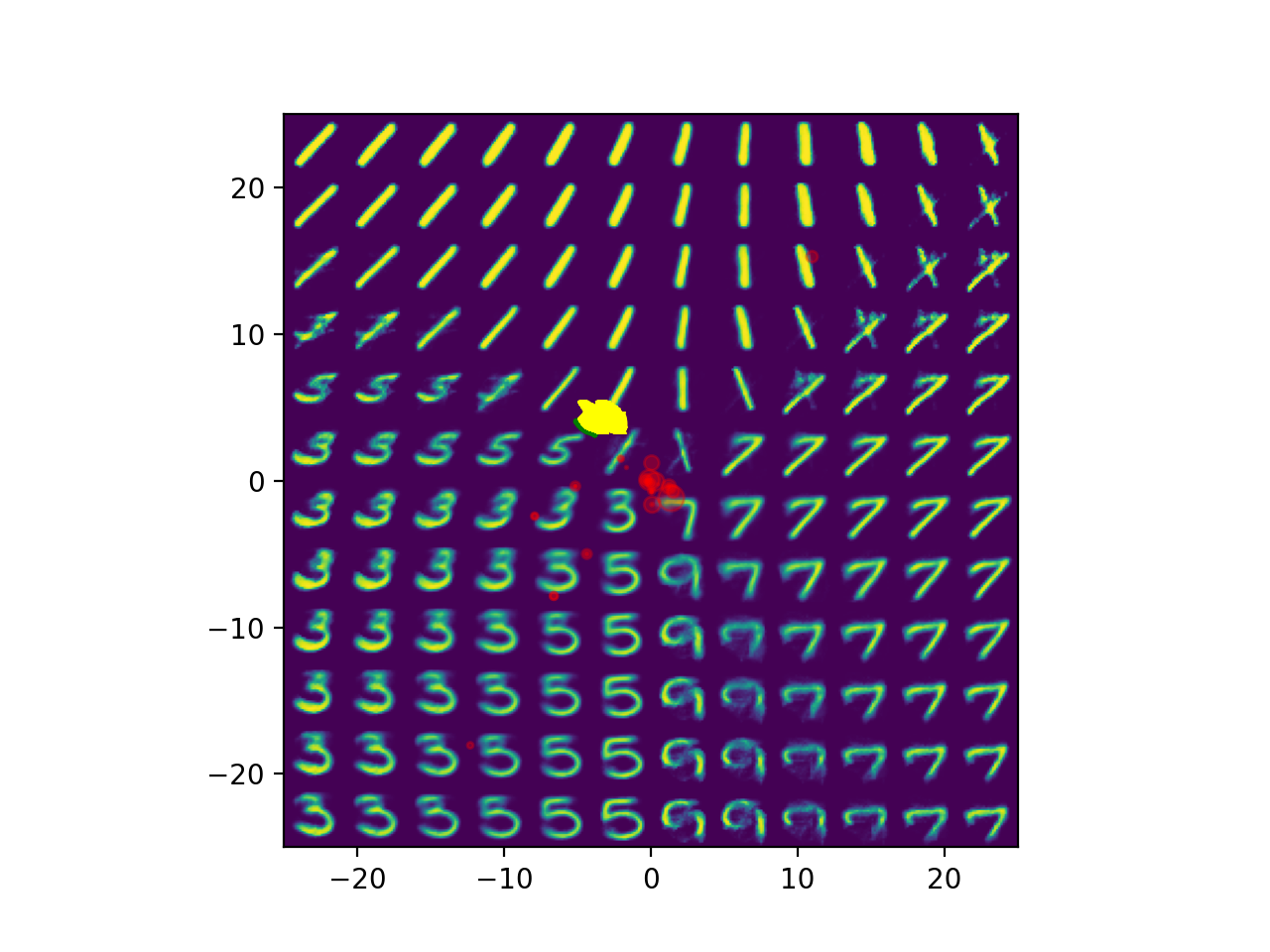}}
      \subfigure[]{\label{fig:image-b}
        \includegraphics[width=0.8\linewidth]{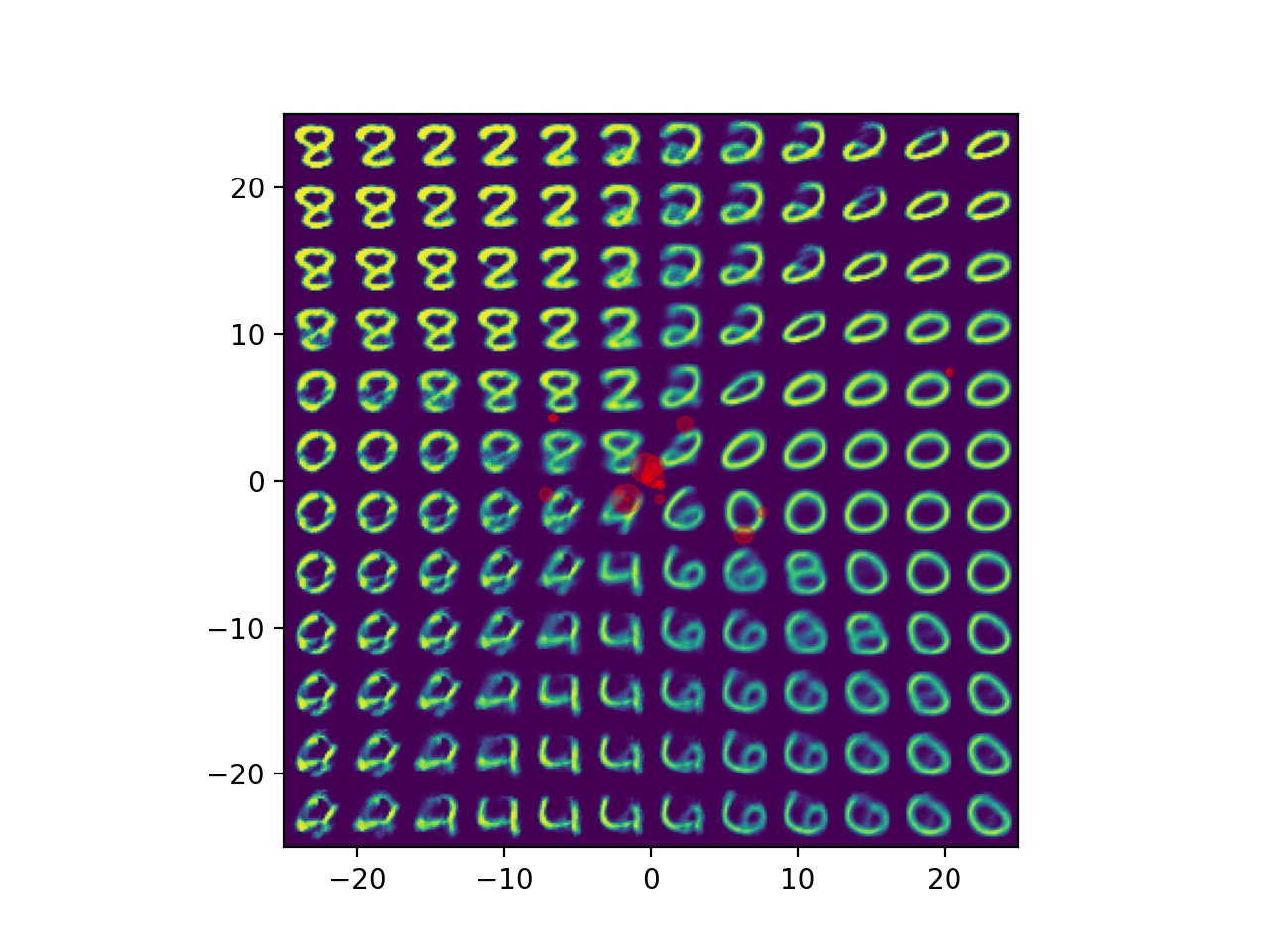}}
    }
    \caption{
Visualization of the latent spaces of  undercomplete autoencoders trained on the odd and even MNIST digits is shown in (a) and (b), respectively. Red circles indicate attractor locations, with radii proportional to the number of random samples falling into the attractor basins. In (a), some samples lead to sequences converging to cycles, as highlighted in yellow.  Refer to \figureref{fig:static_points_P2cycles}.
      }
    \label{fig:bins_as_circles}
  \end{figure}

We visualized the latent space of the trained autoencoders similarly to \citet{vae}. For each autoencoder, we partitioned the rectangle encompassing the range of encoder values in the 2-dimensional latent space into small squares. Each square was visualized by plotting the image obtained from applying the decoder operation to the center of the square. See \figureref{fig:bins_as_circles}.

To identify the attractors, we randomly sampled 2,000 points in the latent space of each autoencoder. We then iteratively applied decoder-encoder steps to each point until the consecutive vectors became sufficiently close, indicating convergence to a fixed point. Each converged vector was assigned to a predefined small bin in the latent space to which the vector belongs. Multiple vectors falling into the same bin were considered to represent the same attractor.

Due to computational constraints, we did not verify whether the identified fixed points are true attractors. This verification would involve calculating the largest eigenvalue of the Jacobian matrix for each fixed point and checking if it exceeds 1 (see \cite{B2_paper}).

The number of elements in each bin indicates the cardinality of the respective attractor's basin. The distribution of these counts for the autoencoders is visualized in \figureref{fig:bins_as_circles}, where the bins are represented by circles with radii proportional to the bin counts.

\begin{figure}[H]
    \centering
    {
      \subfigure[]{\label{fig:static_points_P2cycles:a}
        \includegraphics[width=0.4\linewidth]{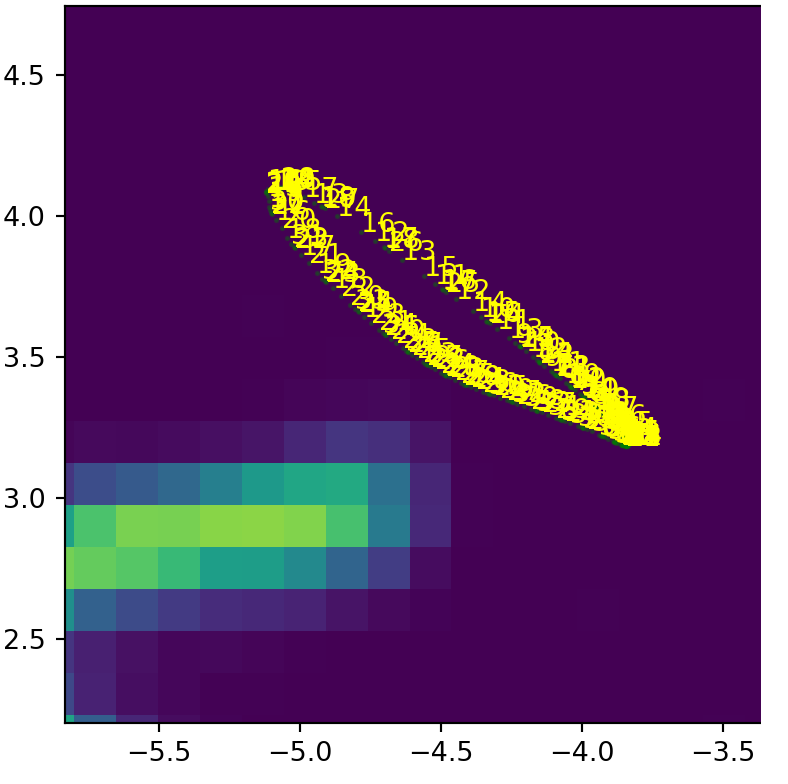}}
      \qquad
      \subfigure[]{\label{fig:static_points_P2cycles:b}
        {\includegraphics[width=0.4\linewidth]{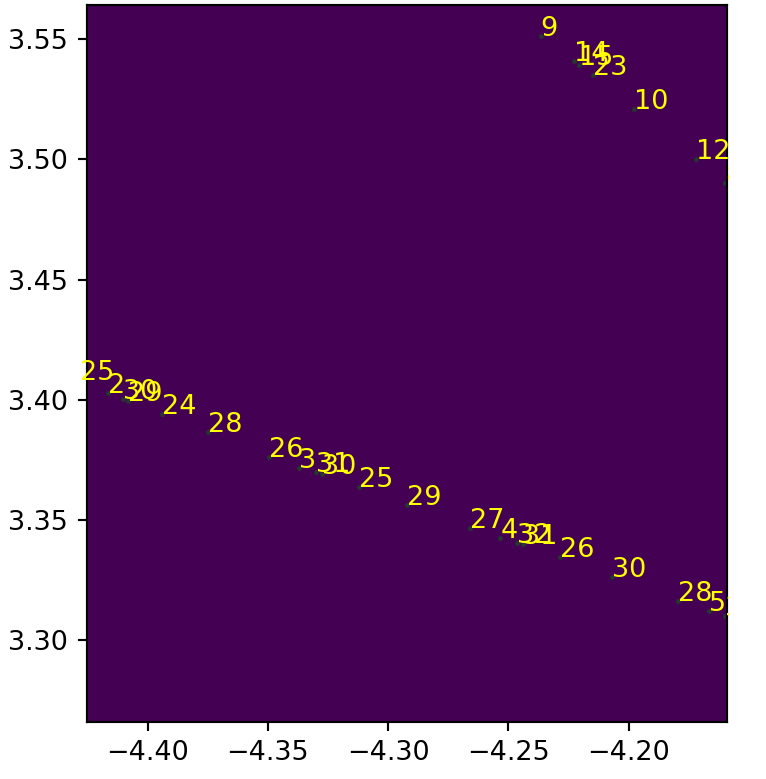}}}
    }
    \caption{
A cycle produced by
the undercomplete autoencoder trained on the odd MNIST digits.
  (a): Zoomed-in view of the yellow "ring" from \figureref{fig:image-a},
  (b): A closer detail of a section of the 'ring' from (a). Refer to the text for additional details
   }
    \label{fig:static_points_P2cycles}
\end{figure}

For 247 sampled points, the sequences for autoencoder $A_{P_1}$ did not converge to the fixed points. Instead, they exhibited convergence to cyclic sequences. In \figureref{fig:static_points_P2cycles}, we visualize these cycles as follows: for every element in the cycle, we display its subsequent number within the cycle at the element's location in the latent space.

\section{Details of
Training Overparameterized  Autoencoders at Restricted MNIST Datasets}
\label{SI:training details}

Our fully convolutional overparameterized autoencoder
(\citet{isn}, Section 7.3), has an architecture similar to that of \citet{Belkin}. It has five encoder  and eight decoder layers, each with a kernel size of 3. Cosine activation function is used. Each internal layer operates on 256 channels.

During training, we employed the Adam optimizer with a learning rate of 1e-5 and a seed of 3, using a minibatch size of 4. The networks were trained on restricted MNIST datasets (\citet{mnielsen, mnielsen2}), which are referenced in Table \ref{tab:TrainTable}. For instance, RMnist8 represents a training database with 8 examples per digit randomly selected from the training MNIST dataset.

The training was done for 110,000 epochs. Starting from the 100,000th epoch, the model that minimized the training error over the interval from the 100,000th epoch to the current epoch was selected. The goal of this strategy was to approach the point of overfitting on the training set. Due to resource limitations, we trained the autoencoders for fewer epochs compared to \citet{BelkinSI}. As a result, our training errors are still higher than those reported there. \tableref{tab:TrainTable} shows the minimal training errors attained for each restricted MNIST dataset.

\begin{table}[hbtp]
\centering
  \caption{Training errors achieved by the autoencoders trained on restricted MNIST datasets}
  {
    \begin{tabular}{|l|l|l|l|l|}
    \hline
    \bfseries RMnist5 & \bfseries RMnist6 & \bfseries RMnist7 & \bfseries RMnist8 & \bfseries RMnist9\\\hline
    0.03 & 0.04 & 0.04 & 0.04 & 0.04\\\hline
    \bfseries RMnist10 & \bfseries RMnist20 & \bfseries RMnist30 & \bfseries RMnist40 & \bfseries RMnist50\\\hline
    0.03 & 0.04 & 0.04 & 0.04 & 0.04\\\hline
    \end{tabular}
  }
  \label{tab:TrainTable}
\end{table}

\pagestyle{empty}
\clearpage
\pagestyle{plain}

\begin{figure}[H]
  \centering
  \begin{tabular}{ccc}
    \includegraphics[width=0.17\linewidth]{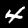} &
    \includegraphics[width=0.17\linewidth]{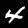} &
    \includegraphics[width=0.17\linewidth]{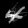} \\
    '4' & '4' & '4' \\
    [6.pt] 
    \includegraphics[width=0.17\linewidth]{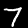} &
    \includegraphics[width=0.17\linewidth]{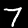} &
    \includegraphics[width=0.17\linewidth]{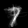} \\
    '7' & '7' & '7' \\
    [6.pt] 
    \includegraphics[width=0.17\linewidth]{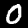} &
    \includegraphics[width=0.17\linewidth]{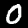} &
    \includegraphics[width=0.17\linewidth]{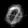} \\
    '0' & '0' & '0' \\
  \end{tabular}
  \caption{
    Simulation of perceptual inference in the vanilla and stochastic classifiers on the training examples used in the experiments.
    Shown are examples from the training set and the corresponding images "perceived" by the vanilla and stochastic CONN classifiers.
    Left column: Original "observed" images from the training set RMnist8, each annotated with its ground truth label.
    Middle column: The respective images "perceived" by the vanilla classifier, annotated with the labels assigned by the classifier.
The images in the middle column are indistinguishable from the original images due to the memorization
of the training examples.
    Right column: The respective images "perceived" by the stochastic classifier, annotated with the labels assigned by the classifier.
      }
  \label{fig:obs_seen_both_train}
\end{figure}

\begin{figure}[H]
  \centering
  \begin{tabular}{ccc}
    \includegraphics[width=0.17\linewidth]{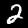} &
    \includegraphics[width=0.17\linewidth]{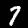} &
    \includegraphics[width=0.17\linewidth]{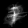} \\
    '2' & '7' & '2' \\
    [6.pt] 
    \includegraphics[width=0.17\linewidth]{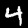} &
    \includegraphics[width=0.17\linewidth]{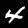} &
    \includegraphics[width=0.17\linewidth]{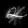} \\
    '4' & '4' & '4' \\
    [6. pt] 
    \includegraphics[width=0.17\linewidth]{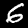} &
    \includegraphics[width=0.17\linewidth]{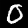} &
    \includegraphics[width=0.17\linewidth]{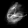} \\
    '6' & '0' & '6' \\
  \end{tabular}
  \caption{
    Simulation of perceptual inference in the vanilla and stochastic classifiers on the test examples used in the experiments.
    Shown are examples from the test set and the corresponding images "perceived" by the vanilla and stochastic CONN classifiers.
    Left column: Original "observed" images from the MNIST test set, each annotated with its ground truth label.
    Middle column: The respective images "perceived" by the vanilla classifier, annotated with the labels assigned by the classifier.
    Right column: The respective images "perceived" by the stochastic classifier, annotated with the labels assigned by the classifier.
  }
  \label{fig:obs_seen_both_test}
\end{figure}

\begin{figure}[]
\centering
  {
    \subfigure[]{\label{fig:image-a3}
      \includegraphics[width=0.3\linewidth]{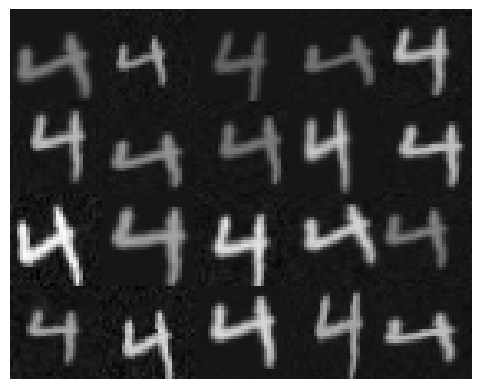}}
    \qquad
    \subfigure[]{\label{fig:image-b3}
      \includegraphics[width=0.3\linewidth]{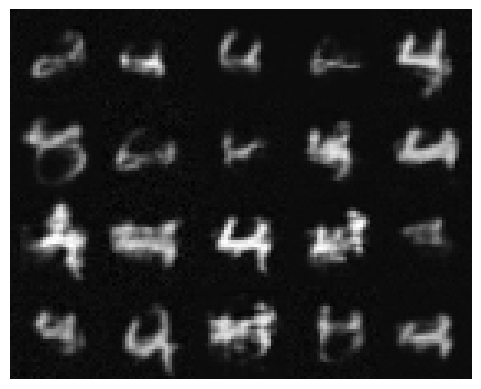}}
    \\\vspace{2ex}
    \subfigure[]{\label{fig:image-c3}
      \includegraphics[width=0.3\linewidth]{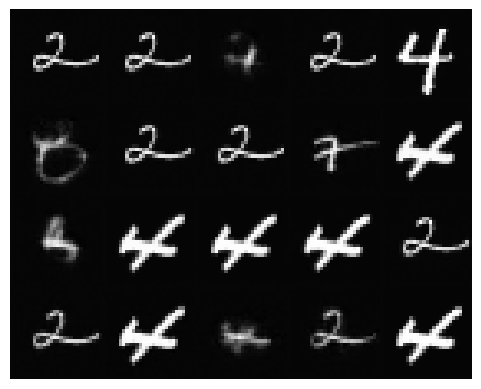}}
    \qquad
    \subfigure[]{\label{fig:image-d3}
      \includegraphics[width=0.3\linewidth]{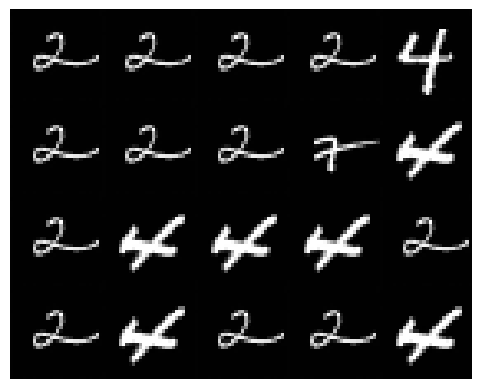}}
  }
  \caption{
Construction of the image "perceived" by the stochastic classifier, shown in the middle row's right column of
\figureref{fig:obs_seen_both_test}.
    For each iteration, 20 images from the generated ensemble of attractors are displayed.
    (a): Random augmentations of the input image before the encoder/decoder operation at the first iteration.
    (b), (c), and (d) depict the ensemble
    images after the 2nd, 5th, and 14th iterations, respectively
}
  \label{fig:SI:hconn_results_series1}
\end{figure}

\section{
Construction of the "Perceived" Images for the Vanilla
and Stochastic CONN Classifiers}
\label{SI:seen for stochastic}

Below, we illustrate the construction of the "perceived"
images for both the vanilla and stochastic CONN classifiers
described in \citet{isn}, Section 7.3.
The classifiers were trained on
the dataset RMnist8
(see \sectionref{SI:training details}).

Figures~\ref{fig:obs_seen_both_train} and \ref{fig:obs_seen_both_test}
illustrate this construction for
the "observed"  images selected
from the classifier's training set
and MNIST test set, respectively.

Note that since the autoencoders for the CONN
classifiers are trained on the
classifiers' training sets
(RMnist8, in our case), and due to the memorization of training examples,
for the vanilla classifier,
for "observed" images selected from the training set,
the "perceived" images coincide with the
observed (see \citet{isn}, Section 5.1).
Comparison of the left and middle columns
in \figureref{fig:obs_seen_both_train}
confirms this phenomena.

In \figureref{fig:SI:hconn_results_series1},
the details of the
construction of the image "perceived" by the stochastic
classifier, shown in the middle row's right column of
\figureref{fig:obs_seen_both_test}, are provided.

\FloatBarrier
\newpage

\section{Resilience to Adversarial Attacks of Vanilla CONN Classifiers}
\label{resil_vanilla}

It is widely acknowledged that human perception exhibits greater resilience against adversarial attacks compared to neural networks (for example, \citet{Adversarial_examples_in_the_physical_world,limitations_deep_learning}). Are CONN classifiers, which mimic certain properties of human visual perception, also resilient to adversarial attacks?

Below, we explain our assumption that vanilla CONN classifiers (\citet{isn}),
trained on small sets of  examples with sufficiently large distances between the
examples,
are inherently resilient to adversarial attacks.
This assumption is based on the fact that the vanilla classifier assigns the same label to all testing images within the basin $\mathcal{B}(a)$ of attractor $a$ (\citet{isn}, Section 5.1).

Specifically, any image $Im$ from an attractor basin
may be surrounded by an
open neighborhood included in the basin.
Thus, for a vanilla classifier, this neighborhood may serve
as a "container," such that classical adversarial attacks
(see below) using examples from this container cannot
succeed. On the scale of the entire image space $X$,
the resilience of the classifier is determined by whether, for
a large proportion of images $Im$, the image can be
immersed in a "container" that is sufficiently large
to encompass all examples that are perceptually indistinguishable from $Im$.

Intuitively, one may expect that
for perceptually distinct training examples (attractors),
the proportion of such "defended" images
$Im$ will be large.
In this section, we justify this assertion.

Our approach is twofold:
first, through mathematical definitions,
including the introduction and analysis of
the Basins Separation Index, and second,
through informal notions and experimental
observations regarding the behavior
of overparameterized autoencoders.

We proceed as follows.
In \sectionref{resil_vanilla_which}
we specify the adversarial attacks under consideration.
In \sectionref{resil_vanilla_subset}, given a vanilla CONN classifier, constructed for a set $\mathcal{A}$ of training examples memorized as attractors, we consider a subset $R_T(X)$ where  resilience to adversarial attacks against the elements of $R_T(X)$ is guaranteed (\conjectureref{conject1}). Then, in \sectionref{resil_vanilla_index}, we define a measure (Class Separation Index)
reflecting the "proportion" of $R_T(X)$ in $X$,  and explain why the measure is elevated for overparameterized autoencoders. This will explain our assumption about the resilience of vanilla CONN classifiers.

\subsection{Assumptions on Adversarial Attacks and  Baseline Classifiers}
\label{resil_vanilla_which}

In this section, we specify the adversarial attacks considered against vanilla CONN classifiers. We also introduce reasonable assumptions regarding the training examples and the baseline classifiers.

A classical successful perturbation adversarial
attack against an image $Im$ involves adding
subtle noise to $Im$, yielding
a new image $Im'$. The image $Im'$ appears
indistinguishable from $Im$ to the human eye,
yet a neural network produces an output for $Im'$
that differs from the output for the original image $Im$.

These are the adversarial attacks considered
in this section in relation to the vanilla
CONN classifier. Note that we do not require
the network to return the correct class for $Im$
in this setting. Furthermore, high resilience
to the considered attacks does not necessarily
indicate high accuracy of the classifier.
For example, a classifier that returns the
same label for all MNIST examples is resilient
against any attack but has very low accuracy.

Various approaches have been proposed to achieve resilience
in deep learning models, addressing  various types  of resilience.
See, for example, \citet{Long2022}.
These include the use of autoencoders for denoising (\citet{adv_autoenc,adv_autoenc_2})
and ensembles (\citet{adv_autoenc, adv_Lin}). Another approach involves adding noise
to the training examples (\citet{adv_noise_train_only}), test examples (\citet{adv_Lin}),
or both (\citet{adv_noise_train_test}). However, these strategies do not
guarantee perfect resilience, as no method has yet demonstrated
complete robustness (\citet{failed_defences}).

The representation of input samples by attractors, as explored in vanilla CONN classifiers, has a distinct potential to hinder perturbation attacks, differing from the aforementioned approaches, as we demonstrate in this section.

In our analysis, we will consider attacks against such images
$Im$  for which the sequence in \equationref{eq:nakaonetc} converges to
\begin{equation}
\label{eq:nakaonetc_B}
\widehat{F}(Im) =
\lim \limits_{n\to\infty} [dec(enc)]^n(Im)
,
\end{equation}
which is a memorized training
example.
Consider to what extent this assumption is justified.

Note that convergence to an attractor in the above equation
for an arbitrary
image $Im$
is not guaranteed. For instance,  \figureref{fig:uniform} illustrates a scenario where any sequence, starting from any image on the border between attractor basins, consists of images that remain on the border and thus does not converge to an attractor.

However, for overparameterized autoencoders, non-convergence to an attractor is rare.
For example, in our experiments with overparameterized autoencoders (\sectionref{SI:training details}), we observed convergence to training examples memorized as
 attractors for
every
image $Im$. Empirical evidence in \citet{Belkin} demonstrates that, for a given image
 $Im$, the sequence of encoding and decoding operations typically converges to an
 attractor, which can be either a memorized example or a spurious attractor.
 Hence, we restrict our analysis
to attacks on images
where the sequence in
\equationref{eq:nakaonetc_B} converges
to an attractor,
leaving the case of non-convergence
for further research.

\begin{remark}
\label{rem:the scenario}
While convergence to spurious attractors
is rare, as demonstrated by \citet{Belkin},
we focus on the scenario
where the attacked images are
those $Im$ for which the sequence
\equationref{eq:nakaonetc_B} converges
to an attractor which is a
memorized training example.
\end{remark}

These images comprise the union of the
basins of training examples memorized as
attractors. We denote the set as
\begin{equation}
\label{eq:H_A}
H_A,
\end{equation}
where $A$ refers to the autoencoder used in the vanilla classifier.

Let us define now how
the vanilla CONN
classifier
handles
input samples $Im'$ for which the sequence
\begin{equation}
\label{eq:nakaonetc Im'}
\lim \limits_{n\to\infty} [dec(enc)]^n(Im')
\end{equation}
 does not converge to
a memorized training example. It is generally unknown whether
this represents a rare case of the original sample with
such a property, or a sample resulting from the adversarial
attack. 
To facilitate the analysis presented in this section, 
we consider an inference method for the vanilla 
classifier that is defined slightly differently from 
the inference method
in \citet{isn}:
\begin{remark}
\label{rem:infer}
Given an input $Im'$,
we first check whether \equationref{eq:nakaonetc} converges
to a memorized training example. If this does not hold,
we classify the sample as "suspicious of an adversarial
attack" and assign it a randomly selected class label.
\end{remark}

As follows from the above, this adjustment in inference practically does
not affect the performance of the classifier when it is
not subjected to adversarial attacks.

Consider now the assumptions regarding the classifiers addressed  in the following subsections.
As mentioned at the beginning of \sectionref{resil_vanilla},
we consider the vanilla CONN classifiers,
trained on small sets of  examples, with sufficiently large distances between them
(specifically, these conditions will be discussed in \sectionref{Lower}). We stipulate the following reasonable assumption regarding the baseline classifiers $M$ embedded in the workflow of the considered
vanilla CONN classifiers:
\begin{remark}
\label{rem:M}
We assume that
every trained baseline classifier
$M$ properly
classifies the
examples from its training set.
\end{remark}
(See also \lemmaref{vanilla_properly_at_train}.)

The following obvious remark will help us avoid overloading the context in the following subsections:
\begin{remark}
\label{rem:nexamples}
We assume that the number of training examples for the considered autoencoders is at least 2.
\end{remark}

\subsection{Subset with Guaranteed Resilience}
\label{resil_vanilla_subset}

In this section, for a given vanilla CONN classifier $C$ constructed for the set $\mathcal{A}$
of training examples memorized  as attractors, we define a subset $R_T(X)$ (\equationref{eq:S_1}) of the  image space $X$. Our goal is to show that resilience to the attacks against
elements of $R_T(X)$ is guaranteed.
This will follow from 
\lemmaref{lemma_adv}.

First, let us note the following:
\begin{remark}
\label{rem:the_same_label}
As shown in
\citet{isn}, Section 5.1,
for
a vanilla CONN classifier   $C$
constructed for an autoencoder $A$,
and
an attractor $a$ of $A$,
the classifier
 assigns the same label to all images in the basin $\mathcal{B}(a)$.
\end{remark}

Let $T$ denote a threshold of similarity between images, such that images with a distance $d$ between them greater than or equal to $T$ are distinguishable from each other, in the context of an adversarial attack described above.
Let $B_{T}(Im)$
denote the
open ball with radius $T$ around $Im$,
and let
$\mathcal{B}(a)$ denote
the basin of
an attractor $a$.
Let, for an image $Im$, $a(Im)$ denote the attractor of
the basin to which $Im$ belongs, if such a basin exists.

\begin{definition}[Subset with Guaranteed Resilience of an Autoencoder]
\label{def_R}
Let $\mathcal{A}$ be a
the set of training examples memorized  as attractors.
Let $A$ be an autoencoder constructed for  $\mathcal{A}$.
Let $c_{gt}: X \rightarrow Q$ assign ground truth labels from set $Q$.
Consider the subset of $X$:
\begin{equation}
\label{eq:S_1}
\begin{aligned}
&\hspace{-0em}
R_{T,\mathcal{A},A,c_{gt}}(X) = \{ Im \in H_A \;|\;
 \text{
$Im$  belongs to basin of
some $a \in  \mathcal{A}$},
\\
&\hspace{1em}
 \text{
 and
for any $a' \in  \mathcal{A}$ such that
$B_T(Im) \cap \mathcal{B}(a')$ is
non-empty,}
\\
&\hspace{1em}
 \text{
it holds that
$c_{gt}(a')=c_{gt}(a)$
 }
  \}
.
\end{aligned}
\end{equation}
\end{definition}
For simplicity,  denote $R_{T,\mathcal{A},A,c_{gt}}(X)$
as
$R_{T}(X)$.

The set $R_T(X)$ consists of images $Im$ such that
all attractors of the
basins intersecting with
the ball $B_T(Im)$ have the same ground truth label.
Note that
 $R_T(X)$ does not depend on
the baseline classifier $M$
used in construction of
the CONN classifier.

The rationale behind the introduced term will follow from \conjectureref{conject1} below.

\begin{definition}[Label-Distinct Basins]
\label{def_dif}
We call two basins
of attractors from $\mathcal A$
label-distinct if their attractors have different
ground truth labels.
\end{definition}

The proof of the following lemma is straightforward.

\begin{lemma}
\label{lemma_RR}
The set $R_T(X)$ consists of the points of
basins $x$ such that the distance from the point
to any basin $y$ label-distinct from $x$ is
greater than or equal to $T$.
\qed
\end{lemma}

\begin{figure}[]
    \centering
    {
      \subfigure[]{\label{fig:BasinsSepartionIndex_large}
        {\includegraphics[width=0.36\linewidth]{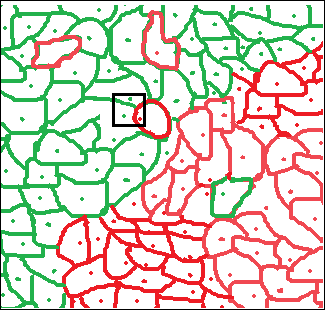}}}
      \qquad
      \subfigure[]{\label{fig:BasinsSepartionIndex}
        {\includegraphics[width=0.4\linewidth]{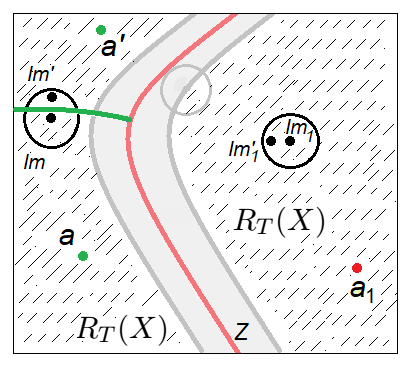}}}
    }

\caption{
Illustration of the subset with guaranteed resilience of the vanilla CONN two-class classifier.
(a): A partition of the image space $X$ by attractor basins is shown. Attractors are marked by points colored according to their assigned ground truth labels ("green" and "red").
(b): A zoom-in of the fragment of $X$ indicated by the black rectangle in (a). Attractors $a$, $a'$, and $a_1$ have labels "green," "green," and "red," respectively.
The green line represents the boundary between the
basins of $a$ and $a'$, while the red line marks both
the boundary between the basins of $a'$ and $a_1$, and
the boundary between the basins of $a$ and $a_1$.
Several balls with radius $T$ are shown as circles. The intersection of the zoomed fragment with $R_T(X)$ is marked by dashed hatching. The set $Z$ (in gray) consists of centers of the $T$-radius balls that contain points from at least two distinct basins.
$Z \subseteq Z_T(X)$
(\equationref{eq:D}).
An attack against $x \in Z$ is not
guaranteed to fail
}
    \label{fig:BasinsSepartionIndexes}
\end{figure}

A fragment of a set $R_T(X)$
for $\mathcal{A}$, which includes three attractors
$a$, $a'$, and $a_1$,
is depicted
in \figureref{fig:BasinsSepartionIndex}
in dashed hatching.
In this hypothetical scenario,
images $Im$
and $Im_1$  belong to
$R_T(X)$.

For image $Im$ in the figure, there are two attractors, $a$ and $a'$, such that $B_T(Im)$ contains elements from their respective basins. Both attractors are assigned a "green" ground truth label.
For image $Im_1$, there is one attractor, $a_1$, such that $B_T(Im_1)$ contains elements from its basin, and $a_1$ is assigned a "red" ground truth label.

We want to show that
an adversarial attack against
the images in $R_T(X)$
cannot succeed.
Let us first note that if an $Im \in
R_T(X)$ is attacked by an adversarial example
$Im'$, then $Im'$ belongs to an open $T$-ball around
$Im$. Two options are possible: either $Im'$ belongs to the basin
of some attractor $a' \in \mathcal{A}$, or $Im'$ does
not belong to the basin of any attractor.

Consider the
former case first.

\begin{lemma}
\label{lemma_adv}
Let $\mathcal{A}$ be a
the set of training examples
of an autoencoder $A$
memorized  as attractors.
Let  $C$ be
a vanilla CONN classifier
constructed for $A$.
Let an
element $Im$
from $R_T(X)$
belongs to basin of
an attractor $a \in \mathcal{A}$.
Then an adversarial attack against
classifier $C$ targeting $Im$,
using an adversarial example $Im'$ that
belongs to the basin of an attractor $a' \in \mathcal{A}$,
cannot be successful.

\begin{proof}
Consider an adversarial
example $Im'$ as stated in the lemma's condition.
Let $C(x)$ denote
the label assigned by classifier $C$ to a sample $x \in X$.

Since $Im \in R_T(X)$, it holds
\[
c_{gt}(a)=c_{gt}(a').
\]

By \remarkref{rem:M},
\begin{align*}
& C(a) = c_{gt}(a), \\
& C(a') = c_{gt}(a').
\end{align*}


By
\remarkref{rem:the_same_label},
\begin{align*}
& C(Im) = C(a), \\
& C(Im') = C(a').
\end{align*}


Therefore,
\[
C(Im)=C(Im'),
\]
and thus,
the attack against $Im$
using $Im'$
can not be successful.
\end{proof}
\end{lemma}


Refer now to the latter
case of an
attack against $Im \in R_T(X)$ using
an adversarial example $Im'$,
which does not belong to the
basin of
a memorized training example.
(In \figureref{fig:BasinsSepartionIndex},
such $Im'$ are marked by 
are marked by a portion of the green curve within the $T$-ball
around $Im$.)

Our 
defense strategy
includes 
the testing whether the sequence of
\equationref{eq:nakaonetc Im'}
converges to a training
example memorized as an
attractor.\footnote{
Practically, the attractors
for the vanilla classifier may be
constructed following
\equationref{eq:nakaonetc}, with
iterations terminated at a
predefined large number $n$. In
\citet{isn}, Section 7.3, for $n =
100$, subsequent members of the
sequence are indistinguishable.
A similar approach is used in
\citet{Belkin,B2_paper}.}
 If this condition is not met, we
classify the sample as "suspicious of an adversarial attack",
 and assign it a randomly selected label, resulting in the attack's failure (see \remarkref{rem:infer}).

In this way, in both aforementioned cases,
the attack against $Im$
cannot be successful, leading us to the following
\begin{conjecture}
\label{conject1}
Let  $C$ be
a vanilla CONN classifier
constructed for $A$.
Then, a classical adversarial attack
against $Im \in R_T(X) $
cannot be successful.
\end{conjecture}

This explains the name given
to $R_T(X)$ in the section title.

\subsection{Class Separation Index}
\label{resil_vanilla_index}

We continue considering a vanilla CONN classifier $C$ constructed for the set $\mathcal{A}$ of training examples memorized as attractors. From \conjectureref{conject1}, it follows that the resilience of $C$ may be characterized by the "proportion" of the subset with guaranteed resilience, $R_T(X)$, within $X$. In this section, we explain why this measure is high for small sets $\mathcal{A}$ with sufficiently large distances between examples.

We proceed as follows.
In \sectionref{Formalization of the Measure}, we formalize the notion of
this "proportion" in $X$
as the class separation index
(\definitionref{def_class_Separation_Index}.)
In \sectionref{Lower},
we consider scenarios
of small subsets with guaranteed resilience.
Although this section is not logically necessary,
it prepares us for
\sectionref{R_is_large},
where we
explain why
$R_T(X)$
is high for the
autoencoders under consideration.

\subsubsection{Formalization of the Measure}
\label{Formalization of the Measure}

Note that,
in the absence of straightforward
methods to measure $R_T(X)$
using conventional measures,
such as the Lebesgue measure,
accurately defining this
proportion may not be feasible.
However, we may avoid this
difficulty as follows.

Let $X' \subset X$ denote a set of images that serve as potential targets for adversarial attacks within a computer environment.
 Since the storage capacity for images in the environment is finite,
 $X'$ is necessarily finite.
This allows us to represent
the "proportion" of
$R_T(X) $ in $X$
using the following
definition:
\begin{definition}[Class  Separation Index
of an Autoencoder]
\label{def_class_Separation_Index}
We define the Class Separation Index $I$ of
an autoencoder $A$ as the ratio
of the number of
images in the intersection of $R_T(X)$ and
$X'$ to the total number of images in $X'$:
\begin{equation}
\label{eq:ind}
I=
\frac
{|R_T(X) \cap X'|}
{| X'|}
.
\end{equation}
\end{definition}

By
\conjectureref{conject1}, larger index values indicate a
greater proportion of images where adversarial
attacks cannot succeed. A value close to 0 means
the system is vulnerable, while a value near 1
indicates that the system covers most of the
input space, making it resilient to such attacks.
Thus, the index serves as a measure of the
system's robustness against adversarial threats.

Note that high values of the index alone do not necessarily indicate higher classifier accuracy. Indeed, a trained classifier with a high index value may be treated as a classifier for a new task with arbitrary new class labels added. Clearly, the index value of the classifier will remain unchanged. However, applying the classifier to test examples with the new labels, which were not included in the training, may result in many misclassifications.

\subsubsection{Scenarios of Low Class Separation Index}
\label{Lower}

Below we consider the scenarios where
$R_T(X)$ is small. We will do this by considering
the complement to $R_T(X)$ (\lemmaref{lemma_compl}).

\begin{remark}
\label{rem:shortnad}
When we refer to the value of $R_T(X)$
as small or large, it is a shorthand notation.
We are actually referring to the corresponding
proportion in $X'$ as defined in
\equationref{eq:ind}. The same note
applies to other measures discussed below.
\end{remark}

The following definition will allow us to represent the complement to $R_T(X)$.

\begin{figure}[]
    \centering
    {
      \subfigure[]{\label{fig:sharedball}
        {\includegraphics[height=0.275\linewidth]{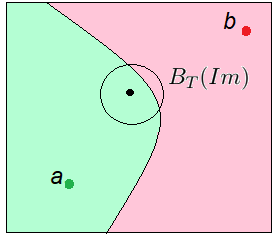}}}
      \qquad
      \subfigure[]{\label{fig:BasinsSepartionIndex3}
        {\includegraphics[height=0.275\linewidth]{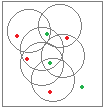}}}
    }
    \caption{
(a): Illustration of the shared $T$-ball as defined in
\definitionref{def_shared}.
(b): As inter-class distances become sufficiently small,
the size of the set of centers of $T$-balls tends to
increase. The drawing illustrates a fragment of the
image space with training examples assigned
"green" and "red" labels. Several shared balls are
denoted by gray circles. Elements of $Z_T(X)$ occupy
a significant part of the fragment
   }
    \label{fig:2figs}
\end{figure}

\begin{definition}[Shared $T$-ball]
\label{def_shared}
Assume the conditions
of
\definitionref{def_R}.
If for an image $Im \in H_A$, the $T$-ball
$B_T(Im)$ shares common points
with two label-distinct basins,
we call $B_T(Im)$ a shared $T$-ball.\end{definition}
The definition is illustrated in
\figureref{fig:sharedball}.

\begin{figure}[H]
    \centering
    {
      \subfigure[]{\label{fig:BasinsSepartionIndex4}
        {\includegraphics[width=0.3\linewidth]{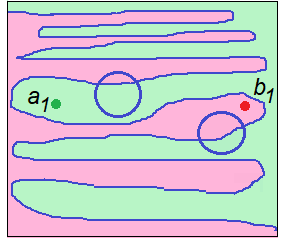}}}
      \qquad
      \subfigure[]{\label{fig:BasinsSepartionIndex6}
        {\includegraphics[width=0.3\linewidth]{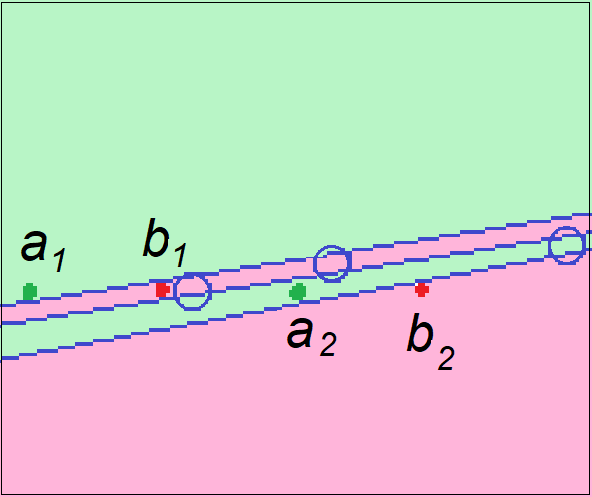}}}
    }
\caption{
Hypothetical scenarios where
large inter-class distances
do not lead to a small
size of $Z_T(X)$.
Depicted are image spaces
where memorized training examples are assigned
"green" and "red" ground truth labels.
Basins corresponding to these labels
are marked in green and red, respectively.
Sample $T$-balls are shown as blue circles.
(a):
Centers of shared $T$-balls span
the union of all basin regions
in the  fragment shown.
(b):
Centers of shared $T$-balls span the entire basins of examples $b_1$ and $a_2$ in the fragment shown
}
    \label{fig:BasinsSepartionIndex23}
\end{figure}

Define the subset of $H_A$
(see
\equationref{eq:H_A}
):
\begin{equation}
\label{eq:D}
\begin{aligned}
&\hspace{-0em}
Z_T(X) = \{ Im \in X \;|\;
 \text{
$B_T(Im)$
is a shared $T$-ball.
}
\end{aligned}
\end{equation}

The lemma below  directly follows  from
\lemmaref{lemma_RR}:
\begin{lemma}
\label{lemma_compl}
$R_T(X) \cup Z_T(X) = H_A$.
\qed
\end{lemma}

Thus, we may estimate $R_T(X)$ using $Z_T(X)$:
the larger $R_T(X)$ is, the smaller $Z_T(X)$ is, and vice versa.

Note that inter-class distances less than $T$
lead to a larger size $Z_T(X)$, since every
$T$-ball that includes examples with different
ground truth labels is shared. This is
illustrated in \figureref{fig:2figs}.

Therefore, large inter-class distances
are a necessary condition for a large $R_T(X)$.

Recall that our analysis is limited to
autoencoders trained on small sets of examples
with sufficiently large inter-example distances.
This is intended to ensure large inter-class
distances, thereby satisfying the necessary
condition for a large $R_T(X)$.

But does large inter-class distance
always lead to a
small size of $Z_T(X)$?
In \figureref{fig:BasinsSepartionIndex23}
we illustrate
scenarios where, despite
the large inter-class distances,
the centers of the shared
$T$-balls occupy a significant
part of the image space.
Thus, $Z_T(X)$ will be high
and consequently $R_T(X)$
will be low.

In \sectionref{R_is_large}, we motivate our assumption that overparameterized autoencoders  have high values of $R_T(X)$.

\subsubsection{High
Class Separation Index
Values at Small Training Datasets}
\label{R_is_large}

In this section we
consider a subset
of $R_T(X)$  and explain why
this subset is large for
overparameterized
autoencoders,
trained on small sets of
examples with sufficiently
large distances between them.
This will explain
the intrinsic resilience of vanilla
CONN classifiers to classical
adversarial attacks.

\begin{definition}[T-Interior of Basins of an Autoencoder]
\label{def_T}
Let $\mathcal{A}$ be a
the set of training examples
of an autoencoder $A$
memorized  as attractors.
We define the subset of $X$ as follows:
\begin{equation}
\label{eq:S_1_GAG}
\begin{aligned}
&\hspace{-0em}
\widehat{R}_T(X)
 = \{ Im \in X \;|\;
 \text{
$B_T(Im)$ is a subset of
a basin of
some $a \in  \mathcal{A}$ }
\}
.
\end{aligned}
\end{equation}
We refer to \(\widehat{R}_T(X)\) as the T-Interior of the basins of the autoencoder.
\end{definition}

\begin{figure}[H]
    \centering
    {
      \subfigure[]{\label{fig:BasinsSepartionIndex_large_B}
        {\includegraphics[width=0.4\linewidth]{BasinsInterIndex2.png}}}
      \qquad
      \subfigure[]{\label{fig:BasinsInterIndex2_purely}
        {\includegraphics[width=0.4\linewidth]{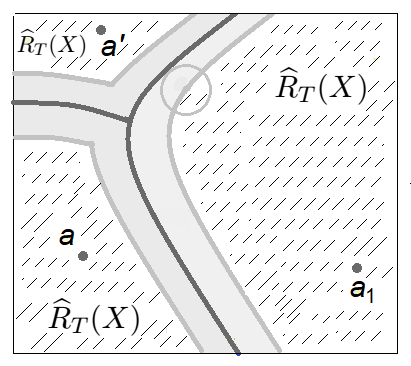}}}
    }
\caption{ Illustration of $\widehat{R}_T(X)$ and \equationref{eq:z_in_z}. Both drawings show a zoomed-in fragment, indicated by the black rectangle in \figureref{fig:BasinsSepartionIndex_large}. (a): Copy of \figureref{fig:BasinsSepartionIndex}, shown here for comparison with (b). (b): Illustration of the T-Interior of basins for an autoencoder. The intersection of the zoomed fragment with $\widehat{R}_T(X)$ is marked by dashed hatching
}
    \label{fig:BasinsInterIndex2_purely_comp}
\end{figure}

The definition is illustrated in
\figureref{fig:BasinsInterIndex2_purely}.
Note that $\widehat{R}_T(X)$
depends
on the basins of
the autoencoder and
does not depend on a label
assignment on $X$.

Using
\lemmaref{lemma_RR},
it is easily shown that
\begin{equation}
\label{eq:z_in_z}
\widehat{R}_T(X) \subseteq
R_T(X)
.
\end{equation}
See
\figureref{fig:BasinsInterIndex2_purely_comp}.

Concerning the note at the beginning of
Section \ref{Lower}, is $\widehat{R}_T(X)$
large in our task?

Experimental results by \citet{Belkin} visualize the partition of the image space by basins of an overparameterized autoencoder trained on a small number of examples.  The basins appear well-separated, each with distinct internal regions. We assume these findings represent the general case; thus, we expect a large $T$-interior for such autoencoders when trained on perceptually distinct examples.

This confirms the intuition regarding autoencoders
trained on a small number of perceptually distinct
training examples, as discussed in the
introduction to \sectionref{resil_vanilla}. Namely,
for an arbitrarily selected $Im$ such that
\equationref{eq:nakaonetc_B} converges to an
attractor $a$, it is likely that for any
 perceptually indistinguishable image  $Im'$, the sequence
\equationref{eq:nakaonetc Im'} will also
converge to $a$. One may see that this is actually another way of stating
that $\widehat{R}_T(X)$ is large.

Larger values of $\widehat{R}_T(X)$ result in larger values of the subset of guaranteed resilience $R_T(X)$, and hence larger class separation indices of the autoencoders.

We, therefore, assume that vanilla CONN classifiers
trained on small datasets possess intrinsic
resilience to classical perturbation attacks.

\bibliographystyle{unsrtnat}
\bibliography{SI_arxiv}